\documentclass[twoside,11pt]{article}

\newcommand{\cO}{\mathcal{O}}

\newcommand{\EE}{\mathbb{E}}
\newcommand{\RR}{\mathbb{R}}

\newcommand{\PP}{\mathbb{P}}

\newcommand{\TG}{\widetilde{G}}
\newcommand{\BG}{\bar{G}}

\newcommand{\HSig}{\hat{\Sigma}}
\newcommand{\Hu}{\hat{\mu}}

\newcommand{\tg}{\tilde{g}}
\newcommand{\argmin}{\mathop{\mathrm{argmin}}}

\newcommand{\tu}{\ti{u}}

\usepackage{bbm}

\def\cS{\mathcal{S}}

\newcommand{\mL}{\mathscr{L}}
\newcommand{\bu}{\bar{u}}

\usepackage{bm, mathrsfs, graphics,float,amssymb,amsmath,subeqnarray,setspace,graphicx,epstopdf,subfigure, enumerate, color}
\usepackage{multirow}
\usepackage{tablefootnote}
\usepackage{colortbl}
\usepackage{hhline}

\usepackage{algorithm}
\usepackage{algorithmic}
\usepackage{bbm}

\usepackage[abbrvbib, preprint]{jmlr2e}

\newcommand{\ti}[1]{\tilde{#1}}
\def\diag{\mathrm{diag}}
\newcommand{\norm}[1]{\left\|#1\right\|}
\newcommand{\dotprod}[1]{\left\langle #1\right\rangle}
\def\tr{\mathrm{tr}}

\ShortHeadings{Mirror Natural Evolution Strategies}{Haishan Ye}
\firstpageno{1}

\begin{document}

\numberwithin{equation}{section}

\title{Mirror Natural Evolution Strategies}

\author{\name 	Haishan Ye \email hsye\_cs@outlook.com \\
	\addr Shenzhen Research Institute of Big Data\\
	The Chinese University of Hong Kong, Shenzhen}

\editor{XXX and XXX}

\maketitle

\begin{abstract}
	The zeroth-order optimization has been widely used in machine learning applications. However, the theoretical study of the zeroth-order optimization  focus on the algorithms which approximate (first-order) gradients using (zeroth-order) function value difference at a random direction. 
	The theory of algorithms which  approximate the gradient and Hessian information by zeroth-order queries is much less studied.
	In this paper, we focus on the theory of zeroth-order optimization which utilizes both the first-order and second-order information approximated by the zeroth-order queries.
	We first propose a novel reparameterized objective function with parameters $(\mu, \Sigma)$. 
	This reparameterized objective function achieves its optimum at the minimizer and the Hessian inverse of the original objective function respectively, but with small perturbations.
	Accordingly, we propose a new algorithm to minimize our proposed reparameterized objective, which we call \texttt{MiNES} (mirror descent natural evolution strategy).
	We show that the estimated covariance matrix of \texttt{MiNES} converges to the inverse of Hessian  matrix of the objective function with a  convergence rate $\widetilde{\cO}(1/k)$, where $k$ is the iteration number and $\widetilde{\cO}(\cdot)$ hides the constant and $\log$ terms.
	We also provide the explict convergence rate of \texttt{MiNES} and how the covariance matrix promote the convergence rate.
\end{abstract}


\section{Introduction}

The zeroth-order optimization has been widely used in machine learning applications such as deep reinforcement learning \citep{salimans2017evolution,conti2018improving}, the black-box adversarial attack of deep neural networks \citep{ilyas18a,tu2019autozoom,dong2019efficient,chen2019boundary}, tuning the hyper-parameters of deep neural networks \citep{loshchilov2016cma} and etc.
Because of its wild applications, the zeroth-order optimization has also been widely studied \citep{Nesterov2017,ghadimi2013stochastic,duchi2015optimal}.
These works mainly use the first-order information, that is, the gradient is approximated by the function value difference at a random direction, to achieve the convergence. 
These works all have explicit convergence rates.

It is well-known that the second-order information is helpful to improve the convergence rate of algorithms. 
Can we approximate the Hessian (inverse) by the zeroth-order queries?
Evolution Strategies such as \texttt{CMA-ES} (covariance matrix adaptation evolution strategy) and \texttt{NES} (natural evolution strategy) use the zeroth-order information to approximate both the gradient and Hessian matrix.
Thus, in practice, \texttt{CMA-ES} and \texttt{NES} outperform the algorithms which only use the zeroth-order information to approximate the gradient \citep{hansen2016cma}.
However, to obtain the explicit convergence rates of this kinds of algorithms, especially the convergence rates of the covariance matrix, are difficult. 

\citet{wierstra2014natural} propose a reparameterized function $J(\theta)$ with parameters $\theta = (\mu, \Sigma)$. 
Based on it,  \citet{Akimoto2012} tries to derive the convergence rates of the covariance matrix of \texttt{NES}. 
However, in the analysis, it requires to use the gradient of $J(\theta)$ with respect to the  covariance matrix but not the approximate one which is used in the practical \texttt{NES}-type algorithms by zeroth-order queries.
Furthermore, the reparameterized function $J(\theta)$ proposed in \citep{wierstra2014natural} is not good enough to help design and understand zeroth-order algorithms with  the both gradient and Hessian being approximated by the zeroth-order queries. 
For the quadratic function, the derivative of $J(\theta)$ with respect to $\Sigma$ equals to the Hessian. 
Thus, we can not obtain the useful information by the first order necessary condition of optimization. 
What we want to is the solution to the equation $\frac{\partial J(\cdot)}{\partial \Sigma} = 0$ equals to the Hessian inverse.

Thus, it remains open to propose a reparameterized function who can help design and understand zeroth-order algorithms with  the both gradient and Hessian being approximated by the zeroth-order queries.
It is also an open question whether one can design a zeroth-order algorithm with  explicit convergence rates both for the function value and the covariance matrix, which only uses the zeroth-order information to approximate both the gradient and Hessian matrix.
In this paper, we give affirmative answers to above open questions and summarize our contributions as follows:
\begin{enumerate}
	\item Based on the reparameterized objective function $J(\theta )$ proposed in \citep{wierstra2014natural}, we improve it and propose a regularized objective function $Q_\alpha(\theta)$. 
	We show that the mean vector and the  covariance part of the minimizer of $Q_\alpha(\theta)$  are close to the minimum of the objective function and  close to the Hessian inverse up to some perturbations up to at most $\cO(\alpha)$. 
	These properties of $Q_\alpha(\theta)$ make the mathematical foundation for designing query efficient algorithm. 
	\item  Based on this new objective function, we propose a novel algorithm called \texttt{MiNES}, which guarantees that the covariance matrix converges to the inverse of the Hessian. Instead of using natural gradient descent, we resort to the mirror descent to update the covariance matrix. 
	\item We provide a convergence analysis of \texttt{MiNES} and give the explicit convergence rates of \texttt{MiNES}. Especially, we show the inverse of covariance matrix will converge to the Hessian with  a rate $\widetilde{\cO}(1/k)$.
	To the best of our knowledge, this is the first explicit convergence rate for the covariance matrix in this kind of algorithms.
	Our convergence analysis also shows how the  covariance matrix helps the algorithm converge to the minimizer of the objective function.
\end{enumerate}

\paragraph{More Related Literature.}

Zeroth-order (derivative-free) optimization  has a long history \citep{matyas1965random,kiefer1952stochastic}.
The idea behind these zeroth-order algorithms is to create a stochastic oracle to approximate (first-order) gradients using (zeroth-order) function value difference at a random direction, and then apply the update rule of (sub-)gradient descent \citep{Nesterov2017,ghadimi2013stochastic,duchi2015optimal}. 
On the other hand, \citet{conn2009global} propose to  utilize curvature information in constructing quadratic approximation model under a slightly
modified trust region regime.
Recently, \citet{ye2018hessian} propose the \texttt{ZOHA} algorithm which utilizes curvature information of the objective function. 
They show that the approximate Hessian can help to improve the convergence rate over zeroth-order algorithms without second-order information.
However, the work of \citet{ye2018hessian} requires extra queries to construct the approximate Hessian.
In contrast,  no extra   queries are needed to approximate the Hessian in our work. 
Thus, our \texttt{MiNES}  is more query-efficient.

Evolutionary strategies (\texttt{ES}) are another important class of zeroth-order algorithms for optimization problems that only have access to function value evaluations.
ES attracts much attention since it was introduced by Ingo Rechenberg and Hans-Paul Schwefel in the 1960s and 1970s \citep{schwefel1977numerische}, and many variants have been proposed \citep{beyer2001self,hansen2001completely,wierstra2008natural,glasmachers2010exponential}. 
\texttt{ES} tries to evaluate the fitness of real-valued genotypes in batches, after which only the best genotypes are kept and used to produce the next batch of offsprings. 
A covariance matrix is incorporated into evolutionary strategies to capture the dependency variables so that independent `mutations' can be generated for the next generation. 
In this general algorithmic framework, the most well-known algorithms are the covariance matrix adaptation evolution strategy (\texttt{CMA-ES}) \citep{hansen2001completely} and natural evolution strategies (\texttt{NES}) \citep{wierstra2008natural}.

There have been many efforts to better understand these methods in the literature \citep{Beyer14,OllivierAAH17,auger2016linear,malago2015information,AkimotoNOK10}. 
For example, the authors in \citet{AkimotoNOK10} revealed a connection between \texttt{NES} and \texttt{CMA-ES}, and showed that  \texttt{CMA-ES}  is a special version of \texttt{NES}. 
\texttt{NES} can be derived from information geometry because it employs a natural gradient descent \citep{wierstra2014natural}. 
Therefore,  the convergence of \texttt{NES} and \texttt{CMA-ES} has been studied from the information geometry point of view  \citep{Beyer14,auger2016linear}. 
\citet{Beyer14} analyzed the convergence of \texttt{NES} with infinitestimal learning rate using ordinary differential equation with the objective function being quadratic and strongly convex. However,
it does not directly lead to a convergence rate result with finite learning rate. 
Moreover, \citet{Beyer14} does not show how the covariance matrix converges to the inverse of the Hessian, although this is conjectured in \citep{hansen2016cma}. 
\citet{Akimoto2012} proves that  the covariance matrix of \texttt{NES} will converge to the Hessian inverse with a geometric convergence rate under the condition that \texttt{NES} takes a natural gradient.
However, in practice, \texttt{NES} takes a stochastic natural gradient to update the covariance matrix. 
In contrast, our work gives an explicit convergence rate of the covariance matrix update of \texttt{MiNES} using the stochastic mirror gradient descent obtained by zeroth-order queries.  
\citet{Glasmachers2022} show that the covariance matrix of Hessian Estimation Evolution Strategies converge to the Hessian inverse but no explicit rate is provided.

Recently, \citet{auger2016linear} showed that for strongly convex functions, the estimated mean value in ES with comparison-based step-size adaptive randomized search (including \texttt{NES} and \texttt{CMA-ES}) can achieve linear convergence. However, \citet{auger2016linear} have not shown how covariance matrix converges and how covariance matrix affects the convergence properties of the estimated mean vector. Although it has been conjectured that for the quadratic function, the covariance matrix of \texttt{CMA-ES} will converge to the inverse of the Hessian  up to a constant factor, no rigorous proof has been provided \citep{hansen2016cma}.

The derivative free algorithms and evolutionary strategies are closely related even they are motivated from different ideas and are of totally different forms.
For example, to improve the convergence rate of \texttt{NES}, \citet{salimans2017evolution} proposed ‘antithetic sampling’ technique. In this case, \texttt{NES} shares the same algorithmic form with derivative free algorithm \citep{Nesterov2017}. 
\texttt{NES} with ‘antithetic sampling’ is widely used in black-box adversarial attack \citep{tu2019autozoom,ilyas18a}. 
Nevertheless, the mathematical relationship between \texttt{NES} and derivative free algorithms have not been explored.

\noindent\textbf{Organization.}
The rest of this paper is organized as follows.
In Section~\ref{sec:background}, we introduce the background and preliminaries will be used in this paper.
In Section~\ref{sec:ROF}, we propose a novel regularized objective function and prove that the mean part of its minimizer is close to the minimizer of the original problem and the covariance part is close to the corresponding Hessian inverse.
Section~\ref{sec:MiNES} gives the detailed description of the proposed mirror natural evolution strategies.
In Section~\ref{sec:conv}, we analyze the convergence properties of \texttt{MiNES}.
Finally, we conclude our work in Section~\ref{sec:conclusion}.
The detailed proofs are deferred to the appendix in appropriate orders. 

\section{Background and Preliminaries}
\label{sec:background}
In this section, we will introduce the natural evolutionary strategies and preliminaries.

\subsection{Natural Evolutionary Strategies}

The Natural Evolutionary Strategies (\texttt{NES}) reparameterize the objective function $f(z)$ ($z\in\RR^d$) as follows \citep{wierstra2014natural}:
\begin{equation}\label{eq:J_org}
	J(\theta) = \EE_{z \sim \pi(\cdot|\theta)} [f(z)] = \int f(z)\pi(z|\theta)\; dz,
\end{equation}
where $\theta$ denotes the parameters of density $\pi(z|\theta)$ and $f(z)$  is commonly referred as the fitness function for samples $z$.
Such transformation can help to develop algorithms to find the minimum of $f(z)$ by only accessing to the function value. 

\paragraph{Gaussian Distribution and Search Directions.}

In this paper, we will  only investigate the Gaussian distribution, that is, 
\begin{equation}
	\label{eq:pi_z}
	\pi(z|\theta) \sim N(\mu,\bar{\Sigma}).
\end{equation}
Accordingly, we have
\begin{equation}\label{eq:z}
	z = \mu + \bar{\Sigma}^{1/2} u, \quad u\sim N(0, I_d),
\end{equation}
where $d$ is the dimension of $z$.
Furthermore, the density function $\pi(z|\theta)$ can be presented as 
\begin{equation*}
	\pi(z|\theta) = \frac{1}{\sqrt{(2\pi)^d\det(\bar{\Sigma})}}\cdot \exp\left(-\frac{1}{2}(z - \mu)^\top \bar{\Sigma}^{-1}(z-\mu)\right)
\end{equation*}
In order to compute the derivatives of $J(\theta)$, we can use the so-called `log-likelihood trick' to obtain the following \citep{wierstra2014natural}
\begin{align}
	\nabla_\theta J(\theta) = \nabla_\theta \int f(z) \pi(z|\theta)\; dz
	=\EE_z \big[ f(z) \nabla_\theta \log \pi(z|\theta) \big]. \label{eq:log_lh}
\end{align}
We also have that 
\begin{align*}
	\log \pi(z|\theta) = -\frac{d}{2}\log(2\pi) -\frac{1}{2} \log \det \bar{\Sigma} - \frac{1}{2}(z-\mu)^\top \bar{\Sigma}^{-1}(z-\mu).
\end{align*} 
We will need its derivatives with respect to $\mu$ and $\bar\Sigma$, that is, $\nabla_\mu\log\pi(z|\theta)$ and $\nabla_{\bar{\Sigma}}\log\pi(z|\theta)$. The first is trivially
\begin{align}
	\nabla_\mu\log\pi(z|\theta) = \bar{\Sigma}^{-1}(z-\mu) , \label{eq:nab_mu}
\end{align}
while the latter is 
\begin{equation}\label{eq:nab_Sig}
	\nabla_{\bar{\Sigma}}\log\pi(z|\theta) = \frac{1}{2}\bar{\Sigma}^{-1}(z-\mu)(z-\mu)^\top\bar{\Sigma}^{-1} - \frac{1}{2} \bar{\Sigma}^{-1}.
\end{equation}
Let us denote 
\begin{equation*}
	\theta = [\mu^\top,\;\mathrm{vec}({\bar{\Sigma}})^\top]^\top,
\end{equation*}
where $\theta\in\RR^{d(d+1)}$-dimensional column vector consisting of all the elements of the mean vector $\mu$ and the covariance matrix $\bar{\Sigma}$. $\mathrm{vec}(\cdot)$ denotes a rearrangement operator from  a matrix to a column vector.
The Fisher matrix with respect to $\theta$ of $\pi(z|\theta)$  for a Gaussian distribution is well-known \citep{AkimotoNOK10}, 
\begin{align}
	F_\theta 
	= 
	\EE_z \left[ \nabla_\theta \log \pi(z|\theta)  \nabla_\theta \log \pi(z|\theta)^\top\right] 
	=
	\left[
	\begin{array}{cc}
		\bar{\Sigma}^{-1}, & 0 \\
		0, &\frac{1}{2}\bar{\Sigma}^{-1}\otimes\bar{\Sigma}^{-1}
	\end{array}
	\right] \notag
\end{align}
where $\otimes$ is the Kronecker product. 
Therefore, the natural gradient of the log-likelihood of $\pi(z|\theta)$ is 
\begin{equation}
	\label{eq:nat_grad}
	F^{-1}_\theta \nabla_\theta \log \pi(z|\theta) 
	=
	\left[
	\begin{array}{c}
		z - \mu\\
		\mathrm{vec}\left((z-\mu)(z-\mu)^\top - \bar{\Sigma}\right)
	\end{array}
	\right]
\end{equation} 
Combining Eqn.~\eqref{eq:log_lh} and \eqref{eq:nat_grad}, we obtain the estimate of the natural gradient from samples $z_1, \dots, z_b$ as
\begin{equation}
	\label{eq:ng}
	F^{-1}_\theta \nabla J(\theta) 
	\approx \frac{1}{b}\sum_{i=1}^{b} f(z_i) 
	\left[
	\begin{array}{c}
		z_i - \mu\\
		\mathrm{vec}\left((z_i-\mu)(z_i-\mu)^\top - \bar{\Sigma}\right)
	\end{array}
	\right]
\end{equation} 
Therefore, we can obtain the meta-algorithm of \texttt{NES} (Algorithm 3 of \citet{wierstra2014natural} with $F^{-1}_\theta \nabla J(\theta) $ approximated as Eqn.~\eqref{eq:ng})
\begin{equation*}
	\left\{
	\begin{aligned}
		\mu =&  \mu - \eta \cdot\frac{1}{b}\sum_{i=1}^{b} f(\mu+\bar{\Sigma}^{1/2}u_i) \bar{\Sigma}^{1/2}u_i\\
		\bar{\Sigma} =& \bar{\Sigma} - \eta\cdot\frac{1}{b}\sum_{i=1}^{b} f(\mu+\bar{\Sigma}^{1/2}u_i) \left(\bar{\Sigma}^{1/2}u_i u_i^\top \bar{\Sigma}^{1/2} - \bar{\Sigma}\right),
	\end{aligned}
	\right.
\end{equation*}
where $\eta$ is the step size.

\subsection{Notions}

Now, we introduce some important notions which is widely used in optimization.
\paragraph{$L$-smooth} 
A function $f(\mu)$ is \textit{$L$-smooth}, if it holds that,  for all $\mu_1,\mu_2\in\RR^d$
\begin{equation}\label{eq:L_1}
	\norm{\nabla f(\mu_1) - \nabla f(\mu_2)} \leq L\norm{\mu_1 - \mu_2}
	.
\end{equation}

\paragraph{$\sigma$-Strong Convexity} 
A function  $f(\mu)$ is \textit{$\sigma$-strongly convex}, if it holds that, for all $\mu_1,\mu_2\in\RR^d$
\begin{equation}\label{eq:sig_1}
	f(\mu_1) - f(\mu_2) \geq \dotprod{\nabla f(\mu_2), \mu_1 - \mu_2} + \frac{\sigma}{2}\norm{\mu_1 - \mu_2}^2.
\end{equation}

\paragraph{$\gamma$-Lipschitz Hessian}
A function $f(\mu)$ \textit{admits $\gamma$-Lipschitz Hessians} if it holds that,  for all $\mu_1,\mu_2 \in \RR^d$, it holds that 
\begin{align}
	\norm{\nabla^2 f(\mu_1) - \nabla^2f(\mu_2)} \leq \gamma \norm{\mu_1-\mu_2}
	.
	\label{eq:gamma_1}
\end{align}

Note that $L$-smoothness and $\sigma$-strongly convexity imply $\sigma I\preceq \nabla^2 f(\mu) \preceq L I$.

\section{Regularized Objective Function}
\label{sec:ROF}

Conventional \texttt{NES} algorithms are going to minimize $J(\theta)$ (Eqn.~\eqref{eq:J_org}) \citep{wierstra2014natural}. 
Instead, we propose a novel regularized objective function to reparameterize $f(z)$:
\begin{equation}\label{eq:Q}
	Q_\alpha(\theta) = J(\theta) - \frac{\alpha^2}{2} \log\det\Sigma,
\end{equation}
where $\alpha$ is a positive constant. 
Furthermore, we represent $\bar{\Sigma}$ in Eqn.~\eqref{eq:pi_z} as $\bar{\Sigma} = \alpha^2\Sigma$. 
Accordingly, $\theta(\mu, \Sigma)$ denotes the parameters of a Gaussian density $\pi(z|\theta) = N(\mu,\alpha^2 \Sigma)$.
By such transformation, $J(\theta)$ can be represent as 
\begin{equation*}
	J(\theta) = \EE_u [f(\mu + \alpha \Sigma^{1/2}u)], \quad\mbox{with}\quad u\sim N(0,I_d).
\end{equation*}
Then $J(\theta)$ is the \emph{Gaussian approximation function} of $f(z)$ and  $\alpha$  plays a role of smoothing parameter \citep{Nesterov2017}.
Compared with $J(\theta)$, $Q_\alpha(\theta)$ has several advantages and we will first introduce the intuition we propose   $Q_\alpha(\theta)$.

\paragraph{Intuition Behind $Q_\alpha(\theta)$}

Introducing the regularization brings an important benefit which can help to clarify the minimizer of $\Sigma$. 
This benefit can be shown when $f(z)$ is a quadratic function where $f(z)$ can be expressed as  
\begin{equation}\label{eq:quad}
	f(z) = f(\mu)+\dotprod{\nabla f(\mu), z-\mu}+ \frac{1}{2}(z-\mu)^\top  H (z-\mu) ,
\end{equation}
where $H=\nabla^2 f(\mu)$ denotes the Hessian matrix. 
Note that when $f(z)$ is quadratic, the Hessian matrix is independent on the value of $z$. 
In the rest of this paper, we will use $H$ to denote the Hessian matrix of a quadratic function.
Since we have $z = \mu+\alpha\Sigma^{1/2}u$ (by Eqn.~\eqref{eq:z}), $J(\theta)$ can be explicitly expressed as
\begin{align}
	J(\theta) = \EE_u\left[f(\mu)+\alpha\dotprod{\nabla f(\mu), \Sigma u}+\frac{\alpha^2}{2}u^\top \Sigma^{1/2} H \Sigma^{1/2}u\right]
	=f(\mu) + \frac{\alpha^2}{2}\dotprod{H, \Sigma}. \label{eq:J_quad}
\end{align}
where $\dotprod{A,B} = \tr(A^\top B)$.
By setting $\nabla_{\theta} Q_\alpha(\theta) = 0$, we can obtain that
\begin{equation*}
	\frac{\partial Q_\alpha}{\partial \mu} = \nabla_\mu f(\mu) = 0, \qquad \frac{\partial Q_\alpha}{\partial \Sigma} = \frac{\alpha^2}{2}H - \frac{\alpha^2}{2} \Sigma^{-1} = 0.
\end{equation*}
Thus, we can obtain that the minimizer $\mu$ of  $Q_\alpha$ is $\mu^*$ --- the minimizer of $f(\mu)$ and the minimizer $\Sigma$ of  $Q_\alpha$ is $H^{-1}$ --- the inverse of the  Hessian matrix.
In contrast, without the regularization, $\frac{\partial Q_\alpha}{\partial \Sigma} $ will reduce to $\frac{\partial J(\theta)}{\partial \Sigma}$:
\begin{equation*}
	\frac{\partial J(\theta)}{\partial \Sigma} =  \frac{\alpha^2}{2}H.
\end{equation*} 
Thus, the  $\partial  J(\theta)/\partial \Sigma$ does not provide useful information about what covariance matrix is the optimum of $J(\theta)$. 

Therefore, in this paper, we will consider the regularized objective function $Q_\alpha(\theta)$. 
To obtain a concise theoretical analysis of the convergence rate of $\Sigma$, we are going to solve the following constrained optimization problem
\begin{equation}\label{eq:modi_prob}
	\min_{\mu\in\RR^d, \Sigma\in \mathcal{S}} Q_\alpha(\theta(\mu,\Sigma))
\end{equation}
with $\cS$ is defined as
\begin{equation}\label{eq:cS}
	\cS = \bigg\{\Sigma\bigg|\zeta^{-1}\cdot I\preceq\Sigma\preceq \tau^{-1}\cdot I \bigg\},
\end{equation}
where $\zeta$ and $\tau$ are positive constants which satisfy $\tau\leq \zeta$. 
Note that, the constraint on $\Sigma$ is used to keep $\Sigma$ bounded and this property will be used in the convergence analysis of $\Sigma$.

In the rest of this section, we show that the mean vector $\Hu^*$ of minimizer of Eqn.~\eqref{eq:modi_prob} is close to $\mu^*$ which is the minimizer of $f(\mu)$. 
Furthermore, if $\zeta$ and $\tau$ in Eqn.~\eqref{eq:cS} are chosen properly, the covariance matrix $\HSig_*$ will be close to the Hessian inverse. 
As a special case, if $f(\cdot)$ is quadratic, then we show that $\Hu^*$ is equal to $\mu^*$ and $\HSig_*$ is equal to $[\nabla^2f(\mu^*)]^{-1}$.

\subsection{Quadratic Case}

We will first investigate the case that function $f(\cdot)$ is quadratic, because the solution in this case is considerably simple. 
\begin{theorem}\label{prop:quad_case}
	If the function $f(\cdot)$ is quadratic so that $f(z)$ satisfies Eqn.~\eqref{eq:quad}. 
	Define  $\Pi_{\cS}(A)$ be the projection of symmetric $A$ on to $\cS$, that is, $\Pi_{\cS}(A) = \argmin_{X\in\cS}\norm{A-X}$ with $\norm{\cdot}$ being the Frobenius norm.
	Let $\mu^*$ be the minimizer of $f(\cdot)$, then the minimizer of $Q_\alpha(\theta)$ is 
	\begin{equation*}
		\left(\mu^*, \Pi_{\cS}\left(H^{-1}\right)\right) = \argmin_{\mu, \Sigma\in\cS} Q_\alpha(\mu, \Sigma).
	\end{equation*}
\end{theorem}

If $\zeta\geq L$ and $\tau\leq \sigma$ in $\cS$ (defined in Eqn.~\eqref{eq:cS}), we can observe that the above proposition shows that the optimal covariance matrix $\Sigma$ is the Hessian inverse matrix. 
\subsection{General Strongly Convex Function with Smooth Hessian} 

Next we will consider the general convex function case with its Hessian being $\gamma$-Lipschitz continuous.
First, we can rewrite $Q_\alpha(\theta)$ as 
\begin{equation*}
	Q_\alpha(\theta) = J(\theta) + R(\Sigma), \quad\mbox{ with } \quad R(\Sigma) = - \frac{\alpha^2}{2} \log\det\Sigma.
\end{equation*}
We will show that the value $Q_\alpha(\theta)$ at $\theta = (\mu, \Sigma)$ is close to $f(\mu)$ when $\alpha$ is sufficiently small.
\begin{theorem}\label{prop:close}
	Let $Q_\alpha(\theta)$ be defined in Eqn.~\eqref{eq:Q}. Assuming that function $f(\cdot)$ is $L$-smooth and its Hessian is $\gamma$-Lipschitz,  then $Q_\alpha(\theta)$ satisfies that
	\begin{equation*}
		f(\mu) - \frac{L\alpha^2}{2} \tr(\Sigma) - \alpha^3\phi(\Sigma) + R(\Sigma) \leq Q_\alpha(\theta) 
		\leq f(\mu) + \frac{L\alpha^2}{2} \tr(\Sigma) + \alpha^3\phi(\Sigma) + R(\Sigma) ,
	\end{equation*}
	where $\phi(\Sigma)$ is defined as
	\begin{equation*}
		\phi(\Sigma) = \frac{1}{(2\pi)^{d/2}}\int_u \frac{\gamma}{6} \|\Sigma^{1/2} u\|^3 \exp\left(-\frac{1}{2}\norm{u}^2\right) d u.
	\end{equation*}
\end{theorem}

By the above proposition, we can observe that as $\alpha\to 0$, $\frac{L\alpha^2}{2} \tr(\Sigma)$, $\phi(\Sigma)$ and $R(\Sigma)$ will go to $0$. Therefore,  instead to directly minimizing $f(\mu)$, we can minimize $Q_\alpha(\theta)$ with $\theta = (\mu, \Sigma)$. 
Next, we will prove that the $\mu$ part of the minimizer of $Q_\alpha(\theta)$ is also close to the solver of $\min f(\mu)$.
\begin{theorem}\label{prop:mu}
	Let $f(\cdot)$ satisfy the properties in Theorem~\ref{prop:close}. $f(\cdot)$ is also $\sigma$-strongly convex. 
	Let $\hat{\mu}_*$ be the minimizer of $Q_\alpha(\theta)$ under constraint $\Sigma\in\cS$.
	$\mu^*$ denotes the optimum of $f(\mu)$. 
	Then, we have the following properties
	\begin{align*}
		f(\hat{\mu}_*) -f(\mu^*) 
		\leq& 
		\frac{dL\alpha^2}{\tau} 
		+ 
		\frac{\gamma \alpha^3 (d+3)^{3/2}}{3\tau^{3/2}} 
		+
		\frac{d\alpha^2}{2} \left(\tau^{-1} - \zeta^{-1}  \right)\\
		\norm{\mu^* - \hat{\mu}_*}^2 
		\leq&
		\frac{2dL\alpha^2}{\sigma\tau} 
		+ 
		\frac{2\gamma \alpha^3 (d+3)^{3/2}}{3\sigma\tau^{3/2}} 
		+
		\frac{d\alpha^2}{\sigma} \left(\tau^{-1} - \zeta^{-1}  \right),
	\end{align*}
	where $d$ is the dimension of $\mu$.
\end{theorem}

Finally, we will provide the properties how well $\Sigma$ approximates the inverse of Hessian  matrix.
\begin{theorem}\label{prop:Sigma}
	Let $f(\cdot)$ satisfy the properties in Theorem~\ref{prop:mu}. $\hat{\theta} = (\Hu^*, \HSig_*)$ is the minimizer of $Q_\alpha(\theta)$ under the constraint $\Sigma\in\cS$. $\mu^*$ is the minimizer of $f(\mu)$ and $\Sigma_*$ is the minimizer of $Q_\alpha(\theta)$ given $\mu = \mu^*$ under the constraint $\Sigma\in\cS$. 
	Assume that $ \alpha$ satisfies 
	\begin{equation}\label{eq:alp_cond}
		\alpha
		\leq
		\frac{3\tau^{3/2}\sigma}{\gamma\zeta}
		\cdot
		\left(
		(d+5)^{5/2} + d (d+3)^{3/2}
		\right)^{-1},
	\end{equation}
	where $d$ is the dimension of $\mu$,
	then $\Sigma_*$ has the following properties
	\begin{equation*}
		\norm
		{
			\HSig_* - \Pi_{\cS}\left(\left(\nabla^2f(\Hu^*) 	
			\right)^{-1}\right)
		}
		\leq 
		\frac{\alpha\gamma\zeta}{3\tau^{3/2}\sigma^2}
		\cdot
		\left(
		(d+5)^{5/2} + d (d+3)^{3/2}
		\right),
	\end{equation*}
	and
	\begin{align*}
		\norm{\Sigma_* - \hat{\Sigma}_*}
		\leq&
		\frac{2\alpha\gamma\zeta}{3\tau^{3/2}\sigma^2}
		\cdot
		\left(
		(d+5)^{5/2} + d (d+3)^{3/2}
		\right)
		\\
		&+
		\frac{\gamma}{\sigma^2}
		\cdot
		\left(
		\frac{2dL\alpha^2}{\sigma\tau} 
		+ 
		\frac{2\gamma \alpha^3 (d+3)^{3/2}}{3\sigma\tau^{3/2}} 
		+
		\frac{d\alpha^2}{\sigma} \left(\tau^{-1} - \zeta^{-1}  \right)
		\right)^{1/2},
	\end{align*}
	where $\Pi_{\cS}(\cdot)$ is the projection operator which projects a symmetric matrix on to $\cS$ with Frobenius norm as distance measure.
\end{theorem}

\begin{remark}
	Theorem~\ref{prop:Sigma} shows that when $\alpha$ is small, then $\HSig_*$  (the $\Sigma$ minimizer of problem~\eqref{eq:modi_prob}) is close to $\Pi_{\cS}\left(\left(\nabla^2f(\mu^*)\right)^{-1}\right)$. If $\tau \leq \sigma$ and $\zeta\geq L$, then $\HSig_*$ is close to the inverse of the Hessian at $\mu^*$.  
\end{remark}

The above propositions show that if $\alpha$ is small, then $\left(\Hu^*, \HSig_*\right)$, which is the minimizer of problem~\ref{eq:modi_prob}, will be close to $\left(\mu^*, \Pi_{\cS}\left(\left(\nabla^2f(\mu^*)\right)^{-1}\right)\right)$. 
In the next section, we propose a mirror natural evolution strategy to minimize problem~\eqref{eq:modi_prob}.

\section{Mirror Natural Evolution Strategies}
\label{sec:MiNES}

In the previous sections, we have shown that one can obtain the minimizer of $f(\mu)$ by minimizing its reparameterized function $Q_\alpha(\theta)$ with $\theta = (\mu, \Sigma)$. 
Instead of solving the optimization problem \eqref{eq:modi_prob} by the natural gradient descent, we propose a novel method called MIrror Natural Evolution Strategy (\texttt{MiNES}) to minimize $Q_\alpha(\theta)$. 
\texttt{MiNES} consists of two main update procedures. 
It updates $\mu$ by natural gradient descent but with `antithetic sampling' (refers to Eqn.~\eqref{eq:g_mu}).
Moreover, \texttt{MiNES}  updates $\Sigma$ by the mirror descent method. 
The mirror descent of $\Sigma$ can be derived naturally because $\nabla_\Sigma R(\Sigma) = -\frac{\alpha^2}{2}\Sigma^{-1}$ is a mirror map  widely used in convex optimization \citep{kulis2009low}.    

In the rest of this section, we will first describe our algorithmic procedure in detail. Then we will discuss the connection between \texttt{MiNES} and existing works.

\subsection{Algorithm Description}

We will give the update rules of $\mu$ and $\Sigma$ respectively.

\paragraph{Natural Gradient Descent of $\mu$}

The natural gradient of $Q_\alpha(\theta)$ with respect to $\mu$ is defined as 
\begin{equation}
	\label{eq:ng_mines}
	g(\mu) = F_\mu^{-1}\frac{\partial Q_\alpha}{\partial \mu}.
\end{equation}
where $F_\mu$ is the Fisher information matrix with respect to $\mu$. 
First, by the properties of the Gaussian distribution, we have the following property.
\begin{lemma}
	Let $F_\mu$ be the Fisher information matrix with respect to $\mu$ and the natural gradient $g(\mu)$ be defined as Eqn.~\eqref{eq:ng_mines}. Then $g(\mu)$ satisfies 
	\begin{equation*}
		g(\mu) = 
		\alpha^2\cdot \left(\frac{1}{2\alpha}\cdot\EE_u\left[(f(\mu+\alpha \Sigma^{1/2}u) - f(\mu-\alpha \Sigma^{1/2}u)) \Sigma^{1/2}u\right]\right).
	\end{equation*}
\end{lemma}
\begin{proof}
	The Fisher matrix $F_\mu$ can be computed as follows \citep{wierstra2014natural}:
	\begin{align}
		F_\mu = \EE_z\left[\nabla_\mu \log \pi(z|\theta)\nabla_\mu \log \pi(z|\theta)^\top\right] 
		\overset{\eqref{eq:nab_mu}}{=}\EE_u\left[\alpha^{-2}\Sigma^{-1/2}uu^\top \Sigma^{-1/2}\right] 
		=\alpha^{-2}\Sigma^{-1}. \label{eq:F_mu}
	\end{align}
	Note that $\frac{\partial Q_\alpha(\theta)}{\partial \mu} = \frac{\partial J(\theta)}{\partial \mu}$, by Eqn.~\eqref{eq:log_lh}, so we have
	\begin{equation*}
		\frac{\partial Q_\alpha(\theta)}{\partial \mu} =\frac{\partial J(\theta)}{\partial \mu} = \EE[f(z)\nabla_\mu\log\pi(z|\theta)].
	\end{equation*}
	We also have $z = \mu + \alpha \Sigma^{1/2}u$ with $u\sim N(0, I_d)$, that is, $z\sim N(\mu, \alpha^2 \Sigma)$. We first consider $\mu$ part, by Eqn.~\eqref{eq:nab_mu} with $\bar{\Sigma} = \alpha^2\Sigma$, we have
	\begin{align*}
		\frac{\partial Q_\alpha(\theta)}{\partial \mu} =& \EE_z\left[f(z)\nabla_\mu\log\pi(z|\theta)\right]
		=\EE_u\left[f(\mu+\alpha \Sigma^{1/2}u)\left(\alpha^2\Sigma\right)^{-1}\cdot\alpha \Sigma^{1/2}u\right]\\
		=&\EE_u  \left[f(\mu+\alpha \Sigma^{1/2}u)\alpha^{-1}\Sigma^{-1/2} u\right].
	\end{align*}
	Because of the symmetry of Gaussian distribution, by setting $z = \mu - \alpha\Sigma^{1/2}u$, we can obtain
	\begin{align*}
		\frac{\partial Q_\alpha(\theta)}{\partial \mu} 
		=-\EE_u  \left[f(\mu-\alpha \Sigma^{1/2}u)\alpha^{-1}\Sigma^{-1/2} u\right].
	\end{align*}
	Combining above two equations, we can obtain that
	\begin{equation}
		\frac{\partial Q_\alpha(\theta)}{\partial \mu} = \frac{1}{2\alpha}\cdot\EE_u\left[(f(\mu+\alpha \Sigma^{1/2}u) - f(\mu-\alpha \Sigma^{1/2}u)) \Sigma^{-1/2}u\right].
	\end{equation}
	
	With the knowledge of $\partial Q_\alpha(\theta)/\partial \mu$ and $F_\mu$ in Eqn.~\eqref{eq:F_mu}, we can obtain the result.
\end{proof}

With the natural gradient $g(\mu)$ at hand, we can update $\mu$ by the natural gradient descent as follows:
\begin{align*}
	\mu_{k+1} = \mu_k - \eta_1'g(\mu)
	=\mu_k - \eta_1 \cdot \frac{1}{2\alpha}\EE_u\left[(f(\mu+\alpha \Sigma^{1/2}u) - f(\mu-\alpha \Sigma^{1/2}u)) \Sigma^{1/2}u\right]
\end{align*} 
where $\eta_1 = \alpha^2\eta_1'$ is the step size. Note that, during the above update procedure, we need to compute the expectations which is infeasible in real applications. 
Instead,  we sample a mini-batch of size $b$ to approximate $\frac{1}{2\alpha}\EE\left[(f(\mu+\alpha \Sigma^{1/2}u) - f(\mu-\alpha \Sigma^{1/2}u)) \Sigma^{1/2}u\right]$, and we define
\begin{equation}\label{eq:g_mu}
	\ti{g}(\mu_k) = \frac{1}{b}\sum_{i=1}^b \frac{f(\mu_k+ \alpha\Sigma_k^{1/2} u_i) -f(\mu_k - \alpha\Sigma_k^{1/2}u_i)}{2\alpha} \Sigma_k^{1/2} u_i\quad \mbox{with}\quad  u_i\sim N(0, I_d).
\end{equation}
Using $\ti{g}(\mu_k)$, we update $\mu$ as follows:
\begin{equation*}
	\mu_{k+1} = \mu_k - \eta_1 \ti{g}(\mu_k) .
\end{equation*}

\paragraph{Mirror Descent of $\Sigma$}

Recall from the definition of $Q_\alpha$, we have $Q_\alpha(\theta) = J(\theta) + R(\Sigma)$.
With the regularizer $R(\Sigma)$, we can define the Bregman divergence with respect to $R(\Sigma)$ as
\begin{align*}
	B_R(\Sigma_1, \Sigma_2) =& R(\Sigma_1) - R(\Sigma_2) - \dotprod{\nabla_\Sigma R(\Sigma_2), \Sigma_1 - \Sigma_2}\\
	=& -\frac{\alpha^2}{2}\left(\log\det(\Sigma_1\Sigma_2^{-1})
	-\dotprod{\Sigma_2^{-1}, \Sigma_1} + d\right) .
\end{align*}
The update rule of $\Sigma$ employing mirror descent is defined as 
\begin{equation}\label{eq:mirror_prox}
	\Sigma_{k+1} = \argmin_{\Sigma} \eta_2 \dotprod{\frac{\partial Q_\alpha(\theta)}{\partial \Sigma_k}, \Sigma} + B_R(\Sigma, \Sigma_k) .
\end{equation}
Using $\nabla_\Sigma R(\Sigma)$ as the mapping function, the update rule of above equation can be reduced to 
\begin{align}\label{eq:mirror}
	\nabla_\Sigma R(\Sigma_{k+1}) = \nabla_\Sigma R(\Sigma_{k}) - \eta_2 \nabla_\Sigma Q_\alpha(\theta_k).
\end{align}
In the following lemma, we will compute $\frac{\partial Q_\alpha(\theta)}{\partial \Sigma}$. 
\begin{lemma}
	Let $Q_\alpha(\theta)$ be defined in Eqn.~\eqref{eq:Q}. Then it holds that
	\begin{equation}
		\small
		\label{eq:Q_Sig}
		\frac{\partial Q_\alpha(\theta) }{\partial \Sigma}= \frac{1}{4}\cdot\EE_u\left[\left(f(\mu - \alpha \Sigma^{1/2}u) + f(\mu + \alpha \Sigma^{1/2}u) - 2f(\mu) \right)\left(\Sigma^{-1/2}uu^\top \Sigma^{-1/2} -  \Sigma^{-1}\right)\right] - \frac{\alpha^2}{2}\Sigma^{-1}.
	\end{equation}
\end{lemma}
\begin{proof}
	Note that $\partial Q_\alpha(\theta)/\partial \Sigma = \partial J(\theta)/\partial \Sigma +  \partial R(\Sigma)/\partial \Sigma$.
	First, we will compute $\frac{\partial J(\theta)}{\partial \Sigma}$. 
	Let $z = \mu + \alpha \Sigma^{1/2}u$ with $u\sim N(0, I_d)$. 
	By Eqn.~\eqref{eq:nab_Sig}, we can obtain that
	\begin{align*}
		\frac{\partial J(\theta)}{\partial \Sigma} =&\frac{\partial J(\theta)}{\partial \bar{\Sigma}} \cdot \frac{\partial \bar{\Sigma}}{\partial \Sigma}
		\overset{\eqref{eq:nab_Sig}}{=} \EE_z\left[f(z) \left(\frac{1}{2}\Sigma^{-1}(z-\mu)(z-\mu)^\top\Sigma^{-1}\alpha^{-2} - \frac{1}{2} \Sigma^{-1}\alpha^{-2}\right)\right]\cdot\alpha^2\\
		=&\frac{1}{2}\cdot\EE_u\left[f(\mu+ \alpha \Sigma^{1/2}u) \left(\Sigma^{-1/2}uu^\top \Sigma^{-1/2} -  \Sigma^{-1}\right)\right].
	\end{align*}
	Because of the symmetry of Gaussian distribution, we can also have $z = \mu - \alpha \Sigma^{1/2} u$. Then, we can similarly derive that 
	\begin{align*}
		\frac{\partial J(\theta)}{\partial \Sigma}
		=\frac{1}{2}\cdot\EE_u\left[f(\mu- \alpha \Sigma^{1/2}u) \left(\Sigma^{-1/2}uu^\top \Sigma^{-1/2} -  \Sigma^{-1}\right)\right].
	\end{align*}
	Note that, we also have the following identity
	\begin{equation*}
		\EE_u\left[f(\mu) \left(\Sigma^{-1/2}uu^\top \Sigma^{-1/2} -  \Sigma^{-1}\right)\right] = 0.
	\end{equation*}
	Combining above equations, we can obtain that 
	\begin{equation}\label{eq:J_Sig}
		\frac{\partial J(\theta)}{\partial \Sigma} = \frac{1}{4}\cdot\EE_u\left[\left(f(\mu - \alpha \Sigma^{1/2}u) + f(\mu + \alpha \Sigma^{1/2}u) - 2f(\mu) \right)\left(\Sigma^{-1/2}uu^\top \Sigma^{-1/2} -  \Sigma^{-1}\right)\right].
	\end{equation}
	Furthermore, by the definition of $Q_\alpha$ in Eqn.~\eqref{eq:Q}, we have
	\begin{equation*}
		\frac{\partial Q_\alpha(\theta) }{\partial \Sigma}
		= \frac{\partial J(\theta)}{\partial \Sigma} - \frac{\alpha^2}{2}\Sigma^{-1} .
	\end{equation*}
	Therefore, we can obtain the result.
\end{proof}

Since, we also have
$
\nabla_\Sigma R(\Sigma) = -\frac{\alpha^2}{2}\Sigma^{-1}
$,
substituting $\nabla_{\Sigma} Q_\alpha(\theta)$ and $\nabla_\Sigma R(\Sigma)$ into Eqn.~\eqref{eq:mirror}, we have
\begin{equation}\label{eq:Sig_up}
	\begin{aligned}
		-\frac{\alpha^2}{2}\Sigma_{k+1}^{-1}=& -\frac{\alpha^2}{2}\Sigma_k^{-1}  - \eta_2 \nabla_{\Sigma} Q_\alpha(\theta_k) \\
		\Rightarrow \Sigma_{k+1}^{-1} = &\Sigma_k^{-1} + 2\eta_2\alpha^{-2} \frac{\partial Q_\alpha(\theta_k)}{\partial \Sigma}. 
	\end{aligned}
\end{equation}

Similar to the update of $\mu$, we only sample a small batch points to query their values and use them to estimate $\partial Q_\alpha(\theta)/\partial \Sigma$. 
We can construct the approximate gradient with respect to $\Sigma$ as follows: 
\begin{equation}\label{eq:TG}
	\TG(\Sigma_k) 
	= 
	\frac{1}{2b\alpha^2} \sum_{i=1}^{b}
	\left[\left(f(\mu_k - \alpha \Sigma_k^{1/2}u_i) + f(\mu_k + \alpha \Sigma_k^{1/2}u_i) - 2f(\mu_k) \right)\left(\Sigma_k^{-1/2}u_iu_i^\top \Sigma_k^{-1/2} -  \Sigma_k^{-1}\right)\right] 
	- \Sigma_k^{-1}.
\end{equation}
The following lemma shows that an important property of $\TG(\Sigma_k) $.
\begin{lemma}\label{prop:TG}
	Letting $\TG(\Sigma_k)$ be defined in Eqn.~\eqref{eq:TG}, then $\TG(\Sigma_k)$ is an unbiased estimation of $2 \frac{\partial Q_\alpha(\theta)}{\alpha^2\partial \Sigma}$ at $\Sigma_k$.
\end{lemma}
\begin{proof}
	By Eqn.~\eqref{eq:J_Sig}, we can observe that 
	\begin{equation}\label{eq:bG}
		\BG(\Sigma_k)=\frac{1}{2b\alpha^2} \sum_{i=1}^{b}
		\left[\left(f(\mu_k - \alpha \Sigma_k^{1/2}u_i) + f(\mu_k + \alpha \Sigma_k^{1/2}u_i) - 2f(\mu_k) \right)\left(\Sigma_k^{-1/2}u_iu_i^\top \Sigma_k^{-1/2}\Sigma_k^{-1/2} -  \Sigma_k^{-1}\right)\right] 
	\end{equation}
	is an unbiased estimation of $2\alpha^{-2}\partial J(\theta)/\partial \Sigma$ at $\Sigma_k$. Furthermore, we have
	\begin{equation*}
		\nabla_\Sigma R(\Sigma) = -\frac{\alpha^2}{2}\Sigma^{-1}.
	\end{equation*}
	Therefore, we can conclude that $\TG(\Sigma_k)$ is an unbiased estimation of $2\alpha^{-2} \partial Q_\alpha(\theta)/\partial \Sigma$ at $\Sigma_k$.
\end{proof}
Replacing $2\alpha^{-2} \partial Q_\alpha(\theta_k)/\partial \Sigma$ with  $\TG(\Sigma_k)$ in Eqn.~\eqref{eq:Sig_up}, we update $\Sigma$ as follows
\begin{equation*}
	\Sigma_{k+1}^{-1} = \Sigma_k^{-1} + \eta_2 \TG(\Sigma_k).
\end{equation*}
where $\eta_2$ is the step size. 

\paragraph{Projection to The Constraint}
Because of the constraint that $\Sigma_{k+1}\in\cS$, we need to project $\Sigma_{k+1}$ back to $\cS$. 
Since we update $\Sigma^{-1}$ instead of directly updating $\Sigma$,  we define another convex set
\begin{equation}\label{eq:cS_p}
	\mathcal{S}'=\bigg\{\Sigma^{-1}\bigg| \Sigma\in\mathcal{S}\bigg\}
\end{equation}
It is easy to check that for any $\Sigma^{-1}\in\cS'$, then it holds that $\Sigma\in\cS$. 
Taking the extra projection to $\cS'$, 
we modify the update rule of $\Sigma$ as follows;
\begin{equation}\label{eq:Sig_modi}
	\left\{
	\begin{aligned}
		\Sigma_{k+0.5}^{-1} =& \Sigma_k^{-1} + \eta_2 \TG(\Sigma_k)\\
		\Sigma_{k+1}^{-1} =& \Pi_{\mathcal{S}'}\left(\Sigma_{k+0.5}^{-1}\right)
	\end{aligned}
	\right.
\end{equation}

The projection $\Pi_{\mathcal{S}'}(\Sigma^{-1})$ is conducted as follows. 
First, we conduct the spectral decomposition $\Sigma^{-1} = U\Lambda U^\top$, where $U$ is an orthonormal matrix and $\Lambda$ is a diagonal matrix with $\Lambda_{i,i} = \lambda_i$. 
Second, we truncate $\lambda_i$'s. If $\lambda_i > \zeta$, we set $\lambda_i = \zeta$. If $\lambda_i < \tau$, we set $\lambda_i = \tau$, that is
\begin{equation}\label{eq:proj_cs_prime}
	\Pi_{\cS'}\left(\Sigma^{-1}\right) = U\bar{\Lambda}U^\top, 
	\quad
	\mbox{with}
	\quad
	\bar{\Lambda}_{i,i} = \left\{
	\begin{aligned}
		&\tau^{-1}    \qquad \mbox{if}\;\lambda_i(\Sigma^{-1})>\tau^{-1}\\
		&\zeta^{-1}   \qquad \mbox{if}\;\lambda_i(\Sigma^{-1})<\zeta^{-1}\\
		&\lambda_i(\Sigma^{-1})    \qquad \mbox{otherwise}
	\end{aligned}
	\right.
\end{equation} 
where $\bar{\Lambda}$ is a diagonal matrix. It is easy to check the
correctness of Eqn.~\eqref{eq:proj_cs_prime}. For completeness, we
prove it in Proposition~\ref{prop:project} in the Appendix.

\begin{algorithm}[tb]
	\caption{Meta-Algorithm \texttt{MiNES}}
	\label{alg:zero_order}
	\begin{small}
		\begin{algorithmic}[1]
			\STATE {\bf Input:} $\mu_1$, $\Sigma_1^{-1}$, $\alpha$ and the target iteration number $K$.
			\FOR {$k=1,\dots, K$ }
			\STATE Compute $\ti{g}(\mu_k) = \frac{1}{b}\sum_{i=1}^b \frac{f(\mu_k+ \alpha\Sigma_k^{1/2} u_i) -f(\mu_k - \alpha\Sigma_k^{1/2}u_i)}{2\alpha} \Sigma_k^{1/2} u_i$ with $u_i\sim N(0, I_d)$
			\STATE Compute $\TG(\Sigma_k) = \frac{1}{2b\alpha^2} \sum_{i=1}^{b}\left[\left(f(\mu_k - \alpha \Sigma_k^{1/2}u_i) + f(\mu_k + \alpha \Sigma_k^{1/2}u_i) - 2f(\mu_k) \right)\left(\Sigma_k^{-1/2}u_iu_i^\top \Sigma_k^{-1/2} -  \Sigma_k^{-1}\right)\right]-\Sigma_k^{-1}$
			\STATE Update $\mu_{k+1} = \mu_k - \eta_{1,k} \ti{g}(\mu_k)$  \label{step:update} 
			\STATE Update $\Sigma^{-1}_{k+1} = \Pi_{\cS'}\left(\Sigma^{-1}_k + \eta_{2,k}\TG(\Sigma_k)\right) $ \label{step:Sig_update}
			\ENDFOR
		\end{algorithmic}
	\end{small}
\end{algorithm}	

\paragraph{Algorithmic Summary of \texttt{MiNES}}

Now, we summarize the algorithmic procedure of \texttt{MiNES}. First, we update $\mu$ by  natural gradient descent and update $\Sigma$ by mirror descent as 
\begin{align*}
	\mu_{k+1} =& \mu_k - \eta_{1,k} \ti{g}(\mu_k)\\
	\Sigma^{-1}_{k+1} =& \Pi_{\cS'}\left(\Sigma^{-1}_k +
	\eta_{2,k}\TG(\Sigma_k)\right) ,
\end{align*} 
where $\ti{g}(\mu_k)$ and $\TG(\Sigma_k)$ are defined in Eqn.~\eqref{eq:g_mu} and~\eqref{eq:TG}, respectively. The detailed algorithm description is in Algorithm~\ref{alg:zero_order}. 

\subsection{Relation to Existing Work}

First, we compare \texttt{MiNES} to derivative free algorithms in the
optimization literature, which uses function value differences to estimate the gradient. In the work of \cite{Nesterov2017}, one approximates the gradient as follows
\begin{equation}
	g(\mu_k) = \frac{1}{b}\sum_{i=1}^b \frac{f(\mu_k+ \alpha u_i) -f(\mu_k
		- \alpha u_i)}{2\alpha}  u_i \quad\mbox{with} \quad u_i\sim N(0,
	I_d) ,
\end{equation}
and update  $\mu$ as 
\begin{equation}
	\mu_{k+1} = \mu_k - \eta_1 g(\mu_k) .
\end{equation}
Comparing $\ti{g}(\mu)$ to $g(\mu)$, we can observe that the
difference lies on the estimated covariance matrix. 
By utilizing $\Sigma_k$ to approximate the inverse of the Hessian , $\ti{g}(\mu_k)$  is an estimation of the natural gradient. 
In contrast, $g(\mu_k)$ just uses the identity matrix hence it only estimates the gradient of $f(\mu_k)$. 
Note that if we don't track the Hessian information by updating
$\Sigma_k$, and set $\Sigma_k$ to the identity matrix, then
\texttt{MiNES} will reduce to the derivative free algorithm of \cite{Nesterov2017}. This establishes a connection between \texttt{NES} and derivative free algorithms.

Then we compare \texttt{MiNES} with classical derivative-free algorithm \citep{conn2009global}. 
The algorithm proposed by \citet{conn2009global} is also a kind of second order method.
Unlike \texttt{MiNES}, the gradient and Hessian of function $f(\cdot)$ are computed approximately by regression method which takes $O(d^2)$ queries to the function value.
Thus, this kind of algorithms is much different from \texttt{MiNES} and other \texttt{NES}-type algorithms.

We may also compare \texttt{MiNES} to traditional \texttt{NES} algorithms (including \texttt{CMA-ES} since \texttt{CMA-ES} can be derived from \texttt{NES} \citep{AkimotoNOK10}). 
There are  two differences between \texttt{MiNES} and the conventional \texttt{NES} algorithms.
First, \texttt{MiNES} minimizes $Q_\alpha$, while \texttt{NES} minimizes $J(\theta)$ defined in Eqn.~\eqref{eq:J_org}. 
Second, the update rule of $\Sigma$ is different. 
\texttt{MiNES} uses the mirror descent to update $\Sigma^{-1}$.
In comparison, \texttt{NES} uses the natural gradient to update $\Sigma$.

\section{Convergence Analysis}
\label{sec:conv}

In this section we analyze the convergence properties of \texttt{MiNES}. 
\texttt{MiNES} has a two-stage convergence convergence properties just as the ones of classical Newton methods \citep{boyd2004convex}.
In the first stage, \texttt{MiNES} we set a constant step size and we have the following convergence rate.

\begin{lemma}\label{lem:gd_dec_1}
	Let the objective function $f(\cdot)$ be $L$-smooth and $\sigma$-strongly convex. 
	Given $0<\delta <1$, define that
	\begin{equation}\label{eq:c1} 
		c_1 = \left(\sqrt{d} + \sqrt{b} + \sqrt{2\log(2/\delta)}\right)^2, \quad  c_2 = b - 2\sqrt{b\log(1/\delta)}, \quad
		c_3 = 2d + 3\log (1/\delta).
	\end{equation}
	Setting the step size $\eta_{1,k} = \frac{b\tau}{4L c_1 }$ in Algorithm~\ref{alg:zero_order}.
	Then with a probability at least $1-\delta$ , it holds that
	\begin{equation}\label{eq:dec_0}
		f(\mu_{k+1}) - f(\mu^*) 
		\le
		\left(1 -  \frac{c_2\tau\sigma  }{16c_1L\zeta  }\right) \cdot \left(f(\mu_k) - f(\mu^*)\right)+ \Delta_{\alpha,1},
	\end{equation}
	with $\Delta_{\alpha,1}$  defined as 
	\begin{align*}
		\Delta_{\alpha,1}	= \frac{c_3^3 \gamma b  \alpha^4}{2^5\cdot 3^2\cdot c_1 L\zeta\tau^3}\left(1 + \frac{c_1c_3}{2\tau\zeta}\right).
	\end{align*}
\end{lemma}

After several iterations and entering the local region, then $\Sigma_k^{-1}$ can be helpful to improve the convergence rate of \texttt{MiNES}. 
Accordingly, we have the following convergence properties.

\begin{lemma}
	\label{lem:dec_local}
	Assume that $f(\cdot)$ has the properties in Lemma~\ref{lem:gd_dec_1} and also admits $\gamma$-Lipschitz Hessian. 
	Let $ \xi_k \cdot \Sigma_k^{-1} \preceq \nabla^2f(\mu_k) \preceq \mL_k \cdot \Sigma_k^{-1} $ and $f(\mu_k)$ satisfy the following property
	\begin{align}
		f(\mu_k) - f(\mu^*) 
		\le 
		\min\left\{ \frac{2^3\cdot 3^2 \cdot \mL_k^4 \tau^4}{L\gamma^2 }, \frac{\xi_k^2\sigma^3}{8\gamma^2 (L\tau^{-1} + 2\xi_k)^2}\right\}. \label{eq:cond}
	\end{align}  
	Given $0<\delta<1$, by setting the step size $\eta_{1,k} = \frac{b}{4\mL_k c_1}$,  then with a probability at least $1-\delta$, Algorithm~\ref{alg:zero_order} has the following property
	\begin{equation}\label{eq:dec}
		\begin{aligned}
			f(\mu_{k+1}) - f(\mu^*) 
			\le 
			\left( 1 - \frac{\xi_k c_2}{8\mL_k c_1}  \right) \cdot \Big( f(\mu_k) - f(\mu^*)\Big) + \Delta_{\alpha,2}
		\end{aligned}
	\end{equation} 
	where $c_1$, $c_3$, $c_2$ is defined in Eqn.~\eqref{eq:c1} and $\Delta_{\alpha,2}$ is defined as 
	\begin{align*}
		\Delta_{\alpha,2} = \frac{b^{3/2}\zeta^3\gamma^4  c_3^6 \alpha^6 }{2^8\cdot 3^4\cdot  \sigma^3 c_1^3 \tau^6} + \frac{b\zeta\gamma^2 c_3^4 \alpha^4}{2^5 \cdot 3^2 \cdot \sigma\tau^3 c_1^2} + \frac{b\gamma c_3^3 \alpha^4}{2^5\cdot 3^2 \cdot \tau^3 c_1}.
	\end{align*}
\end{lemma}

Lemma~\ref{lem:dec_local} shows that when $\Sigma_{k}^{-1}$ is a good preconditioner for $\nabla^2 f(\mu_k)$, that is, $\mL_k / \xi_k$ is of small value, then \texttt{MiNES} can achieve a fast convergence rate.
For example, if $\mL_k / \xi_k = 2$, Eqn.~\eqref{eq:dec} shows that  \texttt{MiNES} converges with a rate independent of the condition number $L/\sigma$.
In this case, \texttt{MiNES} achieves a much faster converge rate than the vanilla zeroth-order algorithm in \citep{Nesterov2017}.

Next, we will prove the how $\Sigma_{k}^{-1}$ converge to be a good preconditioner for $\nabla^2 f(\mu_k)$ and provide an explicit convergence rate.
\begin{lemma}
	\label{lem:Sig_dec}
	Given the target iteration number $K$ and $0<\delta<1$, we denote that 
	\begin{align*}
		&C_1 :=  \frac{A_K L (\zeta A_K + d \zeta) }{\tau}\cdot \sqrt{\frac{2}{b}\log \frac{K^2}{\delta}}, \mbox{ with } A_K = 8\log \frac{1}{b\delta} + 16 \log( K + 1) + d\\ 
		&C_2 := \sqrt{ \frac{32c_1L\zeta\gamma^2}{c_2\tau\sigma^2 } \cdot \Big( \big(f(\mu_1) - f(\mu^*)\big) +  (K-1) \max\left\{ \Delta_{\alpha,1},  \Delta_{\alpha,2}\right\}\Big) },\\
		&C_3 := \frac{c_3^{3/2}\cdot (c_3 + 1) \cdot \zeta\cdot \sqrt{K-1}}{4\tau^{3/2} d^{1/2} } \cdot \alpha,  \qquad
		C_4 := \frac{2L^2 c_3^2 \left( c_3\zeta  + \sqrt{d} \zeta\right)^2 }{b\tau^2} + \frac{d \zeta^2}{2b},
	\end{align*}
	with $c_1$, $c_3$, $c_2$ are defined in Eqn.~\eqref{eq:c1}, and 
	\begin{align}
		C = \max\left\{\frac{9(C_1 + C_2 + C_3)^2}{4} + 3 C_4, \; 2 C_4 + \frac{L^2}{b}, \; 	\norm{\Sigma_1^{-1}-H^*}^2 \right\}. \label{eq:C}
	\end{align}
	Letting the sequence $\{\Sigma_k\}_{k=1}^K$ generated by Algorithm~\ref{alg:zero_order} by setting $\eta_{2,k} = 1/k$, then with a probability at least $1-\delta$, it holds  that
	\begin{align*}
		\norm{\Sigma_k^{-1}-\Pi_{\cS'}(\nabla^2 f(\mu^*))}^2 \le \frac{C}{k}.
	\end{align*}
\end{lemma}

Above lemma shows that $\Sigma_k^{-1}$ will converge to the $\nabla^2 f(\mu^*)$ if $\zeta\ge L$ and $\tau \le \sigma$ with a rate $\cO(\frac{\log k}{k})$. 
This rate is the same to the one of the stochastic gradient descent on the strongly convex function shown in \citep{rakhlin2012making}.
To understand the result in Lemma~\ref{lem:Sig_dec}, we only need to consider the quadratic case. 
Let us consider  the following the problem:
\begin{align}
	\min_{\Sigma^{-1} \in \cS'} h(\Sigma^{-1}) \triangleq \norm{ \Sigma^{-1} - \nabla^2 f(\mu) }^2. \label{eq:h_Sig}
\end{align}
It is easy to check that 
\begin{align*}
	\nabla_{\Sigma^{-1}} h(\Sigma^{-1}) = 2 \left(\Sigma^{-1} - \nabla^2 f(\mu)\right) = -2 \cdot \EE\left[ \TG(\Sigma) \right],
\end{align*}
where the last equality is because of Lemma~\ref{lem:exp_tg} and the definition of $\TG$.
Thus, we can observe that the update rule of $\Sigma_k$ in Step~\ref{step:Sig_update} of Algorithm~\ref{alg:zero_order} is trying to solve the  problem~\eqref{eq:h_Sig} with the stochastic gradient descent method.

Next, we will show that the rate of $\widetilde{\cO}(1/k)$ for the covariance matrix is almost tight by proving that $\EE\left[\norm{\TG(\Sigma)}^2\right]$ will not converge to the zero. 
Without loss of generality, we only consider the case $d = 1$ and $b = 1$. We have
\begin{align*}
	\EE\left[ \norm{\TG(\Sigma)}^2  \right] 
	=& \EE\left[ \norm{ 	\frac{1}{2}\left(u^\top \Sigma^{1/2}H\Sigma^{1/2} u \cdot \left(\Sigma^{-1/2}uu^\top \Sigma^{-1/2} -  \Sigma^{-1}\right)\right) -\Sigma^{-1} }^2\right]\\
	=&
	\EE\left[ \norm{ 	\frac{1}{2}\left(u^\top \Sigma^{1/2}H\Sigma^{1/2} u \cdot \left(\Sigma^{-1/2}uu^\top \Sigma^{-1/2} -  \Sigma^{-1}\right)\right)  }^2\right] - 2 \dotprod{\Sigma^{-1}, H} + \norm{\Sigma^{-1}}^2\\
	=&\frac{1}{4} \EE\left[ (u^8 + u^4 - 2 u^6 )H^2 \right] - 2 \dotprod{\Sigma^{-1}, H} + \norm{\Sigma^{-1}}^2 \\
	=& \frac{39}{2} H^2 -  2 \Sigma^{-1}H + \norm{\Sigma^{-1}}^2 \\
	\ge&
	\frac{37}{2} H^2>0,
\end{align*}
where the forth equality is because of the moments of standard Gaussian distribution.
Since the variance of $ \TG(\Sigma) $ is always positive constant, the stochastic gradient descent wit $\TG$ can not achieve a linear convergence rate. 
Thus, Lemma~\ref{lem:Sig_dec} can only obtain a rate the same to  the one of the stochastic gradient descent on the strongly convex function shown in \citep{rakhlin2012making} and this rate is tight.

Combining above results, we can obtain the final convergence properties of \texttt{MiNES}.
\begin{theorem}\label{thm:main_cvg}
	Let the objective function $f(\cdot)$ be $L$-smooth and $\sigma$-strongly convex and admit  $\gamma$-Lipschitz Hessians. 
	Given $0<\delta < 1$, $c_1$, $c_3$, $c_3$ are defined in Eqn.~\eqref{eq:c1} and $C$ is defined in Lemma~\ref{lem:Sig_dec}. 
	Setting $\eta_{2,k} = 1/k$ in  Algorithm~\ref{alg:zero_order},  the sequence $\{\mu_k\}$ generated by Algorithm~\ref{alg:zero_order} has the following properties
	\begin{enumerate}
		\item If $f(\mu_k)$ does not satisfy condition \eqref{eq:cond}, then by setting the step size $\eta_{1,k} = \frac{b\tau}{4L c_1 }$, then Eqn.~\eqref{eq:dec_0} holds with a probability at least $1-\delta$.
		\item If Condition~\eqref{eq:cond} is satisfied, then by setting the step size $\eta_{1,k} = \frac{b}{4\mL_k c_1}$, with a probability at least $1-\delta$,  Eqn.~\eqref{eq:dec} holds with 
		\begin{equation}\label{eq:lxi}
			\small
			\begin{aligned}
				\mL_k =& \min\left\{ \frac{L}{\tau},  1 + \tau^{-1} \cdot \left(\sqrt{\frac{C}{k}} + \gamma \sqrt{ \frac{2\big(f(\mu_k) - f(\mu^*)\big)}{\sigma}} + \norm{ \nabla^2 f(\mu_k)  - \Pi_{\cS'}\left(\nabla^2 f(\mu_k)\right)}\right) \right\} \\
				\xi_k =& \max\left\{ \frac{\sigma}{\zeta},  1 - \tau^{-1} \cdot \left(\sqrt{\frac{C}{k}} + \gamma \sqrt{ \frac{2\big(f(\mu_k) - f(\mu^*)\big)}{\sigma}}+\norm{ \nabla^2 f(\mu_k)  - \Pi_{\cS'}\left(\nabla^2 f(\mu_k)\right)}\right)\right\}.
			\end{aligned}
		\end{equation}
	\end{enumerate}   
\end{theorem}

\begin{remark}
	Theorem~\ref{thm:main_cvg} shows that \texttt{MiNES} converges with a slow linear convergence rate before entering the local region defined in Eqn.~\eqref{eq:cond}.
	After entering the local region, $\Sigma_k^{-1}$ as the preconditioner begins to improve the convergence rate of \texttt{MiNES}. Note that Eqn.~\eqref{eq:lxi} implies that the convergence rate in the local region is no slower than the one before entering the local region. 
	As the iteration goes, the terms $\sqrt{C/k}$ and $ \sqrt{2\big(f(\mu_k) - f(\mu^*)\big) /\sigma} $ will converge to zero. 
	Furthermore, without loss of generality, we assume that $ \zeta \ge L $ and $\tau \le \sigma$ which implies that $\norm{ \nabla^2 f(\mu_k)  - \Pi_{\cS'}\left(\nabla^2 f(\mu_k)\right)} = 0$.
	Then, Theorem~\ref{thm:main_cvg} shows that  \texttt{MiNES} will converge faster and faster gradually until achieve a linear rate independent of the condition number of the objective. 
	Though, the convergence rate of \texttt{MiNES} will increase as iteration goes after entering the local region. 
	However,  \texttt{MiNES} can not achieve the superlinear convergence rate eventually because the final convergence rate of \texttt{MiNES} is a linear rate $ 1 - \frac{c_2}{8c_1} $ which can be obtain by Eqn.~\eqref{eq:dec} and~\eqref{eq:lxi}.
	
\end{remark}
\begin{remark}
	By the definition of $c_1$ and $c_2$ in Eqn.~\eqref{eq:c1}, the convergence rate of \texttt{MiNES} depend on $\cO(b/ d)$. For example, before entering the local region, Eqn.~\eqref{eq:dec_0} shows that \texttt{MiNES} converges with a rate $1 - \cO\left( \frac{b\tau\sigma}{dL\zeta} \right)$. Thus,, we can effectively improve the convergence rate by increasing the batch size.
	Similarly, by the definition of $C$ in Eqn.~\eqref{eq:C} and Eqn.~\eqref{eq:lxi}, the convergence rate of the covariance matrix can also be improved by increasing the batch size. 
\end{remark}

\section{Conclusion}
\label{sec:conclusion}

In this paper,   we proposed a novel reparameterized objective function $Q_\alpha(\theta)$. Accordingly, we proposed a new algorithm called \texttt{MiNES} to minimize $Q_\alpha(\theta)$. 
We showed that the covariance matrix of \texttt{MiNES} converges to the Hessian inverse with a rate $\widetilde{\cO}(1/k)$, and we presented a rigorous convergence analysis of \texttt{MiNES}.
To the best of our knowledge, we provided the first explicit convergence rates of a zeroth-order algorithm with approximate Hessian being approximated by the zeroth-order queries. Furthermore, \texttt{MiNES} can be viewed as an extension of the traditional first order derivative free algorithms in the optimization literature and a special kind of \texttt{NES} algorithm.
This clarifies the connection between \texttt{NES} algorithm and derivative free methods.

Finally, because the minimizer $(\mu^*, \Sigma^*)$ of $Q_\alpha(\theta)$ are close to the minimizer and the Hessian inverse of the objective function $f(\cdot)$ but with some perturbation depending on $\alpha$, we believe our work can help design better evolutionary strategies algorithms to minimizing $Q_\alpha(\theta)$. Our work can also help  better understand the convergence properties of existent algorithms.





\newpage

\appendix
\section{Convexity of $\cS$ and $\cS'$}

First, we will show that $\cS$ and $\cS'$ are convex.
\begin{proposition}
	The sets $\cS$ and $\cS'$ defined in Eqn.~\eqref{eq:cS} and~\eqref{eq:cS_p} are convex.
\end{proposition}
\begin{proof}
	Letting $\Sigma_1$ and $\Sigma_2$ belong to $\cS$ and given $0 \le \beta \le 1$, then we have
	\begin{equation*}
\beta \Sigma_1 + (1-\beta) \Sigma_2
		\preceq
		\beta \tau^{-1} I + (1-\beta)\tau^{-1} I
		=\tau^{-1} I.
	\end{equation*}
	Similarly, we have $	\beta \Sigma_1 + (1-\beta) \Sigma_2
	\succeq
	\zeta^{-1} I$. 
	Therefore, $\cS$ is a convex set. 
	The convexity of $\cS'$ can be proved similarly.
\end{proof}

\begin{proposition}\label{prop:project}
	Let $A$ be a symmetric matrix. $A = U\Lambda U^\top$ is the spectral decomposition of $A$. The diagonal matrix $\bar{\Lambda}$ is defined as 
	\begin{equation*}
		\bar{\Lambda}_{i,i} = \left\{
		\begin{aligned}
			&\tau^{-1}    \qquad \mbox{if}\;\Lambda_{i,i}>\tau^{-1}\\
			&\zeta^{-1}  \qquad \mbox{if}\;\Lambda_{i,i}<\zeta^{-1}\\
			&\Lambda_{i,i}    \qquad \mbox{otherwise}
		\end{aligned}
		\right.
	\end{equation*}
	Let $\Pi_{\cS}(A)$ be the projection of symmetric $A$ on to $\cS$ defined in Eqn.~\eqref{eq:cS}, that is, $\Pi_{\cS}(A) = \argmin_{X\in\cS}\norm{A-x}$ with $\norm{\cdot}$ being Frobenius norm, then we have
	\begin{equation*}
		\Pi_{\mathcal{S}'}(A) = U\bar{\Lambda}U^\top.
	\end{equation*}
\end{proposition}
\begin{proof}
	We have the following Lagrangian \citep{lanckriet2004learning}
	\begin{equation*}
		L(X, A_1, A_2) = \norm{A-X}^2 + 2\dotprod{A_1, X - \tau^{-1} I} + 2\dotprod{A_2, \zeta^{-1} I - X},
	\end{equation*}
	where $A_1$ and $A_2$ are two positive semi-definite matrices. 
	The partial derivative $\partial L(X, A_1, A_2)/\partial{X}$ is 
	\begin{equation*}
		\frac{\partial L(X, A_1, A_2)}{\partial{X}} 
		= 
		2(X-A+A_1 - A_2).
	\end{equation*}
	By the general Karush-Kuhn-Tucker (KKT) condition \citep{lanckriet2004learning}, we have
	\begin{align*}
		&X_* = A-A_1+A_2 \\
		&A_1X_* = \tau^{-1} A_1, \quad A_2X_* = \zeta^{-1} A_2\\
		& A_1\succeq 0,\quad A_2\succeq 0.
	\end{align*}
	
	Since the optimization problem is strictly convex, there is a unique solution $(X_*,A_1,A_2)$ that satisfy the above KKT condition.
	Let $A = U \Lambda U^\top$ be the spectral decomposition of $A$. We construct $A_1$ and $A_2$ as follows:
	\begin{align}
		A_1 =& U\Lambda^{(1)}U^\top \quad\mbox{with}\quad \Lambda^{(1)}_{i,i} = \max\{\Lambda_{i,i} - \tau^{-1}, 0\} \label{eq:A1}\\ 
		A_2 =& U\Lambda^{(2)}U^\top \quad\mbox{with}\quad \Lambda^{(2)}_{i,i} = \max\{\zeta^{-1} - \Lambda_{i,i}, 0\}. \label{eq:A2}
	\end{align}
	$X_*$ is defined as $X_* = U\bar{\Lambda} U^\top$. 
	We can check that $A-A_1 + A_2 = U(\Lambda - \Lambda^{(1)} + \Lambda^{(2)}) U^\top = U\bar{\Lambda} U^\top  = X_*$. 
	The construction of $A_1$ and $A_2$ in Eqn.~\eqref{eq:A1},~\eqref{eq:A2} guarantees these two matrix are positive semi-definite.
	Furthermore, we can check that $A_1$ and $A_2$  satisfy $A_1 X_* = \tau^{-1} A_1$ and $A_2X_* =\zeta^{-1} A_2$. 
	Thus, $A_1$, $A_2$ and $X_*$ satisfy the KKT's condition which implies $X_* = U\bar{\Lambda} U^\top$ is the projection of $A$ onto $\cS$. 
\end{proof}

\section{Some Useful Lemmas }

\begin{lemma}[\cite{Nesterov2017}]\label{lem:gauss_power}
	Let $p\geq 2$, $u$ be from $N(0, I_d)$, then we have the following bound
	\begin{align}
		d^{p/2} \leq \EE_u\left[\norm{u}^p\right]\leq (p+d)^{p/2}. \label{eq:gauss_power}
	\end{align}
\end{lemma}

\begin{lemma}[$\chi^2$ tail bound~\cite{Laurent2000Adaptive}]
	\label{lem:chi}
	Let $q_1, \dots, q_n$ be independent $\chi^2$ random variables, each with one degree of freedom. For any vector $\gamma = (\gamma_1, \dots , \gamma_n) \in \RR_+^n$ with non-negative
	entries, and any $t > 0$,
	\begin{align*}
		\PP\left[\sum_{i=1}^{n}\gamma_iq_i \geq \norm{\gamma}_1 + 2 \sqrt{\norm{\gamma}_2^2 t} + 2\norm{\gamma}_\infty t\right] \leq& \exp(-t),\\
		\PP\left[ \sum_{i=1}^{n}\gamma_iq_i \leq \norm{\gamma}_1 - 2 \sqrt{\norm{\gamma}_2^2 t}  \right] \le&  \exp(-t)
	\end{align*}
	where $\norm{\gamma}_1 = \sum_{i=1}^{n} |\gamma_i|$.
\end{lemma} 

\begin{lemma}[Corollary 5.35 of \cite{vershynin2010introduction}]
	Let $A$ be an $N \times n$ matrix whose entries are independent standard normal random variables. Then for every $\tau \ge 0$,
	with probability at least $1 - 2 \exp(-\tau^2/2)$, the largest singular value $s_{\max}(A)$ satisfies
	\begin{equation}
		\label{eq:smax}
		s_{\max}(A) \le \sqrt{N} + \sqrt{n} + \tau. 
	\end{equation}
\end{lemma}

\begin{lemma}
	Letting $u\sim N(0, I)$ be a $d$-dimensional Gaussian vector, then with probability at least $1-\delta$, it  holds that
	\begin{equation}
		\label{eq:u_norm}
		\norm{u}^2 \le 2d + 3\log(1/\delta).
	\end{equation}
\end{lemma}
\begin{proof}
	By Lemma~\ref{lem:chi}, we have
	\begin{equation*}
		\norm{u}^2 \le d + 2\sqrt{d\log(1/\delta)} + 2\log(1/\delta) \le 2d + 3\log(1/\delta).
	\end{equation*}
\end{proof}

\begin{lemma}[Hoeffding's inequality]
	\label{lem:hoeffding}
	For bounded random variables $X_i \in [a_i, b_i]$,  where $X_1, \dots, X_n$ are
	independent, then $S_n = \sum_{i=1}^n X_i$ satisfies that
	\begin{align*}
		\PP\left( S_n - \EE\left[ S_n \right]  \ge \psi \right) \le \exp\left( -\frac{-2\psi^2}{\sum_{i=1}^n (b_i - a_i)^2} \right).
	\end{align*}
\end{lemma}

\begin{lemma}[\cite{rakhlin2012making}]
	\label{lem:sgd_dec}
	Let a positive sequence $\{a_t\}_{t=1}^\infty$ satisfy that $a_t \le \frac{a_0}{t}$ for $t = 1,\dots, k$ and 
	\begin{align*}
		a_{k+1} 
		\le
		\frac{b}{k(k-1)} \sqrt{\sum_{t=2}^{k}(t-1)^2 \cdot a_t } + \frac{c}{k}, \qquad\mbox{with}\qquad b, c >0.
	\end{align*}
	Then if $a_0 \ge 9b^2/ 4 + 3c$, then it holds that $a_{k+1} \le a_0/(k+1)$.
\end{lemma}

\section{Proof of Section~\ref{sec:ROF}}
\subsection{Proof of Theorem~\ref{prop:quad_case}}

The proof of Theorem~\ref{prop:quad_case} is similar to that of Proposition~\ref{prop:project}.
\begin{proof}[Proof of Theorem~\ref{prop:quad_case}]
	By the definition of $Q_\alpha(\theta)$ and Eqn.~\eqref{eq:J_quad}, we have
	\begin{equation*}
		Q_\alpha(\theta) = f(\mu) + \frac{\alpha^2}{2}\dotprod{H, \Sigma} - \frac{\alpha^2}{2} \log\det\Sigma.
	\end{equation*}
	Then, taking partial derivative of $Q_\alpha$ with respect to $\mu$, we can obtain that 
	\[
	\frac{\partial Q_\alpha(\theta)}{\partial \mu} = \nabla_\mu f(\mu).
	\] 
	By setting $\frac{\partial Q_\alpha(\theta)}{\partial \mu}$ to zero, we can obtain that $Q_\alpha$ attains its minimum at $\mu^*$.
	
	For the $\Sigma$ part, we have the following Lagrangian \citep{lanckriet2004learning},
	\begin{equation*}
		L(\Sigma, A_1, A_2) 
		= 
		\frac{\alpha^2}{2}\dotprod{H, \Sigma} 
		- 
		\frac{\alpha^2}{2} \log\det\Sigma 
		+
		\frac{\alpha^2}{2}\dotprod{A_1, \Sigma - \tau^{-1}I}
		+
		\frac{\alpha^2}{2}\dotprod{A_2, \zeta^{-1}I - \Sigma},
	\end{equation*}  
	where $A_1$ and $A_2$ are two positive semi-definite matrices and $H$ denotes the Hessian matrix of the quadratic function $f(\cdot)$. 
	The $\partial L(\Sigma, A_1, A_2)/\partial{X}$ is
	\begin{equation*}
		\frac{\partial L(\Sigma, A_1, A_2)}{\partial{\Sigma}} 
		= 
		\frac{\alpha^2}{2}
		\left(
		H - \Sigma^{-1} + A_1 - A_2
		\right).
	\end{equation*}
	By the general Karush-Kuhn-Tucker (KKT) condition \citep{lanckriet2004learning}, we have
	\begin{align*}
		&\Sigma_*^{-1} = H+A_1-A_2 \\
		&A_1\Sigma_* = \tau^{-1}A_1, \quad A_2\Sigma_* = \zeta^{-1}A_2\\
		& A_1\succeq 0,\quad A_2\succeq 0.
	\end{align*}
	
	Since the optimization problem is strictly convex, there is a unique solution $(\Sigma_*,A_1,A_2)$ that satisfy the above KKT condition. We construct such a solution as follows. 
	Let $H = U \Lambda U^\top$ be the spectral decomposition of $H$, where $\Lambda$ is diagonal and $U$ is an orthogonal matrix. We define $\Sigma_*$ as $\Sigma_* = U \bar{\Lambda} U^\top$, where
	$\bar{\Lambda}$ is a diagonal matrix with $\bar{\Lambda}_{i,i} = \tau^{-1}$ if $\Lambda_{i,i} \leq \tau$, $\bar{\Lambda}_{i,i} = \zeta^{-1}$ if $\Lambda_{i,i} \geq \zeta$, and $\bar{\Lambda}_{i,i} = \Lambda_{i,i}^{-1}$ in other cases.
	$A_1$ and $A_2$ are defined as follows, where both $\Lambda^{(1)}_{i,i}$ and $\Lambda^{(2)}_{i,i}$ are diagonal matrices:
	\begin{align*}
		A_1 =& U\Lambda^{(1)}U^\top \quad\mbox{with}\quad \Lambda^{(1)}_{i,i} = \max\{\tau - \Lambda_{i,i}, 0\}  \\
		A_2 =& U\Lambda^{(2)}U^\top \quad\mbox{with}\quad \Lambda^{(2)}_{i,i} = \max\{\Lambda_{i,i} - \zeta, 0\}.
	\end{align*}
	
	Next, we will check that $A_1$, $A_2$ and $\Sigma_*$ satisfy the KKT's condition. First, we have
	\begin{align*}
		H+A_1-A_2 = U\Lambda U^\top + U\Lambda^{(1)}U^\top - U\Lambda^{(2)}U^\top = U \bar{\Lambda}^{-1} U^\top = \Sigma_*^{-1}.
	\end{align*}
	Then we also have $A_1\Sigma_* = U\Lambda^{(1)}U^\top U\bar{\Lambda}U^\top = U\Lambda^{(1)} \bar{\Lambda}U^\top = \tau^{-1}  U\Lambda^{(1)}U^\top = \tau^{-1} A_1$. 
	Similarly, it also holds that $A_2\Sigma_* = \zeta^{-1} A_2$.
	
	Finally, the construction of $A_1$ and $A_2$ guarantees these two matrix are positive semi-definite.
	
	Therefore, $\Sigma_*$ is the covariance part of the minimizer of $Q_\alpha(\theta)$, and $\left(\mu^*, \Pi_{\cS}\left(H^{-1}\right)\right)$ is the optimal solution of $Q_\alpha(\theta)$ under constraint $\cS$.
\end{proof}

\subsection{Proof of Theorem~\ref{prop:close}}
\begin{proof}[Proof of Theorem~\ref{prop:close}]
	By the Taylor's expansion of $f(z)$ at $\mu$, we have 
	\begin{equation}
		\left|
		f(z) -\left[ f(\mu) + \langle \nabla f(\mu),z-\mu \rangle + \frac12 (z-\mu)^\top \nabla^2 f(\mu) (z-\mu)
		\right] 
		\right| \leq \frac{\gamma}6 \| z-\mu\|_2^3 .
		\label{eq:taylor}
	\end{equation}
	By $z = \mu + \alpha \Sigma u$ with $u\sim N(0, I_d)$, we can upper bound the $J(\theta)$ as 
	\begin{align*}
		J(\theta) =& \frac{1}{(2\pi)^{d/2}}\int_u f(\mu + \alpha\Sigma^{1/2} u) \exp\left(-\frac{1}{2}\norm{u}^2\right) du \\
		\stackrel{\eqref{eq:taylor}}{\le}&\frac{1}{(2\pi)^{d/2}}\int_u \left(f(\mu) + \dotprod{\nabla f(\mu), \alpha \Sigma^{1/2}u} + \frac{\alpha^2}{2} u^\top \Sigma^{1/2} \nabla^2 f(\mu) \Sigma^{1/2} u + \frac{\gamma\alpha^3}{6}\|\Sigma^{1/2} u\|_2^3\right)\exp\left(-\frac{1}{2}\norm{u}^2\right) d u \\
		=&f(\mu) + \frac{\alpha^2}{2}\dotprod{\nabla^2f(\mu), \Sigma}+ \alpha^3\phi(\Sigma)\\
		\leq&f(\mu) + \frac{L\alpha^2}{2} \tr(\Sigma) + \alpha^3\phi(\Sigma),
	\end{align*}
	where the last inequality is because of $\|\nabla^2 f(\mu)\|_2 \leq L$, we have $|\tr(\nabla^2 f(\mu) \Sigma) | \leq L \tr(\Sigma)$ .
	
	By the fact that $Q_\alpha(\theta) = J(\theta) + R(\Sigma)$, we have
	\begin{equation*}
		Q_\alpha(\theta) \leq f(\mu) + \frac{L\alpha^2}{2} \tr(\Sigma) + \alpha^3\phi(\Sigma) + R(\Sigma).
	\end{equation*}
	
	Similarly, we can obtain that
	\begin{equation*}
		J(\theta) \geq f(\mu) - \frac{L\alpha^2}{2} \tr(\Sigma) -\alpha^3 \phi(\Sigma).
	\end{equation*}
	Therefore, we can obtain that 
	\begin{equation*}
		f(\mu) - \frac{L\alpha^2}{2} \tr(\Sigma) - \alpha^3\phi(\Sigma) + R(\Sigma)\leq Q_\alpha(\theta) .
	\end{equation*}
\end{proof}

\subsection{Proof of Theorem~\ref{prop:mu}}

\begin{proof}[Proof of Theorem~\ref{prop:mu}]
	Let $\HSig_*$ be $\Sigma$ part the minimizer $\hat{\theta}_*$ of $Q_\alpha$ under the  constraint $\Sigma\in\cS$.
	$\Sigma_*$ denotes the minimizer of $Q_\alpha$ given $\mu = \mu^* $ under the  constraint $\Sigma\in\cS$. Let us denote $\theta_* = (\mu^*, \Sigma_*)$.
	By Theorem~\ref{prop:close}, we have
	\begin{align*}
		f(\mu^*)+R(\Sigma_*)-\frac{L\alpha^2}{2}\tr(\Sigma_*)-\alpha^3\phi(\Sigma_*)\leq& Q_\alpha(\theta_*)   \leq   f(\mu^*)+\frac{L\alpha^2}{2}\tr(\Sigma_*)+\alpha^3\phi(\Sigma_*)+R(\Sigma_*)\\
		f(\hat{\mu}_*)
		+ R(\hat{\Sigma}_*)  
		- \frac{L\alpha^2}{2}\tr(\hat{\Sigma}_*)
		- \alpha^3\phi(\hat{\Sigma}_*)
		\leq&Q_\alpha(\hat{\theta}_*)
		\leq f(\hat{\mu}_*) 
		+ \frac{L\alpha^2}{2}\tr(\hat{\Sigma}_*)
		+ \alpha^3\phi(\hat{\Sigma}_*)
		+ R(\hat{\Sigma}_*)
	\end{align*}
	By the fact that $\hat{\theta}_*$ is the minimizer of $Q_\alpha(\theta)$ under constraint $\Sigma\in\cS$, then we have 
	$
	Q_\alpha(\hat{\theta}_*) \leq Q_\alpha(\theta_*).
	$
	Thus, we can obtain that 
	\begin{align*}
		&f(\Hu^*)+R(\HSig_*)-\frac{L\alpha^2}{2}\tr(\HSig_*)-\alpha^3\phi(\HSig_*) 
		\leq 
		f(\mu^*) 
		+ \frac{L\alpha^2}{2}\tr(\Sigma_*)
		+ \alpha^3\phi(\Sigma_*)
		+ R(\Sigma_*)\\
		\Rightarrow&
		f(\hat{\mu}_*)-f(\mu^*)
		\leq 	
		\frac{L\alpha^2}{2}\tr(\Sigma_* + \hat{\Sigma}_*) 
		+ 
		\alpha^3\left(\phi(\Sigma_*) + \phi(\hat{\Sigma}_*) \right)
		+
		R(\Sigma_*) - R(\hat{\Sigma}_*)
	\end{align*}
	Since $\mu^*$ is the solver of minimizing $f(\mu)$, we have
	$
	0
	\leq 
	f(\hat{\mu}_*) - f(\mu^*)
	$.
	Thus, we can obtain that 
	\begin{equation*}
		0
		\leq 
		f(\hat{\mu}_*) - f(\mu^*)   
		\leq
		\frac{L\alpha^2}{2}\tr(\Sigma_* + \hat{\Sigma}_*) 
		+ 
		\alpha^3\left(\phi(\Sigma_*) + \phi(\hat{\Sigma}_*) \right)
		+
		R(\Sigma_*) - R(\hat{\Sigma}_*). 
	\end{equation*}
	Next, we will bound the terms of right hand of above equation. First, we have
	\begin{equation}\label{eq:tr}
		\frac{L\alpha^2}{2}\tr(\Sigma_* + \hat{\Sigma}_*) 
		\leq
		\frac{dL\alpha^2}{2}(\lambda_{\max}(\Sigma_*)+\lambda_{\max}(\hat{\Sigma}_*) )
		\leq
		\frac{dL\alpha^2}{\tau} ,
	\end{equation}
	where the last inequality is because $\Sigma_*$ and $\hat{\Sigma}_*$ are in $\cS$. Then we bound the value of $\phi(\Sigma_*)$ as follows.
	\begin{align*}
		\phi(\Sigma_*) 
		\leq 
		\frac{\gamma\norm{\Sigma_*^{1/2}}^3}{6} 
		\left(\EE_u \left[\norm{u}^3\right]\right)
		\stackrel{\eqref{eq:gauss_power}}{\le}
		\frac{\gamma (d+3)^{3/2}}{6\tau^{3/2}},
	\end{align*}
	where the second inequality follows from Jensen's inequality. Similarly, we have
	\begin{equation*}
		\phi(\hat{\Sigma}_*) \leq  \frac{\gamma (d+3)^{3/2}}{6\tau^{3/2}}.
	\end{equation*}
	Thus, we obtain that 
	\begin{equation}\label{eq:phi}
		\alpha^3\left(\phi(\Sigma_*) + \phi(\hat{\Sigma}_*) \right)
		\leq
		\frac{\gamma \alpha^3 (d+3)^{3/2}}{3\tau^{3/2}}.
	\end{equation}
	Finally, we bound $R(\Sigma_*) - R(\hat{\Sigma}_*)$. 
	\begin{align*}
		R(\Sigma_*) - R(\hat{\Sigma}_*)
		=\frac{\alpha^2}{2} 
		\left(
		-\log\det\Sigma_* 
		+ 
		\log\det\hat{\Sigma}_* 
		\right)
		\leq
		\frac{d\alpha^2}{2}\left(\lambda_{\max}\left(\hat{\Sigma}_*\right) - \lambda_{\min}\left(\Sigma_*\right)\right)
		\leq
		\frac{d\alpha^2}{2} \left(\tau^{-1} - \zeta^{-1} \right).
	\end{align*}
	Thus, we have
	\begin{equation}\label{eq:R}
		R(\Sigma_*) - R(\hat{\Sigma}_*) 
		\leq 
		\frac{d\alpha^2}{2} \left(\tau^{-1} - \zeta^{-1}  \right)
	\end{equation}
	Combining Eqn.~\eqref{eq:tr},~\eqref{eq:phi} and~\eqref{eq:R}, we obtain that
	\begin{equation*}
		f(\hat{\mu}_*)  - f(\mu^*)
		\leq 
		\frac{dL\alpha^2}{\tau} 
		+ 
		\frac{\gamma \alpha^3 (d+3)^{3/2}}{3\tau^{3/2}} 
		+
		\frac{d\alpha^2}{2} \left(\tau^{-1} - \zeta^{-1}  \right).
	\end{equation*}
	By the property of strongly convex, we have
	\begin{align*}
		\norm{\mu^* - \hat{\mu}_*}^2 \leq \frac{2}{\sigma}\left( f(\mu^*) - f(\hat{\mu}_*)\right) 
		\leq 
		\frac{2dL\alpha^2}{\sigma\tau} 
		+ 
		\frac{2\gamma \alpha^3 (d+3)^{3/2}}{3\sigma\tau^{3/2}} 
		+
		\frac{d\alpha^2}{\sigma} \left(\tau^{-1} - \zeta^{-1}  \right).
	\end{align*}
\end{proof}

\subsection{Proof of Theorem~\ref{prop:Sigma}}

First, we give the following property.
\begin{lemma}\label{lem:exp_tg}
	Letting $u\sim N(0, I)$ and $H$ be a positive semi-definite matrix, then we have
	\begin{equation*}
		\frac{1}{2}\cdot\EE_u\left(u^\top \Sigma^{1/2}H\Sigma^{1/2} u \cdot \left(\Sigma^{-1/2}uu^\top \Sigma^{-1/2} -  \Sigma^{-1}\right)\right)
		= H.
	\end{equation*}
\end{lemma}
\begin{proof}
	Let $J(\theta)$ be defined as Eqn.~\eqref{eq:J_org}. $f(\cdot)$ is a quadratic function with $H$ as its Hessian matrix. Then $J(\theta)$ can be represented as Eqn.~\eqref{eq:J_quad}. Therefore, we have
	\begin{equation*}
		\frac{\partial J(\theta)}{\Sigma} 
		= 
		\frac{\alpha^2}{2} H.
	\end{equation*}
	Let $\TG$ be defined as Eqn.~\eqref{eq:TG} with respect to the quadratic function $f(\cdot)$. $\TG$ can further reduce to 
	\begin{equation*}
		\EE_u\left[\TG\right] = 
		\EE_u\left[ \frac{1}{2} \left(u^\top \Sigma^{1/2}H\Sigma^{1/2} u \cdot \left(\Sigma^{-1/2}uu^\top \Sigma^{-1/2} -  \Sigma^{-1}\right)\right)\right] 
		- 
		\frac{1}{2} \Sigma^{-1} .
	\end{equation*}
	By Lemma~\ref{prop:TG}, we can obtain that 
	\begin{equation*}
		\EE_u\left[ \frac{1}{2} \left(u^\top \Sigma^{1/2}H\Sigma^{1/2} u \cdot \left(\Sigma^{-1/2}uu^\top \Sigma^{-1/2} -  \Sigma^{-1}\right)\right)\right] 
		= 2\alpha^{-2} \cdot
		\frac{\partial J(\theta)}{\Sigma} 
		= H.
	\end{equation*}
	This completes the proof.
\end{proof}
\begin{proof}[Proof of Theorem~\ref{prop:Sigma}]
	We will compute $\HSig_*$. First, we have the following Lagrangian 
	\begin{equation*}
		L(\Sigma, A_1, A_2) 
		= 
		Q_\alpha(\theta) 
		+ 
		\frac{\alpha^2}{2}\dotprod{A_1,  \Sigma - \tau^{-1} I} 
		+ 
		\frac{\alpha^2}{2}\dotprod{A_2,  \zeta^{-1}I - \Sigma }, 
	\end{equation*}
	where $A_1$ and $A_2$ are two positive semi-definite matrices. 
	Next, we will compute $\partial(L)/\partial\Sigma$ 
	\begin{equation}\label{eq:L_partial}
		\partial(L)/\partial\Sigma 
		= 
		\frac{\partial J(\theta)}{\partial\Sigma}
		-\frac{\alpha^2}{2}\Sigma^{-1} 
		+\frac{\alpha^2}{2}A_1
		-\frac{\alpha^2}{2}A_2.
	\end{equation}
	
	Furthermore, by Eqn.~\eqref{eq:log_lh}, ~\eqref{eq:nab_Sig} and $z = \Hu^*+\alpha\Sigma^{1/2}u$, we have
	\begin{equation}
		\label{eq:J_Sig_0}
		\begin{aligned}
			\frac{\partial J(\theta)}{\partial \Sigma} =&\frac{\partial J(\theta)}{\partial \bar{\Sigma}} \cdot \frac{\partial \bar{\Sigma}}{\partial \Sigma} 
			\\
			\overset{\eqref{eq:nab_Sig}}{=}& \EE_z\left[f(z) \left(\frac{1}{2}\Sigma^{-1}(z-\Hu^*)(z-\Hu^*)^\top\Sigma^{-1}\alpha^{-2} - \frac{1}{2} \Sigma^{-1}\alpha^{-2}\right)\right]\cdot\alpha^2
			\\
			=&\frac{1}{2}\cdot\EE_u\left[f(\Hu^*+ \alpha \Sigma^{1/2}u) \left(\Sigma^{-1/2}uu^\top \Sigma^{-1/2} -  \Sigma^{-1}\right)\right] .
		\end{aligned}
	\end{equation}
	We can express $f(\Hu^*+\alpha\Sigma^{1/2}u)$ using Taylor expansion as follows:
	\begin{equation*}
		f(\Hu^*+\alpha\Sigma^{1/2}u) 
		= 
		f(\Hu^*) 
		+ 
		\dotprod{\nabla f(\Hu^*), \alpha\Sigma^{1/2}u} 
		+ 
		\frac{\alpha^2}{2}u^\top \Sigma^{1/2}\nabla^2f(\Hu^*)\Sigma^{1/2} u
		+ \tilde{\rho}\left(\alpha\Sigma^{1/2}u\right),
	\end{equation*}
	where $\tilde{\rho}\left(\alpha\Sigma^{1/2}u\right)$ satisfies that 
	\begin{equation}
		\label{eq:rho_t}
		|\tilde{\rho}\left(\alpha\Sigma^{1/2}u\right)| \leq \frac{\gamma\alpha^3\norm{\Sigma^{1/2}u}^3}{6},
	\end{equation}
	due to Eqn.~\eqref{eq:taylor}.
	By plugging the above Taylor expansion into Eqn.~\eqref{eq:J_Sig_0}, we obtain
	\begin{align*}
		\frac{\partial J(\theta)}{\partial \Sigma}
		=&
		\frac{1}{2}\cdot\EE_u\left[\left(\frac{\alpha^2}{2}u^\top \Sigma^{1/2}\nabla^2f(\Hu^*)\Sigma^{1/2} u
		+ \tilde{\rho}\left(\alpha\Sigma^{1/2}u\right)\right)
		\cdot
		\left(\Sigma^{-1/2}uu^\top \Sigma^{-1/2} -  \Sigma^{-1}\right)
		\right]
		\\
		=&\frac{\alpha^2}{2}\nabla^2 f(\Hu^*) 
		+ 
		\Phi(\Sigma),
	\end{align*}
	where the last equality uses Lemma~\ref{lem:exp_tg},
	and $\Phi(\Sigma)$ is defined as
	\begin{equation*}
		\Phi(\Sigma) 
		= 
		\frac12
		\cdot
		\EE_u
		\left[
		\tilde{\rho}\left(\alpha\Sigma^{1/2}u\right)
		\cdot
		\left(\Sigma^{-1/2}uu^\top \Sigma^{-1/2} -  \Sigma^{-1}\right)
		\right].
	\end{equation*}
	Replacing $\partial J(\theta)/\partial\Sigma$ to Eqn.~\eqref{eq:L_partial}, we have
	\begin{equation*}
		\frac{\partial L}{\partial \Sigma}
		=
		\frac{\alpha^2}{2}\nabla^2f(\Hu^*) 
		+
		\Phi(\Sigma) 
		-\frac{\alpha^2}{2}\Sigma^{-1} 
		+\frac{\alpha^2}{2}A_1
		-\frac{\alpha^2}{2}A_2.
	\end{equation*}
	By the KKT condition, we have
	\begin{align}
		&\HSig_* = \left(\nabla^2f(\Hu^*) + 2\alpha^{-2}\Phi(\HSig_*)  + A_1 - A_2\right)^{-1} \label{eq:Sig_opt}\\
		&A_1\HSig_* = \tau^{-1}A_1,\quad A_2\HSig_* = \zeta^{-1}A_2\notag\\
		&A_1\succeq 0,\quad A_2\succeq 0. \notag
	\end{align}
	
	Because the optimization problem is strictly convex, there is a unique solution $(\HSig_*,A_1,A_2)$ that satisfy the above KKT condition.
	Let $\nabla^2f(\Hu^*)  + 2\alpha^{-2}\Phi(\HSig_*) = U\Lambda U^\top$ be the spectral decomposition of $\nabla^2f(\Hu^*)+ 2\alpha^{-2}\Phi(\HSig_*)$, where $U$ is a orthonormal matrix and $\Lambda$ is a diagonal matrix,  then we construct $A_1$ and $A_2$ as follows
	\begin{align*}
		A_1 =& U\Lambda^{(1)}U^\top \quad\mbox{with}\quad \Lambda^{(1)}_{i,i} = \max\{\tau - \Lambda_{i,i}, 0\}\\
		A_2 =& U\Lambda^{(2)}U^\top \quad\mbox{with}\quad \Lambda^{(2)}_{i,i} = \max\{\Lambda_{i,i} - \zeta, 0\}.
	\end{align*}
	Substituting $A_1$ and $A_2$ in Eqn.~\eqref{eq:Sig_opt}, we can obtain that 
	\begin{equation}\label{eq:Sig_opt_1}
		\HSig_* 
		= 
		\Pi_{\cS}
		\left(
		\left(
		\nabla^2f(\mu^*) 
		+
		2\alpha^{-2}\Phi(\HSig_*)
		\right)^{-1}
		\right).
	\end{equation}
	Similar to the proof of Theorem~\ref{prop:project} and \ref{prop:mu}, we can check that $\HSig_*$, $A_1$ and $A_2$ satisfy the above KKT condition.
	
	Now we begin to bound the error between $\HSig_*$ and $\Pi_{\cS}\left(\left(\nabla^2f(\Hu^*) 
	\right)^{-1}\right)$. We have
	\begin{equation} \label{eq:tmp}
		\begin{aligned}
			&\norm{\HSig_* - \Pi_{\cS}\left(\left(\nabla^2f(\Hu^*) 	
				\right)^{-1}\right)} \\
			\overset{\eqref{eq:Sig_opt_1}}{=}&
			\norm{\Pi_{\cS}
				\left(
				\left(
				\nabla^2f(\Hu^*) 
				+
				2\alpha^{-2}\Phi(\HSig_*)
				\right)^{-1}
				\right) 
				- 
				\Pi_{\cS}
				\left(
				\left(
				\nabla^2f(\Hu^*) 	
				\right)^{-1}
				\right)} \\
			\leq&
			\norm
			{
				\left(
				\nabla^2f(\Hu^*) 
				+
				2\alpha^{-2}\Phi(\HSig_*)
				\right)^{-1}
				-
				\left(
				\nabla^2f(\Hu^*) 	
				\right)^{-1}
			}\\
			\leq&
			\norm{	\left(
				\nabla^2f(\Hu^*) 
				+
				2\alpha^{-2}\Phi(\HSig_*)
				\right)^{-1}
				\left(
				2\alpha^{-2}\Phi(\Sigma_*)
				\right)
				\left(
				\nabla^2f(\Hu^*) 	
				\right)^{-1}
			}\\
			\leq&
			\norm
			{
				\left(
				\nabla^2f(\Hu^*) 
				+
				2\alpha^{-2}\Phi(\HSig_*)
				\right)^{-1}
			}_2
			\cdot
			\norm
			{
				\left(
				\nabla^2f(\Hu^*) 	
				\right)^{-1}
			}_2
			\cdot
			\norm
			{
				2\alpha^{-2}\Phi(\Sigma_*)
			}, 
		\end{aligned}
	\end{equation}
	where $\norm{\cdot}_2$ is the spectral norm. 
	The first inequality is because the projection operator onto a convex set is non-expansive \citep{bertsekas2009convex}.  The second inequality used the following fact: for any two nonsingular matrices $A$ and $B$, it holds that
	\begin{equation}\label{eq:A_inv}
		A^{-1} - B^{-1} = A^{-1}\left(B - A\right)B^{-1}.
	\end{equation}
	The last inequality is because it holds that $\norm{AB} \leq \norm{A}_2\norm{B}$ for two any consistent matrices $A$ and $B$.
	
	Now we bound $2\alpha^{-2}\norm{\Phi(\HSig_*)}$ as follows
	\begin{equation}\label{eq:tmp2}
		\begin{aligned}
			&2\alpha^{-2}\norm{\Phi(\HSig_*)} 
			= 
			\EE_u
			\left[
			\tilde{\rho}\left(\alpha\HSig_*^{1/2}u\right)
			\cdot
			\left(\HSig_*^{-1/2}uu^\top \HSig_*^{-1/2} -  \HSig_*^{-1}\right)
			\right]
			\\
			\overset{\eqref{eq:rho_t}}{\leq}&
			\frac{\alpha\gamma\norm{\HSig_*}_2^{3/2}}{6}
			\EE_u
			\left[
			\norm{u}^3
			\cdot
			\norm{
				\left(\HSig_*^{-1/2}uu^\top \HSig_*^{-1/2} -  \HSig_*^{-1}\right)
			}
			\right]\\
			\leq&
			\frac{\alpha\gamma\norm{\HSig_*}_2^{3/2}\cdot \norm{\HSig_*^{-1/2}}_2^2}{6}
			\EE_u
			\left[
			\norm{u}^3\left(\norm{u}^2 + d\right)
			\right]\\
			\leq& \frac{\alpha\gamma\zeta}{6\tau^{3/2}}
			\EE_u
			\left[\norm{u}^5 + d\norm{u}^3\right]
			\stackrel{\eqref{eq:gauss_power}}{\le}
			\frac{\alpha\gamma\zeta}{6\tau^{3/2}}\left((d+5)^{5/2} + d (d+3)^{3/2}\right)
		\end{aligned}
	\end{equation}
	where the last inequality follows from the fact that $\HSig_*$ is in the convex set $\cS$.
	
	By the condition of $\alpha$ in Eqn.~\eqref{eq:alp_cond},
	we have $2\alpha^{-2}\norm{\Phi(\HSig_*)}  
	\leq 
	\frac{\sigma}{2}$
	which implies 
	\begin{equation}
		\label{eq:tmp1}
		\norm
		{
			\left(
			\nabla^2f(\Hu^*) 
			+
			2\alpha^{-2}\Phi(\HSig_*)
			\right)^{-1}
		}_2
		\leq
		2\sigma^{-1}.
	\end{equation}
	Consequently, we can obtain that 
	\begin{align*}
		&\norm
		{
			\HSig_* - \Pi_{\cS}\left(\left(\nabla^2f(\Hu^*) 	
			\right)^{-1}\right)
		}
		\\
		\overset{\eqref{eq:tmp}}{\leq}&
		\norm
		{
			\left(
			\nabla^2f(\Hu^*) 
			+
			2\alpha^{-2}\Phi(\Sigma_*)
			\right)^{-1}
		}_2
		\cdot
		\norm
		{
			\left(
			\nabla^2f(\Hu^*) 	
			\right)^{-1}
		}_2
		\cdot
		\norm
		{
			2\alpha^{-2}\Phi(\Sigma_*)
		}
		\\
		\overset{\eqref{eq:tmp1}}{\leq}&
		2\sigma^{-1}
		\cdot
		\sigma^{-1}
		\cdot
		\norm
		{
			2\alpha^{-2}\Phi(\Sigma_*)
		}
		\overset{\eqref{eq:tmp2}}{\leq}
		2\sigma^{-1}
		\cdot
		\sigma^{-1}
		\cdot
		\frac{\alpha\gamma\zeta}{6\tau^{3/2}}
		\left(
		(d+5)^{5/2} + d (d+3)^{3/2}
		\right)
		\\
		=&
		\frac{\alpha\gamma\zeta}{3\tau^{3/2}\sigma^2}
		\cdot
		\left(
		(d+5)^{5/2} + d (d+3)^{3/2}
		\right)
		.
	\end{align*}
	
	Similarly, we have
	\begin{equation*}
		\norm
		{
			\Sigma_* - \Pi_{\cS}\left(\left(\nabla^2f(\mu^*) 	
			\right)^{-1}\right)
		}
		\leq 
		\frac{\alpha\gamma\zeta}{3\tau^{3/2}\sigma^2}
		\cdot
		\left(
		(d+5)^{5/2} + d (d+3)^{3/2}
		\right).
	\end{equation*}
	Next, we will bound $\norm{\Sigma_* - \hat{\Sigma}_*}$ as follows
	\begin{align*}
		\norm{\Sigma_* - \hat{\Sigma}_*}
		\leq
		\norm
		{
			\Pi_{\cS}\left(\left(\nabla^2f(\mu^*)  \right)^{-1}\right)
			-
			\Pi_{\cS}\left(\left(\nabla^2f(\hat{\mu}_*) 	
			\right)^{-1}\right)
		}
		+
		\frac{2\alpha\gamma\zeta}{3\tau^{3/2}\sigma^2}
		\cdot
		\left(
		(d+5)^{5/2} + d (d+3)^{3/2}
		\right)
	\end{align*}
	We also have
	\begin{align*}
		&\norm
		{
			\Pi_{\cS}\left(\left(\nabla^2f(\mu^*)\right)^{-1}\right)
			-
			\Pi_{\cS}\left(\left(\nabla^2f(\hat{\mu}_*) 	
			\right)^{-1}\right)
		}
		\leq
		\norm{
			\left(\nabla^2f(\mu^*) \right)^{-1}
			-
			\left(\nabla^2f(\hat{\mu}_*) \right)^{-1}
		}
		\\
		\overset{\eqref{eq:A_inv}}{\leq}&
		\norm{
			\left(\nabla^2f(\mu^*)\right)^{-1}
		}
		\cdot
		\norm{
			\left(\nabla^2f(\hat{\mu}_*) \right)^{-1}
		}
		\cdot
		\norm{
			\nabla^2f(\mu^*) - \nabla^2f(\hat{\mu}_*)
		}
		\leq
		\frac{\gamma}{\sigma^2}
		\norm{
			\mu^* - \hat{\mu}_*
		}
		\\
		\leq&
		\frac{\gamma}{\sigma^2}
		\cdot
		\left(
		\frac{2dL\alpha^2}{\sigma\tau} 
		+ 
		\frac{2\gamma \alpha^3 (d + 3)^{3/2}}{3\sigma\tau^{3/2}} 
		+
		\frac{d\alpha^2}{\sigma} \left(\tau^{-1} - \zeta^{-1}  \right)
		\right)^{1/2}.
	\end{align*}
	The first inequality is because of the property that projection operator onto a convex set is non-expansive \citep{bertsekas2009convex}.
	The third inequality is due to $f(\cdot)$ is $\sigma$-strongly convex and $\nabla^2f(\mu)$ is $\gamma$-Lipschitz continuous. 
	The last inequality follows from Theorem~\ref{prop:mu}.
	
	Therefore, we can obtain that 
	\begin{align*}
		\norm{\Sigma_* - \hat{\Sigma}_*}
		\leq&
		\frac{2\alpha\gamma\zeta}{3\tau^{3/2}\sigma^2}
		\cdot
		\left(
		(d+5)^{5/2} + d (d+3)^{3/2}
		\right)
		\\
		&+
		\frac{\gamma}{\sigma^2}
		\cdot
		\left(
		\frac{2dL\alpha^2}{\sigma\tau} 
		+ 
		\frac{2\gamma \alpha^3 (d+3)^{3/2}}{3\sigma\tau^{3/2}} 
		+
		\frac{d\alpha^2}{\sigma} \left(\tau^{-1} - \zeta^{-1}  \right)
		\right)^{1/2}.
	\end{align*}
\end{proof}

\section{Properties of $\tg(\mu)$ and $\TG(\Sigma)$}

In this section, we will give several important properties related to $\tg(\mu)$ and $\TG(\Sigma)$ that will be used in the proof of the convergence rates of Algorithm~\ref{alg:zero_order}.

\begin{lemma}
	\label{lem:tgm_1}
	Denoting  $\tu_i\sim N(0, \Sigma)$ with $\Sigma^{-1} \in \cS$, and $\ti{U} = [\tu_1,\dots, \tu_b] \in\RR^{d\times b}$, we can represent $\tg(\mu)$ defined in Eqn.\eqref{eq:g_mu} as 
	\begin{equation}
		\label{eq:tgm}
		\begin{aligned}
			\tg(\mu)
			=
			\frac{1}{b}\sum_{i=1}^b \left(\nu_i\cdot \tu_i + \tu_i\tu_i^\top\nabla f(\mu)\right)
			= \frac{1}{b}\left(\ti{U}\nu + \ti{U}\ti{U}^\top \nabla f(\mu)\right),
		\end{aligned}
	\end{equation}
	with
	\begin{align*}
		\nu_i = \frac{f(\mu+\alpha \ti{u}_i)-f(\mu-\alpha\ti{u}_i) - \dotprod{\nabla f(\mu), 2\alpha\tu_i}}{2\alpha},\quad\mbox{ and }\quad
		\nu = [\nu_1,\nu_2,\dots, \nu_b]^\top.
	\end{align*}
	Given $0<\delta<1$, if $f(x)$ is $\gamma$-Hessian smooth, then, with probability at least $1-\delta$, it holds that
	\begin{align}
		|\nu_i| \le \frac{\gamma \alpha^2  (2d + 3\log(1/\delta))^{3/2} }{6\tau^{3/2}}. \label{eq:sigma_1}
	\end{align}
\end{lemma}
\begin{proof}
	By the definition of $\tg$, we have
	\begin{align*}
		\tg(\mu) 
		=& 
		\frac{1}{b} \sum_{i=1}^{b}\frac{f(\mu+\alpha \ti{u}_i)-f(\mu-\alpha\ti{u}_i)}{2\alpha} \cdot \ti{u}_i
		\\
		=&\frac{1}{b} \sum_{i=1}^{b}\Big(\frac{f(\mu+\alpha \ti{u}_i)-f(\mu-\alpha\ti{u}_i) - \dotprod{\nabla f(\mu), 2\alpha\tu_i}}{2\alpha} \cdot \ti{u}_i + \tu_i\tu_i^\top\nabla f(\mu)\Big)
		\\
		=&
		\frac{1}{b}\sum_{i=1}^b \left(\nu_i\cdot \tu_i + \tu_i\tu_i^\top\nabla f(\mu)\right)
		=
		\frac{1}{b}\left(\ti{U}\nu + \ti{U}\ti{U}^\top \nabla f(\mu)\right).
	\end{align*}
	If $f(\mu)$ is $\gamma$-Hessian smooth, denoting 
	$a_{\tu_i}(\alpha) =f(\mu+\alpha \tu_i) - f(\mu) - \dotprod{\nabla f(\mu), \alpha\tu_i} - \frac{\mu^2}{2}\dotprod{\nabla^2 f(\mu)\tu_i, \tu_i}$,
	then we have
	\begin{align*}
		|\nu_i| 
		=&
		\frac{ a_{\tu_i}(\alpha) - a_{\tu_i}(-\alpha)}{2\alpha}
		\overset{\eqref{eq:taylor}}{\le}
		\frac{\gamma\norm{\alpha \tu_i}^3}{6\alpha}
		\le
		\frac{\gamma \alpha^2 \norm{\Sigma^{3/2}}\cdot \norm{u}^3}{6}\\
		\stackrel{\eqref{eq:cS_p}}{\le}& 
		\frac{\gamma \alpha^2 \cdot \norm{u}^3}{6\tau^{3/2}} 
		\overset{\eqref{eq:u_norm}}{\le}
		\frac{\gamma  \alpha^2 (2d + 3\log(1/\delta))^{3/2} }{6\tau^{3/2}}.
	\end{align*}
\end{proof}
\begin{lemma}
	Letting $\nu$ and $\ti{U}$ be defined in Lemma~\ref{lem:tgm_1}, given $0 < \delta< 1$, then with a probability at least $1-\delta$, it holds that
	\begin{align}
		\norm{\nu}^2 \le \frac{\gamma b\alpha^4  (2d + 3\log(1/\delta))^3}{36\tau^3}, \mbox{ and }\norm{\ti{U} \nu}^2 
		\le
		\frac{\gamma b \alpha^4  (2d + 3\log(1/\delta))^4}{36\tau^4}. \label{eq:nu}
	\end{align}
\end{lemma}
\begin{proof} First, by the definition of $\nu$, we have
	\begin{align*}
		\norm{\nu}^2 
		= 
		\sum_{i}^{b}\nu_i^2 
		\stackrel{\eqref{eq:sigma_1}}{\le} \frac{\gamma b\alpha^4  (2d + 3\log(1/\delta))^3}{36\tau^3}.
	\end{align*}
	Furthermore, we have
	\begin{align*}
		\norm{\ti{U} \nu}^2
		=&
		\nu^\top \ti{U}^\top \ti{U} \nu
		=
		\nu^\top U^\top \Sigma U \nu
		\le
		\norm{\Sigma} \cdot \nu^\top U^\top  U \nu
		\le
		\norm{\Sigma} \cdot \max_i{|\nu_i|^2} \cdot \norm{ U \mathbf{1} }^2\\
		\le&
		\frac{\gamma \alpha^4  (2d + 3\log(1/\delta))^3}{36\tau^4} \cdot \norm{ U \mathbf{1} }^2,
	\end{align*}
	where $\mathbf{1} = [1,\dots, 1]^\top \in \RR^b$ and $U \in \RR^{d\times b}$ with each entry being of standard normal distribution.
	Note that each entry of $U \mathbf{1}$ satisfies that $[U \mathbf{1}]_i \sim N(0, b)$.
	By Lemma~\ref{lem:chi}, with a probability at least $1-\delta$, it holds that
	\begin{align}
		\label{eq:U_1}
		\norm{U\mathbf{1}}^2 
		=
		\sum_{i=1}^d [U \mathbf{1}]_i^2 
		\le b(2d + 3\log(1/\delta)).
	\end{align}
	Therefore, we obtain the final result about $\norm{\ti{U} \nu}^2  $ which concludes the proof.
\end{proof}

\begin{lemma}
	Letting $\ti{U}$ be defined in Lemma~\ref{lem:tgm_1} and given $0<\delta<1$, then  with a probability at least $1-\delta$, it holds that
	\begin{equation}\label{eq:gd_low}
		\norm{\ti{U}^\top \nabla f(\mu)  }^2 \ge \left(b - 2\sqrt{b\log(1/\delta)}\right) \norm{ \Sigma^{1/2} \nabla f(\mu) }^2.
	\end{equation}
\end{lemma}
\begin{proof}
	By the definition of $ \ti{U} $, we have $ \ti{U}^\top \nabla f(\mu) = U^\top \Sigma^{1/2} \nabla f(\mu)$ with $U \in \RR^{d\times b}$ whose  entries are of standard normal distribution.
	Note that each entry of $U^\top \Sigma^{1/2} \nabla f(\mu)$ satisfies that $[U^\top  \Sigma^{1/2} \nabla f(\mu)]_i \sim N(0, \norm{ \Sigma^{1/2} \nabla f(\mu) }^2)$ with $i = 1,\dots, b$.
	By Lemma~\ref{lem:chi}, with probability at least $1-\delta$, it holds that  
	\begin{align*}
		\norm{\ti{U}^\top \nabla f(\mu)  }^2  = \sum_{i=1}^{b} [U^\top  \Sigma^{1/2} \nabla f(\mu)]_i^2 \ge \left(b - 2\sqrt{b\log(1/\delta)}\right) \norm{ \Sigma^{1/2} \nabla f(\mu) }^2.
	\end{align*}
\end{proof}

\begin{lemma}
	\label{lem:tgm}
	Assume that $ \nabla^2f(\mu) \preceq \mathscr{L} \cdot \Sigma^{-1}  \preceq \mL \tau^{-1} \cdot I$ and $\nabla^2f(\mu)$ is $\gamma$-Lipschitz continuous. 
	Denote $U = [u_1,u_2,\dots, u_b] \in \RR^{d\times b}$ with $u_i \sim N(0, I_d)$.
	Given $0<\delta<1/2$, with probability $1-\delta$, it holds that
	\begin{equation}\label{eq:hh}
		\begin{aligned}
			&\norm{\tg(\mu)}_{\nabla^2 f(\mu)}^2 \\
			\le&
			\frac{2\mL}{b^2} \left(\nu^\top U^\top U\nu +  	\nabla^\top f(\mu) \ti{U}U^\top  U \ti{U}^\top \nabla f(\mu)\right)\\
			\le&
			\frac{2\mL \left(\sqrt{d} + \sqrt{b} + \sqrt{2\log(2/\delta)}\right)^2}{b^2} \norm{U^\top  \Sigma^{1/2}\nabla f(\mu)}^2
			+
			\frac{\mL\gamma^2 \alpha^4 (2d + 3\log(1/\delta))^4  }{18 b\tau^3}.
		\end{aligned}
	\end{equation}
\end{lemma}
\begin{proof}
	It holds that $\ti{U}  = \Sigma^{1/2} U$ by the definition of $\ti{U}$ in Lemma~\ref{lem:tgm_1}.
	Using $H$ to denote $\nabla^2 f(\mu)$, we have
	\begin{equation*}
		\begin{aligned}
			\norm{\tg(\mu)}_H^2 
			=&
			\tg^\top H \tg
			=\frac{1}{b^2} \Big(\nu^\top \ti{U}^\top H \ti{U}\nu + 2 \nu^\top \ti{U}^\top H \ti{U}\ti{U}^\top \nabla f(\mu) + \nabla^\top  f(\mu) \ti{U}\ti{U}^\top H \ti{U}\ti{U}^\top \nabla f(\mu)\Big)
			\\
			\le&
			\frac{2}{b^2} \left(\nu^\top \ti{U}^\top H \ti{U}\nu + \nabla^\top  f(\mu) \ti{U}\ti{U}^\top H \ti{U}\ti{U}^\top \nabla f(\mu)\right),
		\end{aligned}
	\end{equation*}
	where the last inequality is because of the Cauchy's inequality.
	
	We will bound above two terms in order.
	First, we have
	\begin{align*}
		\nabla^\top f(\mu) \ti{U}\ti{U}^\top H \ti{U}\ti{U}^\top \nabla f(\mu)
		=&
		\nabla^\top f(\mu) \Sigma^{1/2}UU^\top \Sigma^{-1/2} H \Sigma^{1/2} UU^\top \Sigma^{1/2} \nabla f(\mu)
		\\
		\le&
		\mL \cdot 	\nabla^\top f(\mu) \Sigma^{1/2}UU^\top  UU^\top \Sigma^{1/2} \nabla f(\mu),
	\end{align*}
	where the last inequality is because of the assumption that $ \nabla^2f(\mu) \preceq \mathscr{L} \cdot \Sigma^{-1}  $.
	Based on above inequality,	we can furthermore obtain that
	\begin{align*}
		\nabla^\top f(\mu) \ti{U}\ti{U}^\top H \ti{U}\ti{U}^\top \nabla f(\mu)
		\le&
		\mL\cdot\norm{U}^2 \cdot \norm{U^\top \Sigma^{1/2} \nabla f(\mu)}^2
		\\
		\overset{\eqref{eq:smax}}{\le}&
		\mL \left(\sqrt{d} + \sqrt{b} + \sqrt{2\log(2/\delta)}\right)^2 \cdot \norm{U^\top  \Sigma^{1/2}\nabla f(\mu)}^2.
	\end{align*}
	Furthermore, we have
	\begin{align*}
		\nu^\top \ti{U}^\top H \ti{U}\nu 
		=&
		\nu^\top U^\top \Sigma^{1/2} H \Sigma^{1/2} U \nu
		\le
		\mL \cdot \nu^\top U^\top U \nu
		\le
		\mL \max_i \{|\nu_i|^2\} \cdot  \norm{U \mathbf{1}}^2\\
		\stackrel{\eqref{eq:U_1}}{\le}&
		\mL b(2d + 3\log(1/\delta)) \cdot  \max_i \{|\nu_i|^2\}
		\stackrel{\eqref{eq:sigma_1}}{\le} 
		\frac{b\mL\gamma^2 \alpha^4 (2d + 3\log(1/\delta))^4  }{36\tau^3}.
	\end{align*}	
	Combining above results, we can obtain the final result.
\end{proof}

\begin{lemma}
	Let objective function $f(\cdot)$ be $\gamma$-Hessian smooth and $\Sigma^{-1} \preceq \tau^{-1} \cdot I$.  Given $0<\delta<1/2$, it holds with probability at least $1-\delta$ that
	\begin{equation}
		\label{eq:tgm_3}
		\norm{\tg(\mu)} 
		\le
		\frac{ \gamma \alpha^2 (2d + 3\log(1/\delta))^{2} }{6b^{1/2}\tau^2}
		+
		\frac{\left(\sqrt{d} + \sqrt{b} + \sqrt{2\log(2/\delta)}\right) }{b\tau^{1/2}} \cdot \norm{U^\top\Sigma^{1/2} \nabla f(x)}.
	\end{equation}
\end{lemma}
\begin{proof}
	By Eqn.~\eqref{eq:tgm}, we have
	\begin{align*}
		\norm{\tg}  
		\le&
		\frac{1}{b}\left(\norm{\ti{U}\sigma} + \norm{\ti{U}\ti{U}^\top \nabla f(x)}\right).
	\end{align*}
	We will bound above terms in order. 
	First, we have  
	\begin{align*}
		\norm{\ti{U}\sigma} 
		\le&  \norm{\Sigma^{1/2}} \norm{U \mathbf{1}} \cdot \max_i\{\sigma_i\}
		\stackrel{\eqref{eq:sigma_1}\eqref{eq:U_1}}{\le} \frac{ b^{1/2}\gamma \alpha^2 (2d + 3\log(1/\delta))^{2} \norm{\Sigma^{1/2}}}{6\tau^{3/2}}\\
		\le&
		\frac{ b^{1/2}\gamma \alpha^2 (2d + 3\log(1/\delta))^{2} }{6\tau^2}.
	\end{align*}
	We also have
	\begin{align*}
		\norm{\ti{U}\ti{U}^\top \nabla f(x)}^2 
		=& 
		\nabla^\top f(x) \Sigma^{1/2} UU^\top \Sigma UU^\top \Sigma^{1/2} \nabla f(x)
		\\
		\le& \norm{\Sigma} \cdot \norm{U}^2 \cdot \norm{U^\top \Sigma^{-1/2} \nabla f(x)}^2
		\\
		\overset{\eqref{eq:smax}}{\le}& \left(\sqrt{d} + \sqrt{b} + \sqrt{2\log(2/\delta)}\right)^2 \norm{\Sigma}\cdot \norm{U^\top \Sigma^{1/2} \nabla f(x)}^2\\
		\stackrel{\eqref{eq:cS}}{\le}& \tau^{-1}\left(\sqrt{d} + \sqrt{b} + \sqrt{2\log(2/\delta)}\right)^2 \cdot \norm{U^\top \Sigma^{1/2} \nabla f(x)}^2.
	\end{align*}
	Combining above results, we can obtain the final result.
\end{proof}

\begin{lemma}
	Let $\Sigma^{-1} \in \cS'$ and  $\BG(\Sigma)$ be defined in Eqn.~\eqref{eq:bG}. Assume that $\nabla^2f(\mu)$ is $\gamma$-Lipschitz continuous.
	Given $0<\delta<1$, then with a probability at least $1-\delta$, it holds that
	\begin{equation}\label{eq:G_norm}
		\begin{aligned}
			&\norm{\left(f(\mu - \alpha \Sigma^{1/2}u) + f(\mu + \alpha \Sigma^{1/2}u) - 2 f(\mu) \right) \left( \Sigma^{-1/2} uu^\top \Sigma^{-1/2} - \Sigma^{-1} \right)} \\
			\le&
			2L\tau^{-1} \alpha^2 (2d + 2\log(1/\delta)) \left( \zeta (2d + 2\log(1/\delta)) + \sqrt{d} \zeta   \right),
		\end{aligned}
	\end{equation}
	and
	\begin{equation} \label{eq:GG}
		\begin{aligned}
			&\norm{\BG(\Sigma) - \frac{1}{2b} \sum_{i=1}^b u_i^\top \Sigma^{1/2} \nabla^2 f(\mu) \Sigma^{1/2} u_i \left(\Sigma^{-1/2} u_i u_i^\top \Sigma^{-1/2} - \Sigma^{-1}\right)}_2 \\
			\le &
			\frac{(2 d + 3\log(1/\delta))^{3/2}\cdot (2d + 3\log (1/\delta) + 1) \cdot \zeta\gamma}{4\tau^{3/2}  } \cdot \alpha
		\end{aligned}
	\end{equation}
\end{lemma}
\begin{proof}
	Using the Taylor's expansion, we have
	\begin{align*}
		f(\mu - \alpha \Sigma^{1/2} u) =& f(\mu) + \dotprod{\nabla f(\mu), -\alpha \Sigma^{1/2}u} + \alpha^2 u^\top \Sigma^{1/2} \left[\int_{0}^{1}\nabla^2f(\mu - t \alpha \Sigma^{1/2}u) \; dt\right] \Sigma^{1/2} u\\
		f(\mu + \alpha \Sigma^{1/2} u) =& f(\mu) + \dotprod{\nabla f(\mu), \alpha \Sigma^{1/2}u} + \alpha^2 u^\top \Sigma^{1/2} \left[ \int_{0}^{1}\nabla^2f(\mu + t \alpha \Sigma^{1/2}u) \; dt\right]\Sigma^{1/2} u.
	\end{align*}
	Using above Taylor's expansions, we have
	\begin{align*}
		&\norm{\left(f(\mu - \alpha \Sigma^{1/2}u) + f(\mu + \alpha \Sigma^{1/2}u) - 2 f(\mu) \right) \left( \Sigma^{-1/2} uu^\top \Sigma^{-1/2} - \Sigma^{-1} \right)}\\
		=& 
		\alpha^2 \norm{ u^\top \Sigma^{1/2} \left(\int_{0}^{1}\left[\nabla^2f(\mu + t \alpha \Sigma^{1/2}u) +  \nabla^2f(\mu - t \alpha \Sigma^{1/2}u)\right]\; dt   \right) \Sigma^{1/2} u \cdot \left( \Sigma^{-1/2} uu^\top  \Sigma^{-1/2} - \Sigma^{-1} \right) }\\
		\le&
		\alpha^2 \norm{\Sigma^{1/2} u}^2 \cdot \int_{0}^1  \norm{\nabla^2f(\mu + t \alpha \Sigma^{1/2}u) +  \nabla^2f(\mu - t \alpha \Sigma^{1/2}u)} \; dt \cdot \norm{ \Sigma^{-1/2} uu^\top  \Sigma^{-1/2} - \Sigma^{-1} } \\
		\le&
		2L\alpha^2 \cdot  \norm{\Sigma^{1/2} u}^2 \cdot \left( \norm{\Sigma^{-1/2}u}^2 + \norm{\Sigma^{-1}} \right) \\
		\le&
		2L\tau^{-1} \alpha^2 (2d + 2\log(1/\delta)) \left( \zeta (2d + 2\log(1/\delta)) + \sqrt{d} \zeta   \right),
	\end{align*}
	where the second inequality is because of $f(\cdot)$ is $L$-smooth and the last inequality is because of  Eqn.~\eqref{eq:cS} and Eqn.~\eqref{eq:u_norm}.

	Using above Taylor's expansions again, we can obtain that
	\begin{align*}
		&\frac{f(\mu - \alpha \Sigma^{1/2} u_i) + f(\mu - \alpha \Sigma^{1/2} u_i) - 2 f(\mu)}{2\alpha^2} - u_i^\top \Sigma^{1/2} \nabla^2 f(\mu) \Sigma^{1/2} u_i\\
		=&
		u_i^\top \Sigma^{1/2} \left[\int_{0}^{1}\frac{\nabla^2f(\mu - t\alpha \Sigma^{1/2}u_i) + \nabla^2f(\mu + t\alpha \Sigma^{1/2}u_i)}{2}\;dt\right] \Sigma^{1/2}u_i  
		-
		u_i^\top \Sigma^{1/2} \nabla^2 f(\mu) \Sigma^{1/2} u_i \\
		\le&
		u_i^\top \Sigma^{1/2}\left[\int_{0}^{1}\frac{\norm{\nabla^2f(\mu - t\alpha \Sigma^{1/2}u_i) + \nabla^2f(\mu + t\alpha \Sigma^{1/2}u_i) - 2 \nabla^2 f(\mu)}}{2}\; dt\right] \Sigma^{1/2}u_i  \\
		\stackrel{\eqref{eq:gamma_1}}{\le}&
		u_i^\top \Sigma u_i  \cdot \int_0^1 \gamma t \alpha\norm{\Sigma^{1/2}u_i}\; dt
		=
		u_i^\top \Sigma u_i \cdot \frac{\alpha \gamma \norm{\Sigma^{1/2} u_i}}{2} 
		\stackrel{\eqref{eq:cS}\eqref{eq:u_norm}}{\le}
		\frac{\gamma(2 d + 2\log(1/\delta))^{3/2}}{2\tau^{3/2}}\cdot \alpha.
	\end{align*}	
	Thus, 
	\begin{align*}
		&\norm{\BG - \frac{1}{2b} \sum_{i=1}^b u_i^\top \Sigma^{1/2} \nabla^2 f(\mu) \Sigma^{1/2} u_i \left(\Sigma^{-1/2} u_i u_i^\top \Sigma^{-1/2} - \Sigma^{-1}\right)}_2\\
		\le&
		\frac{1}{2b} \sum_{i=1}^b \left|\frac{f(\mu - \alpha \Sigma^{1/2} u_i) + f(\mu - \alpha \Sigma^{1/2} u_i) - 2 f(\mu)}{2\alpha^2} - u_i^\top \Sigma^{1/2} \nabla^2 f(\mu) \Sigma^{1/2} u_i\right| \\
		& \cdot \norm{\Sigma^{-1/2} u_i u_i^\top \Sigma^{-1/2} - \Sigma^{-1}}_2\\
		\le&
		\frac{\gamma(2 d + 2\log(1/\delta))^{3/2}}{4\tau^{3/2}}\cdot \alpha \cdot \left( \norm{  \Sigma^{-1/2} u_i u_i^\top \Sigma^{-1/2} }_2 + \norm{\Sigma^{-1}}_2 \right)\\
		\stackrel{\eqref{eq:cS}\eqref{eq:u_norm}}{\le}&
		\frac{(2 d + 2\log(1/\delta))^{3/2}\cdot (2d + 2\log (1/\delta) + 1) \cdot \zeta \gamma}{4\tau^{3/2}  } \cdot \alpha.
	\end{align*}
\end{proof}

\begin{lemma}
	Let $\Sigma^{-1} \in \cS'$ and  $\TG(\Sigma)$ be defined in Eqn.~\eqref{eq:TG}. 
	Given $0<\delta<1$, denoting $H^* = \Pi_{\cS'}(\nabla^2 f(\mu^*))$,  then with a probability at least $1-\delta$, it holds that
	\begin{align} 
		&\norm{\TG(\Sigma_k)}^2 \le   \frac{2L^2 (2d+3\log(1/\delta))^2 \left( \zeta (2d + 3\log(1/\delta)) + \sqrt{d} \zeta\right)^2 }{b\tau^2} + \frac{d \zeta^2}{2b}, \label{eq:G_norm_1}\\
		&\norm{ \TG(\Sigma_k) - \left(H^* - \Sigma_k^{-1}\right) }^2 \le \frac{2L^2 (2d+3\log(1/\delta))^2 \left( \zeta (2d + 3\log(1/\delta)) + \sqrt{d} \zeta\right)^2 }{b\tau^2} + \frac{d \zeta^2}{2b} 
		. \label{eq:G_norm_2}
	\end{align}
\end{lemma}
\begin{proof}
	By the definition of $\TG(\Sigma)$ in Eqn.~\eqref{eq:TG}, we have
	\begin{align*}
		&\norm{\TG(\Sigma_k)}^2 \\
		=& 
		\norm{\frac{1}{2b\alpha^2} \sum_{i=1}^{b}
			\left[\left(f(\mu_k - \alpha \Sigma_k^{1/2}u_i) + f(\mu_k + \alpha \Sigma_k^{1/2}u_i) - 2f(\mu_k) \right)\left(\Sigma_k^{-1/2}u_iu_i^\top \Sigma_k^{-1/2} -  \Sigma_k^{-1}\right)\right] 
			- \Sigma_k^{-1}}^2 \\
		\le& \frac{1}{4b^2\alpha^4} \sum_{i = 1}^b\norm{ \left(f(\mu_k - \alpha \Sigma_k^{1/2}u_i) + f(\mu_k + \alpha \Sigma_k^{1/2}u_i) - 2f(\mu_k) \right)\left(\Sigma_k^{-1/2}u_iu_i^\top \Sigma_k^{-1/2} -  \Sigma_k^{-1}\right) - \alpha^2 \Sigma_k^{-1} }^2\\
		\le&
		\frac{1}{2b^2 \alpha^4} \left( \sum_{i=1}^b \norm{ \left(f(\mu_k - \alpha \Sigma_k^{1/2}u_i) + f(\mu_k + \alpha \Sigma_k^{1/2}u_i) - 2f(\mu_k) \right)\left(\Sigma_k^{-1/2}u_iu_i^\top \Sigma_k^{-1/2} -  \Sigma_k^{-1}\right) }^2 + b\alpha^4 \norm{\Sigma_k^{-1}}^2 \right)\\
		&\stackrel{\eqref{eq:G_norm}\eqref{eq:cS_p}}{\le}
		\frac{2L^2 (2d+3\log(1/\delta))^2 \left( \zeta (2d + 3\log(1/\delta)) + \sqrt{d} \zeta\right)^2 }{b\tau^2} + \frac{d \zeta^2}{2b}.
	\end{align*}
	
	Similarly, it holds that
	\begin{align*}
		&\norm{ \TG(\Sigma_k) - \left(H^* - \Sigma_k^{-1}\right) }^2 \\
		=&
		\norm{\frac{1}{2b\alpha^2} \sum_{i=1}^{b}
			\left[\left(f(\mu_k - \alpha \Sigma_k^{1/2}u_i) + f(\mu_k + \alpha \Sigma_k^{1/2}u_i) - 2f(\mu_k) \right)\left(\Sigma_k^{-1/2}u_iu_i^\top \Sigma_k^{-1/2} -  \Sigma_k^{-1}\right)\right] 
			- H^*}^2 \\
		\le&
		\frac{2L^2 (2d+3\log(1/\delta))^2 \left( \zeta (2d + 3\log(1/\delta)) + \sqrt{d} \zeta\right)^2 }{b\tau^2} + \frac{d \zeta^2}{2b},
	\end{align*}
	where the last inequality is because of the objective function $f(\cdot)$ is $L$-smooth.
\end{proof}

\section{Proofs of Section~\ref{sec:conv}}

\subsection{Proof of Lemma~\ref{lem:gd_dec_1}}

\begin{proof}
	By the update rule of Algorithm~\ref{alg:zero_order}, we have
	\begin{align*}
		f(\mu_{k+1}) 
		=& f(\mu_k - \eta_1 \tg(\mu_k)) 
		\stackrel{\eqref{eq:L_1}}{\le}
		f(\mu_k) - \eta_1\dotprod{\nabla f(\mu_k), \tg(\mu_k)} + \frac{L\eta_1^2}{2}\norm{ \tg(\mu_k)  }^2\\
		\stackrel{\eqref{eq:tgm}}{=}&
		f(\mu_k) -  \frac{\eta_1}{b} \dotprod{ \nabla f(\mu_k), \ti{U}_k\nu + \ti{U}_k\ti{U}_k\nabla f(\mu_k) } + \frac{L\eta_1^2}{2b^2} \norm{ \ti{U}_k \nu + \ti{U}_k\ti{U}_k^\top \nabla f(\mu_k)  }^2\\
		\le&
		f(\mu_k) - \frac{\eta_1}{2b} \dotprod{\nabla f(\mu_k), \ti{U}_k\ti{U}_k^\top \nabla f(\mu_k)} + \frac{\eta_1\norm{\nu}^2}{2b} \\
		&+ \frac{L\eta_1^2}{b^2} \left( \norm{ \ti{U}_k\ti{U}_k^\top \nabla f(\mu_k) }^2 + \norm{ \ti{U}_k \nu}^2 \right),
	\end{align*}
	where the last inequality is because of Cauchy's inequality and the fact that $2ab \le a^2 + b^2$.
	
	Furthermore, with a probability at least $1-\delta$, it holds that
	\begin{align*}
		&\norm{ \ti{U}_k\ti{U}_k^\top \nabla f(\mu_k) }^2
		=
		\nabla^\top f(\mu_k) \ti{U}_k\ti{U}_k^\top \ti{U}_k\ti{U}_k^\top \nabla f(\mu_k)
		=
		\nabla^\top f(\mu_k) \ti{U}_k U \Sigma_k U^\top \ti{U}_k^\top \nabla f(\mu_k)\\
		\le&
		\norm{\Sigma_k}_2 \norm{U}^2 \cdot \norm{\ti{U}_k^\top \nabla f(\mu_k)  }^2
		\stackrel{\eqref{eq:cS}\eqref{eq:smax}}{\le}
		\tau^{-1} \left(\sqrt{d} + \sqrt{b} + \sqrt{2\log(2/\delta)}\right)^2 \cdot \norm{\ti{U}_k^\top \nabla f(\mu_k)  }^2.
	\end{align*}
	
	By replacing the value of $\eta_{1,k}$, we have
	\begin{align*}
		f(\mu_{k+1}) - f(\mu^*) 
		\le& 
		f(\mu_k) - f(\mu^*) - \frac{1}{16L\zeta} \norm{\ti{U}_k^\top \nabla f(\mu_k)  }^2 + \frac{\norm{\nu}^2}{8Lc_1\zeta } + \frac{\norm{ \ti{U}_k \nu }^2}{16L\zeta^2}\\
		\stackrel{\eqref{eq:gd_low}\eqref{eq:nu}}{\le}&
		f(\mu_k) - f(\mu^*) - \frac{c_2\tau}{16c_1L } \norm{\Sigma^{1/2}\nabla f(\mu_k)}^2 + \frac{c_3^3 \gamma b  \alpha^4}{2^5\cdot 3^2\cdot c_1 L\zeta\tau^3}\left(1 + \frac{c_1c_3}{2\tau\zeta}\right) \\
		\le&
		f(\mu_k) - f(\mu^*) - \frac{c_2\tau }{16c_1L\zeta } \norm{\nabla f(\mu_k)}^2 + \Delta_{\alpha,1}\\
		\le&
		f(\mu_k) - f(\mu^*) - \frac{c_2\tau\sigma  }{16c_1L\zeta } \left(f(\mu_k) - f(\mu^*)\right) + \Delta_{\alpha,1} \\
		=&
		\left(1 -  \frac{c_2\tau\sigma  }{16c_1L\zeta }\right) \cdot \left(f(\mu_k) - f(\mu^*)\right)+ \Delta_{\alpha,1},
	\end{align*}
	where the last inequality is because of $f(\cdot)$ is $\sigma$-strongly convex.
\end{proof}

\subsection{Proof of Lemma~\ref{lem:dec_local}}

Before the proof, we first give a lemma which  constructs connection between $\norm{\nabla f(\mu_k)}_{\Sigma_k}^2$ and $f(\mu_k) - f(\mu^*)$.
\begin{lemma}
	\label{lem:nab_f}
	Let $f(x)$ satisfy the properties described in Lemma~\ref{lem:dec_local} and the covariance matrix $\Sigma_k$ satisfy $ \xi_k \cdot \Sigma_k^{-1} \preceq \nabla^2f(\mu_k) \preceq \mL_k \cdot \Sigma_k^{-1} $. Then $\norm{\nabla f(\mu_k)}_{\Sigma_k}^2$ satisfies that
	\begin{align}
		-\norm{\nabla f(\mu_k)}^2_{\Sigma_k} \leq -\xi_k\left(f(\mu_k)-f(\mu^*)\right), \quad\mbox{ if } \quad f(\mu_k) - f(\mu^*) 
		\le
		\frac{\xi_k^2\sigma^3}{8\gamma^2 (L\tau^{-1} + 2\xi_k)^2}. \label{eq:nab_rst}
	\end{align}
\end{lemma}
\begin{proof}
	Now, we begin to give the connections between $\norm{\nabla f(\mu_k)}_{\Sigma_k}^2$ and $f(\mu_k) - f(\mu^*)$. First, by the Taylor's expansion, we have
	\begin{align*}
		\nabla f(\mu^*) = \nabla f(\mu_k) + \nabla^2f(\mu_k)( \mu^*-\mu_k ) + \int_{0}^{1}\left(\nabla^2f(\mu_k - s (\mu_k - \mu^*)) - \nabla^2f(\mu_k)\right)( \mu^*-\mu_k ).
	\end{align*}
	Combining with the fact that $\nabla f(\mu^*) = 0$, we can obtain that
	\begin{align*}
		\nabla f(\mu_k) = \nabla^2f(\mu_k)(\mu_k -  \mu^* ) + \int_{0}^{1}\left(\nabla^2f(\mu_k - s (\mu_k - \mu^*)) - \nabla^2f(\mu_k)\right)( \mu_k - \mu^* )\;ds.
	\end{align*}
	
	Let us denote that
	\begin{align*}
		\Delta_1 = \int_{0}^{1}\left(\nabla^2f(\mu^* - s (\mu_k - \mu^*)) - \nabla^2f(\mu^*)\right)(\mu_k - \mu^*)\;ds,
	\end{align*}
	which can be bounded as follows
	\begin{equation}
		\label{eq:delta_1}
		\norm{\Delta_1} \overset{\eqref{eq:gamma_1}}{\leq} \gamma \int_{0}^{1} s \norm{\mu_k - \mu^*}^2 \; ds = \frac{\gamma}{2}\norm{\mu_k - \mu^*}^2.
	\end{equation}
	
	Let us denote $H^* = \nabla^2 f(\mu^*)$ and $H = \nabla^2 f(\mu_k)$. We have
	\begin{align*}
		-\norm{\nabla f(\mu_k)}^2_{\Sigma_k}
		=&- \norm{H(\mu_k - \mu^*)+\Delta_1}^2_{\Sigma_k}\\
		\leq&
		-\norm{H(\mu_k - \mu^*)}^2_\Sigma + 2 \dotprod{H(\mu_k - \mu^*), \Sigma_k\Delta_1}\\
		\leq& - \xi_k\norm{\mu_k - \mu^*}_{H}^2 + 2 \dotprod{H(\mu_k - \mu^*), \Sigma_k\Delta_1},
	\end{align*}
	where the last inequality is because of the condition that $\xi_k\cdot \Sigma_k \preceq \nabla^2 f(\mu_k)$.
	Furthermore, we have
	\begin{align*}
		f(\mu_k) \overset{\eqref{eq:taylor}}{\leq}& f(\mu^*) + \langle \nabla f(\mu^*), \mu_k - \mu^* \rangle + \frac{1}{2}\langle \nabla^2f(\mu^*) (\mu_k-\mu^*), \mu_k - \mu^*\rangle + \frac{\gamma}{6}\norm{\mu_k - \mu^*}^3\\
		=&
		f(\mu^*) + \frac{1}{2}\norm{\mu_k-\mu^*}^2_{H^*} + \frac{\gamma}{6}\norm{\mu_k - \mu^*}^3\\
		=&
		f(\mu^*) + \frac{1}{2}\norm{\mu_k-\mu^*}^2_{H} + \frac{1}{2}\norm{\mu_k-\mu^*}^2_{H^* - H} + \frac{\gamma}{6}\norm{\mu_k - \mu^*}^3\\
		\stackrel{\eqref{eq:gamma_1}}{\le}&
		f(\mu^*) + \frac{1}{2}\norm{\mu_k-\mu^*}^2_{H} + \frac{2\gamma}{3}\norm{\mu_k - \mu^*}^3.
	\end{align*}
	Hence, we have
	\begin{equation}
		\small
		-\frac{1}{2}\norm{\mu_k-\mu^*}^2_{H} \leq -\left(f(\mu_k)-f(\mu^*)\right) + \frac{2\gamma}{3}\norm{\mu_k-\mu^*}^3. \label{eq:H_s}
	\end{equation}
	
	Then we begin to bound $\dotprod{H(\mu_k - \mu^*), \Sigma\Delta_1}$. 
	First, we have
	\begin{equation*}
		\dotprod{H(\mu_k - \mu^*), \Sigma \Delta_1} \leq \norm{\mu_k - \mu^*}\norm{\Delta_1} \norm{H\Sigma} 
		\stackrel{\eqref{eq:delta_1}}{\le} \frac{\gamma}{2}\norm{\mu_k - \mu^*}^3 \norm{H\Sigma}
		\stackrel{\eqref{eq:L_1}\eqref{eq:cS}}{\le} \frac{\gamma L }{2\tau} \norm{\mu_k - \mu^*}^3.
	\end{equation*}
	Therefore, we obtain that
	\begin{equation*}
		\small
		\begin{split}
			-\norm{\nabla f(\mu_k)}^2_{\Sigma_k}
			\leq&
			- \xi_k\norm{\mu_k - \mu^*}_{H}^2 + \gamma L \tau^{-1}\norm{\mu_k - \mu^*}^3\\
			\overset{\eqref{eq:H_s}}{\leq}&
			-2\xi_k\left(f(\mu_k)-f(\mu^*)\right) + \frac{4\xi_k\gamma}{3}\norm{\mu_k-\mu^*}^3 + \gamma L \tau^{-1}\norm{\mu_k - \mu^*}^3\\
			\leq&
			-2\xi_k\left(f(\mu_k)-f(\mu^*)\right) + \gamma(L\tau^{-1} +2\xi_k)\norm{\mu_k-\mu^*}^3\\
			\leq&
			-2\xi_k\left(f(\mu_k)-f(\mu^*)\right) + \gamma(L\tau^{-1} +2\xi_k)\cdot \left(\frac{2}{\sigma}\right)^{3/2} \left(f(\mu_k)-f(\mu^*)\right)^{3/2},
		\end{split}
	\end{equation*}
	where the last inequality is because of the $\sigma$-strong convexity of $f(x)$. 
	Replacing the condition on $ f(\mu_k) - f(\mu^*) $ to above inequality, we can obtain that 
	\begin{align*}
		-\norm{\nabla f(\mu_k)}^2_{\Sigma_k} 
		\le 	-2\xi_k\left(f(\mu_k)-f(\mu^*)\right) + \xi_k \left(f(\mu_k)-f(\mu^*)\right)
		= -\xi_k\left(f(\mu_k)-f(\mu^*)\right).
	\end{align*}
\end{proof}

\begin{proof}[Proof of Lemma~\ref{lem:dec_local}]
	Taking a random step from $\mu_k$, we have
	\begin{equation*}
		\begin{aligned}
			&f(\mu_{k+1})
			=f(\mu_k - \eta_{1,k} \tg(\mu_k))\\
			\overset{\eqref{eq:taylor}}{\leq}&
			f(\mu_k)-\eta_{1,k}\dotprod{\nabla f(\mu_k), \tg(\mu_k)}
			+\frac{\eta_{1,k}^2}{2}\norm{\tg(\mu_k)}_H^2
			+\frac{\eta_{1,k}^3\gamma}{6}\norm{\tg(\mu_k)}^3\\
			\stackrel{\eqref{eq:tgm}\eqref{eq:hh}}{\le}&
			f(\mu_k)-\frac{\eta_{1,k}}{b}\dotprod{\nabla f(\mu_k), \ti{U}\nu + \ti{U}\ti{U}^\top \nabla f(\mu_k)} +\frac{\eta_{1,k}^3\gamma}{6}\norm{\tg(\mu_k)}^3  \\
			&+ \frac{\eta_{1,k}^2 \mL_k c_1 }{b^2} \norm{U^\top \Sigma_k^{1/2} \nabla f(\mu_k)}^2 
			+  \frac{\mL_k \eta_{1,k}^2\gamma^2 \alpha^4 c_3^4}{18b\tau^3}\\
			\le& 
			f(\mu_k)- \frac{\eta_{1,k}}{2b} \nabla^\top f(\mu_k)\ti{U}\ti{U}^\top \nabla f(\mu_k) + \frac{\eta_{1,k}}{2b} \norm{\nu}^2 +\frac{\eta_{1,k}^3\gamma}{6}\norm{\tg(\mu_k)}^3 \\
			&+\frac{\eta_{1,k}^2 \mL_k c_1 }{b^2} \norm{U^\top \Sigma_k^{1/2} \nabla f(\mu_k)}^2 
			+  \frac{\eta_{1,k}^2\mL_k \gamma^2 \alpha^4 c_3^4}{18b\tau^3}\\
			\stackrel{\eqref{eq:tgm_3}\eqref{eq:nu}}{\le}&
			f(\mu_k)- \frac{\eta_{1,k}}{2b} \nabla^\top f(\mu_k)\ti{U}\ti{U}^\top \nabla f(\mu_k) + \frac{\eta_{1,k}^2 \mL_k c_1^{1/2} }{b^2} \norm{U^\top \Sigma_k^{1/2} \nabla f(\mu_k)}^2 \\
			&+  \frac{2 \gamma c_1^{3/2} \eta_{1,k}^3}{3b^3\tau^{3/2}} \norm{U^\top \Sigma_k^{1/2} \nabla f(\mu_k)}^2
			+ \frac{\eta_{1,k}^3\gamma^4\alpha^6 c_3^6}{2^2\cdot 3^4\cdot b^{3/2} \tau^6} +  \frac{\mL_k \eta_{1,k}^2\gamma^2 \alpha^4 c_3^4}{18b\tau^3} + \frac{\eta_{1,k} \gamma c_3^3 \alpha^4 }{2^3\cdot 3^2\cdot \tau^3}\\
			=& 
			f(\mu_k) - \frac{1}{4\mL_k c_1} \norm{U^\top \Sigma_k^{1/2} \nabla f(\mu_k)}^2 + \frac{\gamma}{2^5\cdot 3\cdot \mL_k^3 c_1^{3/2} \tau^{3/2}} \norm{U^\top \Sigma_k^{1/2} \nabla f(\mu_k)}^3 \\
			&+\frac{b^{3/2}\gamma^4 c_3^6 \alpha^6 }{2^8\cdot 3^4\cdot  \mL_k^3 c_1^3 \tau^6} + \frac{b\gamma^2 c_3^4 \alpha^4}{2^5 \cdot 3^2 \cdot \mL_k\tau^3 c_1^2} + \frac{b\gamma c_3^3 \alpha^4}{2^5\cdot 3^2 \cdot \tau^3 c_1},
		\end{aligned}
	\end{equation*}
	where the third inequality follows from Cauchy's inequality and the last equality uses the value of the step size $\eta_k$.
	
	Because of $\sigma \cdot I \preceq \nabla^2 f(\mu) \preceq L \cdot I$ and $\tau \cdot I \preceq \Sigma_k^{-1} \preceq \zeta\cdot I$, it holds that
	$ \frac{\sigma}{\zeta} \le \mL_k $. Thus, we can upper bound that
	\begin{align*}
		&\frac{b^{3/2}\gamma^4 c_3^6 \alpha^6 }{2^8\cdot 3^4\cdot  \mL_k^3 c_1^3 \tau^6} + \frac{b\gamma^2 c_3^4 \alpha^4}{2^5 \cdot 3^2 \cdot \mL_k\tau^3 c_1^2} + \frac{b\gamma c_3^3 \alpha^4}{2^5\cdot 3^2 \cdot \tau^3 c_1} \\
		\le&
		\frac{b^{3/2}\zeta^3\gamma^4  c_3^6 \alpha^6 }{2^8\cdot 3^4\cdot  \sigma^3 c_1^3 \tau^6} + \frac{b\zeta\gamma^2 c_3^4 \alpha^4}{2^5 \cdot 3^2 \cdot \sigma\tau^3 c_1^2} + \frac{b\gamma c_3^3 \alpha^4}{2^5\cdot 3^2 \cdot \tau^3 c_1}  := \Delta_{\alpha,2}
	\end{align*}
	
	Furthermore, 
	\begin{align*}
		\norm {\ti{U}^\top \nabla f(\mu_k)} 
		\le&
		\norm{U} \cdot \norm{\Sigma^{1/2} \nabla f(\mu_k)}
		\stackrel{\eqref{eq:smax}}{\le}
		c_1^{1/2} \cdot \norm{\Sigma^{1/2} \nabla f(\mu_k)}
		\le
		\sqrt{2c_1 L\tau^{-1} (f(\mu_k) - f(\mu^*))}.
	\end{align*}
	Combining above equation with the condition on $f(\mu_k) - f(\mu^*)$, we can obtain 
	\begin{align*}
		\frac{\gamma \norm{\ti{U}^\top\nabla f(\mu_k)} }{2^5\cdot 3 \cdot c_1^{3/2} \cdot \tau^{3/2} \mL_k^3}
		\le  \frac{1}{8\mL_k c_1}.
	\end{align*}

	Therefore
	\begin{align*}
		f(\mu_{k+1}) - f(\mu^*) 
		\le&
		f(\mu_k) - f(\mu^*) - \frac{1}{8\mL_k c_1} \norm{U^\top \Sigma_k^{1/2} \nabla f(\mu_k)}^2 + \Delta_{\alpha,2}\\
		\stackrel{\eqref{eq:gd_low}}{\le}&
		f(\mu_k) - f(\mu^*) - \frac{b - 2\sqrt{b\log(1/\delta)}}{8\mL_k c_1} \norm{\Sigma_k^{1/2} \nabla f(\mu_k)}^2 + \Delta_{\alpha,2}\\
		\stackrel{\eqref{eq:nab_rst}}{\le}&
		f(\mu_k) - f(\mu^*) - \frac{\xi_k  c_2}{8\mL_k c_1} \Big( f(\mu_k) - f(\mu^*)\Big) + \Delta_{\alpha,2}\\
		=&
		\left( 1 - \frac{\xi_k c_2}{8\mL_k c_1}  \right) \cdot \Big( f(\mu_k) - f(\mu^*)\Big) + \Delta_{\alpha,2}.
	\end{align*}
\end{proof}

\subsection{Proof of Lemma~\ref{lem:Sig_dec}}

The proof of Lemma~\ref{lem:Sig_dec} consists of several lemmas. 
The main idea is the same to the convergence analysis of stochastic gradient descent in \cite{rakhlin2012making} since our update of $\Sigma_k^{-1}$ can be viewed as a kind of stochastic gradient descent just as discussed in Section~\ref{sec:conv}.
\begin{lemma}
	Let us denote $H^* = \Pi_{\cS'}\left(\nabla^2 f(\mu^*)\right)$ and set $\eta_{2,k} = \frac{1}{k}$. Given $0<\delta<1$, then $\Sigma_k$ generated by Algorithm~\ref{alg:zero_order} has the following property with a probability at least $1-\delta$
	\begin{equation} \label{eq:Sig_dec} \small
		\norm{\Sigma_{k+1}^{-1}-H^*}^2 
		\le 
		\frac{1}{k(k-1)} \sum_{t=2}^k (t-1) \dotprod{\BG(\Sigma_t) - H^*, \Sigma_t^{-1} - H^*}
		+
		\frac{1}{k(k-1)} \sum_{t=2}^{k}\frac{t-1}{t}\norm{\TG(\Sigma_t)}^2.
	\end{equation}
	Specifically, it also holds that
	\begin{equation}
		\norm{\Sigma_2^{-1} - H^*}^2 \le \frac{4L^2 c_3^2 \left( c_3\zeta  + \sqrt{d} \zeta\right)^2 }{b\tau^2} + \frac{d\zeta^2}{b}. \label{eq:Sig_2_bnd}
	\end{equation}
	$\BG(\Sigma)$ and $c_3$ are defined in Eqn.~\eqref{eq:bG} and~\eqref{eq:c1} respectively.
\end{lemma}
\begin{proof}
	By the update rule of $\Sigma_k$, we have
	\begin{align*}
		&\norm{\Sigma_{k+1}^{-1} - H^*}^2 \\
		=& 
		\norm{\Pi_{\cS'}(\Sigma_k^{-1} + \eta_{2,k} \TG(\Sigma_k)) - H^*}^2
		\le
		\norm{\Sigma_k^{-1} + \eta_{2,k} \TG(\Sigma_k) - H^*}^2 \\
		=& 
		\norm{\Sigma_k^{-1} - H^*}^2 + 2\eta_{2,k}\dotprod{\Sigma_k^{-1} - H^*, \TG(\Sigma_k)} + \eta_{2,k}^2\norm{\TG(\Sigma_k)}^2\\
		=& 
		\norm{\Sigma_k^{-1} - H^*}^2 + 2\eta_{2,k}\dotprod{\Sigma_k^{-1} - H^*, H^* - \Sigma_k^{-1}} \\&+ 2\eta_{2,k}\dotprod{\Sigma_k^{-1} - H^*, \TG(\Sigma_k) - \left(H^* - \Sigma_k^{-1}\right)}+ \eta_{2,k}^2\norm{\TG(\Sigma_k)}^2\\
		=&
		\left(1-\frac{2}{k}\right) \norm{\Sigma_k^{-1} - H^*}^2 +\frac{2}{k}\dotprod{\TG(\Sigma_k) - \left(H^* - \Sigma_k^{-1}\right), \Sigma_k^{-1} - H^*} + \frac{1}{k^2}\norm{\TG(\Sigma_k)}^2,
	\end{align*}
	where the first inequality is because of the non-expansiveness of projection onto convex set.
	
	Unwinding this recursive inequality till $k = 2$, we get that for any $k \geq 2$, 
	\begin{align*}
		&\norm{\Sigma_{k+1}^{-1}-H^*}^2 \\
		\leq& 
		2\sum_{t=2}^{k} \frac{1}{t}\left(\prod_{j=t+1}^{k}\left(1-\frac{2}{j}\right)\right)\dotprod{\TG(\Sigma_t) - \left(H^* - \Sigma_t^{-1}\right), \Sigma_t^{-1} - H^*}\\
		&+ 
		\sum_{t=2}^{k}\frac{1}{t^2}\left(\prod_{j=t+1}^{k}\left(1-\frac{2}{j}\right)\right) \norm{\TG(\Sigma_t)}^2\\
		=& 
		\frac{1}{k(k-1)} \sum_{t=2}^k(t-1)\dotprod{\TG(\Sigma_t) - \left(H^* - \Sigma_t^{-1}\right), \Sigma_t^{-1} - H^*}
		+ \frac{1}{k(k-1)} \sum_{t=2}^{k}\frac{t-1}{t}\norm{\TG(\Sigma_t)}^2\\
		=&
		\frac{1}{k(k-1)} \sum_{t=2}^k (t-1) \dotprod{\BG(\Sigma_t) - H^*, \Sigma_t^{-1} - H^*}
		+
		\frac{1}{k(k-1)} \sum_{t=2}^{k}\frac{t-1}{t}\norm{\TG(\Sigma_t)}^2.
	\end{align*}
	
	Furthermore, for $k=1$, we have
	\begin{align*}
		&\norm{\Sigma_2^{-1} - H^*}^2 \\
		\le&
		\left(1 - \frac{2}{1}\right) \norm{\Sigma_1^{-1} - H^*}^2 + \frac{2}{1}\dotprod{\TG(\Sigma_1) - \left(H^* - \Sigma_1^{-1}\right), \Sigma_1^{-1} - H^*} + \frac{1}{1^2}\norm{\TG(\Sigma_1)}^2\\
		\le& 
		- \norm{\Sigma_1^{-1} - H^*}^2 + 2 \norm{ \TG(\Sigma_1) - \left(H^* - \Sigma_1^{-1}\right) } \cdot \norm{ \Sigma_1^{-1} - H^* } + \norm{\TG(\Sigma_1)}^2 \\
		\le& 
		- \norm{\Sigma_1^{-1} - H^*}^2 + \norm{ \TG(\Sigma_1) - \left(H^* - \Sigma_1^{-1}\right) }^2 + \norm{ \Sigma_1^{-1} - H^* }^2 + \norm{\TG(\Sigma_1)}^2 \\
		=& 
		\norm{ \TG(\Sigma_1) - \left(H^* - \Sigma_1^{-1}\right) }^2 + \norm{\TG(\Sigma_1)}^2
		\stackrel{\eqref{eq:G_norm_1}\eqref{eq:G_norm_2}}{\le}
		\frac{4L^2 c_3^2 \left( c_3\zeta  + \sqrt{d} \zeta\right)^2 }{b\tau^2} + \frac{d \zeta^2}{b}.
	\end{align*}
\end{proof}

Next, we mainly to bound the value $\frac{1}{k(k-1)} \sum_{t=2}^k (t-1) \dotprod{\BG(\Sigma_t) - H^*, \Sigma_t^{-1} - H^*}$. 
We first decompose this term into several terms.
\begin{lemma} \label{lem:GH}
	Lettting us denote that $H_t = \Pi_{\cS'}\left( \nabla^2 f(\mu_t)\right)$, it holds that
	\begin{align*}
		&\dotprod{\BG(\Sigma_t) - H^*, \Sigma_t^{-1} - H^*} 
		= \dotprod{ \frac{1}{2b} \sum_{i=1}^b u_i^\top \Sigma_t^{1/2} H_t \Sigma_t^{1/2} u_i \left(\Sigma_t^{-1/2} u_i u_i^\top \Sigma_t^{-1/2} - \Sigma_t^{-1}\right)- H_t,  \Sigma_t^{-1} - H^*  } \\
		& + \dotprod{ \BG - \frac{1}{2b} \sum_{i=1}^b u_i^\top \Sigma_t^{1/2} H_t \Sigma_t^{1/2} u_i \left(\Sigma_t^{-1/2} u_i u_i^\top \Sigma_t^{-1/2} - \Sigma_t^{-1}\right),  \Sigma_t^{-1} - H^*} 
		+\dotprod{ H_t - H^*, \Sigma_t^{-1} - H^* }.
	\end{align*}
\end{lemma}
\begin{proof}
	It holds that
	\begin{align*}
		&\dotprod{\BG - H^*, \Sigma_t^{-1} - H^*} 
		= 
		\dotprod{\BG - H_t + H_t - H^*, \Sigma_t^{-1} - H^*}\\
		=&
		\dotprod{ \frac{1}{2b} \sum_{i=1}^b u_i^\top \Sigma_t^{1/2} H_t \Sigma_t^{1/2} u_i \left(\Sigma_t^{-1/2} u_i u_i^\top \Sigma_t^{-1/2} - \Sigma_t^{-1}\right)- H_t,  \Sigma_t^{-1} - H^*  } \\
		& + \dotprod{ \BG - \frac{1}{2b} \sum_{i=1}^b u_i^\top \Sigma_t^{1/2} H_t \Sigma_t^{1/2} u_i \left(\Sigma_t^{-1/2} u_i u_i^\top \Sigma_t^{-1/2} - \Sigma_t^{-1}\right),  \Sigma_t^{-1} - H^*} \\
		&+\dotprod{ H_t - H^*, \Sigma_t^{-1} - H^* }.
	\end{align*}
	
\end{proof}

\begin{lemma}
	It holds that
	\begin{equation}\label{eq:alpha_err}
		\begin{aligned}
			&\sum_{t=2}^k (t-1)\dotprod{ \BG - \frac{1}{2b} \sum_{i=1}^b u_i^\top \Sigma_t^{1/2} H_t \Sigma_t^{1/2} u_i \left(\Sigma_t^{-1/2} u_i u_i^\top \Sigma_t^{-1/2} - \Sigma_t^{-1}\right),  \Sigma_t^{-1} - H^*} \\
			&+\sum_{t=2}^k (t-1) \dotprod{H_t - H^*, \Sigma_t^{-1} - H^*} \\
			\le&
			\sqrt{ \sum_{t=2}^k \norm{H_t - H^*}^2 } \cdot \sqrt{ \sum_{t=2}^k (t-1)^2  \norm{ \Sigma_t^{-1} - H^* }^2  } \\
			&+\frac{c_3^{3/2}\cdot (c_3 + 1) \cdot \zeta\gamma\cdot \sqrt{k-1}}{4\tau^{3/2} d^{1/2} } \cdot \alpha \cdot \sqrt{\sum_{t=2}^k (t-1)^2 \norm{ \Sigma_t^{-1} - H^* }^2}.
		\end{aligned}
	\end{equation}
\end{lemma}
\begin{proof}
	By the Cauchy' inequality, we have
	\begin{align*}
		&\sum_{t=2}^k (t-1) \dotprod{H_t - H^*, \Sigma_t^{-1} - H^*} 
		\le 
		\sum_{t=2}^k (t-1) \norm{H_t - H^*}\cdot \norm{ \Sigma_t^{-1} - H^* }\\
		\le&
		\sqrt{ \sum_{t=2}^k \norm{H_t - H^*}^2 } \cdot \sqrt{ \sum_{t=2}^k (t-1)^2  \norm{ \Sigma_t^{-1} - H^* }^2  }.
	\end{align*}
	
	Similarly, we can obtain that
	\begin{align*}
		&\sum_{t=2}^k (t-1)\dotprod{ \BG - \frac{1}{2b} \sum_{i=1}^b u_i^\top \Sigma_t^{1/2} H_t \Sigma_t^{1/2} u_i \left(\Sigma_t^{-1/2} u_i u_i^\top \Sigma_t^{-1/2} - \Sigma_t^{-1}\right),  \Sigma_t^{-1} - H^*} \\
		\le&  
		\sum_{t=2}^k (t-1)  \norm{ \BG - \frac{1}{2b} \sum_{i=1}^b u_i^\top \Sigma_t^{1/2} H_t \Sigma_t^{1/2} u_i \left(\Sigma_t^{-1/2} u_i u_i^\top \Sigma_t^{-1/2} - \Sigma_t^{-1}\right) } \cdot \norm{ \Sigma_t^{-1} - H^* } \\
		\stackrel{\eqref{eq:GG}}{\le}&
		\frac{c_3^{3/2}\cdot (c_3 + 1) \cdot \zeta}{4\tau^{3/2} d^{1/2} } \cdot \alpha \cdot \sum_{t=2}^k (t-1) \norm{ \Sigma_t^{-1} - H^* }\\
		\le& 
		\frac{c_3^{3/2}\cdot (c_3 + 1) \cdot \zeta\cdot \sqrt{k-1}}{4\tau^{3/2} d^{1/2} } \cdot \alpha \cdot \sqrt{\sum_{t=2}^k (t-1)^2 \norm{ \Sigma_t^{-1} - H^* }^2}.
	\end{align*}
	Combining above results, we can obtain the finaly result.
\end{proof}

In the next five lemmas, we try to bound the value $$\sum_{t=2}^{k}\dotprod{ \BG - \frac{1}{2b} \sum_{i=1}^b u_i^\top \Sigma_t^{1/2} H_t \Sigma_t^{1/2} u_i \left(\Sigma_t^{-1/2} u_i u_i^\top \Sigma_t^{-1/2} - \Sigma_t^{-1}\right),  \Sigma_t^{-1} - H^*}.$$
\begin{lemma} \label{lem:E}
	Let the eigenvalue decomposition of  $\Sigma_t^{1/2} H_t \Sigma_t^{1/2}$ be defined as follows:
	\begin{align*}
		\Sigma_t^{1/2} H_t \Sigma_t^{1/2} = V_t \Lambda_t V_t^\top, \mbox{ with } V_tV_t^\top = V_t^\top V_t = I, \Lambda_t = \diag\{ \lambda_1^{(t)},\dots, \lambda_d^{(t)}\}.
	\end{align*}
	Denote that $\bu_{t,i} = V_t^\top u_i$ and $E_{t,i}
	\triangleq 
	\bu_{t,i}^\top  \Lambda_t \bu_{t,i} \left(\bu_{t,i}\bu_{t,i}^\top - I\right) - 2\Lambda_t$.
	It holds that
	\begin{equation} \label{eq:E}
		\begin{aligned}
			&\dotprod{\frac{1}{2} u_i^\top \Sigma_t^{1/2} H_t \Sigma_t^{1/2} u_i \left( \Sigma_t^{-1/2} u_iu_i^\top \Sigma_t^{-1/2} -\Sigma_t^{-1} \right) - H_t, \Sigma_t^{-1} - H^*} \\
			=&
			\frac{1}{2} \dotprod{ E_{t,i},  V_t^\top \Sigma_t^{-1/2} \left(\Sigma_t^{-1} - H^*\right) \Sigma_t^{-1/2} V_t}.
		\end{aligned}
	\end{equation}
\end{lemma}
\begin{proof} 
	First, we have
	\begin{align*}
		&\frac{1}{2} u_i^\top \Sigma_t^{1/2} H_t \Sigma_t^{1/2} u_i \left( \Sigma_t^{-1/2} u_iu_i^\top \Sigma_t^{-1/2} -\Sigma_t^{-1} \right) - H_t\\
		=& 
		\frac{1}{2} \Sigma_t^{-1/2} V_t \left( u_i^\top V_t \Lambda_t V_t^\top u_i \left( V_t^\top u_i u_i^\top V_t - I  \right) - 2\Lambda_t \right) V_t^\top \Sigma_t^{-1/2}\\
		=&
		\frac{1}{2} \Sigma_t^{-1/2} V_t \left(\bu_{t,i}^\top  \Lambda_t \bu_{t,i} \left(\bu_{t,i}\bu_{t,i}^\top - I\right) - 2\Lambda_t\right)V_t^\top \Sigma_t^{-1/2}\\
		=& \frac{1}{2} \Sigma_t^{-1/2} V_t  E_{t,i}V_t^\top \Sigma_t^{-1/2} 
		,
	\end{align*}
	where the second equality is because we denote $\bu_{t,i} = V_t^\top u_i$ and the last equality is due to the definition of $E_{t,i}$. 
	Using the properties of inner product of matrices, we can obtain the final result.
\end{proof}

\begin{lemma}\label{lem:EXY}
	Letting $E_{t,i}
	$ be defined in Lemma~\ref{lem:E} and given a parameter $  A_t > 0$, we can represent $E_{t,i}$ that
	\begin{align}
		E_{t,i} =& \Big(X_{t,i} - \EE[X_{t,i}]\Big) + \Big(Y_{t,i} - \EE[Y_{t,i}]\Big), \label{eq:EXY}
	\end{align}
	with 
	\begin{align*}
		X_{t,i} =& \bu_{t,i}^\top  \Lambda_t \bu_{t,i} \left(  \bu_{t,i} \bu_{t,i}^\top - I  \right) \mathbbm{1}_{\norm{\bu_{t,i}}^2 \le A_t}, \quad\mbox{ and } \quad Y_{t,i} = \bu_{t,i}^\top  \Lambda_t \bu_{t,i} \left(  \bu_{t,i} \bu_{t,i}^\top - I  \right) \mathbbm{1}_{\norm{\bu_{t,i}}^2 > A_t}.
	\end{align*}
\end{lemma}
\begin{proof}
	First, it holds that
	\begin{equation*}
		\bu_{t,i} = \bu_{t,i} \mathbbm{1}_{\norm{\bu_{t,i}}^2 \le A_t}  + \bu_{t,i} \mathbbm{1}_{\norm{\bu_{t,i}}^2 > A_t}.
	\end{equation*}
	Accordingly, we can obtain that
	\begin{align*}
		\bu_{t,i}^\top  \Lambda_t \bu_{t,i} \left(\bu_{t,i}\bu_{t,i}^\top - I\right) 
		=& 
		\bu_{t,i}^\top  \Lambda_t \bu_{t,i} \left(  \bu_{t,i} \bu_{t,i}^\top - I  \right) \mathbbm{1}_{\norm{\bu_{t,i}}^2 \le A_t} + \bu_{t,i}^\top  \Lambda_t \bu_{t,i} \left(  \bu_{t,i} \bu_{t,i}^\top - I  \right) \mathbbm{1}_{\norm{\bu_{t,i}}^2 > A_t} \\
		=&
		X_{t,i} + Y_{t, i}.
	\end{align*}
	
	Furthermore, since it  holds that $\bu_{t,i} = V_t^\top u_i$, $u_i \sim N(0, I)$ and $V_t$ is orthonormal, it holds that $\bu\sim N(0, I)$. 
	Combining Lemma~\ref{lem:exp_tg}, we can obtain that
	\begin{align*}
		\EE\left[ \bu_{t,i}^\top  \Lambda_t \bu_{t,i} \left(\bu_{t,i}\bu_{t,i}^\top - I\right)  \right] = 2 \Lambda_t
		= \EE\left[  X_{t,i} + Y_{t, i} \right].
	\end{align*}
	Thus, 
	\begin{align*}
		E_{t,i}
		=&
		\bu_{t,i}^\top  \Lambda_t \bu_{t,i} \left(\bu_{t,i}\bu_{t,i}^\top - I\right) - 2\Lambda_t
		= X_{t,i} + Y_{t, i} - \EE\left[  X_{t,i} + Y_{t, i} \right]\\
		=& \Big(X_{t,i} - \EE[X_{t,i}]\Big) + \Big(Y_{t,i} - \EE[Y_{t,i}]\Big).
	\end{align*}
\end{proof}

\begin{lemma}\label{lem:X}
	Let $A_t$ and $\Lambda_t$  be defined in Lemma~\ref{lem:EXY} and Lemma~\ref{lem:E} respectively. We  define that
	\begin{align*}
		\Gamma_t := A_t \norm{\Lambda_t}_2 \left( \zeta A_t +  \tr\left(\Sigma_t^{-1}\right) \right) \cdot \norm{ \Sigma_t^{-1} - H^* }_2.
	\end{align*}
	Given sequence $\{z_{t,k}\}_{t=1}^k$ with $z_{t,k} \ge 0$, $0<\delta<1$ and $X_{t,i}$ being defined in Lemma~\ref{lem:EXY}, it holds that with a probability at least $1 - \frac{\delta}{k^2}$ that
	\begin{align*}
		\left|\sum_{t=2}^kz_{t,k}\dotprod{X_t - \EE\left[X_t\right] , \frac{1}{2} V^\top \Sigma_t^{-1/2} B_t \Sigma_t^{-1/2} V }\right| 
		\le
		\sqrt{ \frac{2}{b} \log \frac{k^2}{\delta} \cdot \sum_{t=2}^k z_{t,k}^2 \Gamma_t^2 }.
	\end{align*}
\end{lemma}
\begin{proof}
	By the definition of $X_{t,i}$, we have
	\begin{align*}
		&\left|\dotprod{X_{t,i}, \frac{1}{2} V_t^\top \Sigma_t^{-1/2} \left(\Sigma_t^{-1} - H^*\right) \Sigma_t^{-1/2} V_t}\right|\\
		=&
		\frac{1}{2} \left|\dotprod{ \bu_{t,i}^\top  \Lambda_t \bu_{t,i} \left(  \bu_{t,i} \bu_{t,i}^\top - I  \right) \mathbbm{1}_{\norm{\bu_{t,i}}^2 \le A_t},  V_t^\top \Sigma_t^{-1/2} \left(\Sigma_t^{-1} - H^*\right) \Sigma_t^{-1/2} V_t }\right|\\
		\le&
		\frac{1}{2} \left| \dotprod{ \bu_{t,i}^\top  \Lambda_t \bu_{t,i} \cdot \bu_{t,i} \bu_{t,i}^\top \cdot \mathbbm{1}_{\norm{\bu_{t,i}}^2 \le A_t},  V_t^\top \Sigma_t^{-1/2} \left(\Sigma_t^{-1} - H^*\right) \Sigma_t^{-1/2} V_t } \right| \\
		& + \frac{1}{2} \left| \dotprod{ \bu_{t,i}^\top  \Lambda_t \bu_{t,i}  \cdot \mathbbm{1}_{\norm{\bu_{t,i}}^2 \le A_t} \cdot I,  V_t^\top \Sigma_t^{-1/2} \left(\Sigma_t^{-1} - H^*\right) \Sigma_t^{-1/2} V_t } \right| \\
		\le& 
		\frac{A_t^2 \norm{\Lambda_t}_2}{2}  \cdot\norm{\Sigma_t^{-1/2} \left(\Sigma_t^{-1} - H^*\right) \Sigma_t^{-1/2} }_2 + \frac{A_t \norm{ \Lambda_t }_2}{2} \left|\tr\left( \Sigma_t^{-1/2} \left(\Sigma_t^{-1} - H^*\right) \Sigma_t^{-1/2} \right)\right|\\
		\le&
		\frac{ \zeta A_t^2 \norm{\Lambda_t}_2  }{2} \cdot \norm{ \Sigma_t^{-1} - H^* }_2 + \frac{A_t \norm{\Lambda_t}_2 \cdot \tr(\Sigma_t^{-1})}{2} \cdot \norm{ \Sigma_t^{-1} - H^* }_2 = \frac{\Gamma_t}{2} ,
	\end{align*}
	where the last inequality is because of Eqn.~\eqref{eq:cS} and the von  Neumann's trace theorem.
	
	Applying the Hoeffding's inequality (refer to Lemma~\ref{lem:hoeffding}), we can obtain that
	\begin{align*}
		\PP\left( \sum_{t=1}^kz_{t,k} \cdot \frac{1}{b}\sum_{i=1}^b\dotprod{X_{t,i} - \EE\left[X_{t,i}\right] , \frac{1}{2} V^\top \Sigma_t^{-1/2} \left( \Sigma_k^{-1} - H^* \right) \Sigma_t^{-1/2} V } \ge \psi \right) 
		\le
		\exp\left( -\frac{b\psi^2/2}{ \sum_{t=1}^{k}z_{t,k}^2\Gamma_t^2 }   \right).
	\end{align*}
	
	For any $\delta \in (0, 1)$, letting the right hand side of above equation be $\delta / k^2$, then with probability at least $1 - \delta/k^2$, it holds that
	\begin{align*}
		\left|\sum_{t=2}^kz_{t,k}\dotprod{X_t - \EE\left[X_t\right] , \frac{1}{2} V^\top \Sigma_t^{-1/2} B_t \Sigma_t^{-1/2} V }\right| 
		\le
		\sqrt{ \frac{2}{b} \log \frac{k^2}{\delta} \cdot \sum_{t=2}^k z_{t,k}^2 \Gamma_t^2 }
	\end{align*}
\end{proof}

\begin{lemma}{\label{lem:Y}}
	Given parameter $0 <\delta<1$, define $A_t = 8\log \frac{1}{b\delta} + 16 \log( t + 1) + d$. 
	Given sequence $\{z_{t,k}\}_{t=1}^k$ with $z_{t,k} \ge 0$ and $Y_{t,i}$ being defined in Lemma~\ref{lem:EXY}, with a probability at least $1-\delta$, it holds that 
	\begin{align*}
		&\left|\sum_{t=0}^kz_{t,k} \cdot  \frac{1}{b}\sum_{i=1}^b\dotprod{Y_{t,i} - \EE\left[Y_{t,i}\right] , \frac{1}{2} V_t^\top \Sigma_t^{-1/2} \left( \Sigma_t^{-1} - H^* \right) \Sigma_t^{-1/2} V_t }\right| \\
		\le&
		\sum_{t=0}^k z_{t,k} \cdot \frac{3\tr(\Lambda_t)}{d} \cdot \frac{\delta \left(A_t^2 + 16 A_t + 144\right)}{(t+1)^2} \cdot \tr(\Sigma_t^{-1}) \cdot \norm{ \Sigma_t^{-1} - H^*}.
	\end{align*}
\end{lemma}
\begin{proof}
	By the standard Chi-square concentration inequality (refer to Eqn.~(2.18) and Example~2.11 of \citet{wainwright2019high}), it holds that
	\begin{align}
		\PP\left( \norm{\bu_{t,i}}^2 \ge (\beta+1) d \right) \le \exp\left(-d \beta / 8\right), \forall\; \beta\ge 1, \bu_{t,i} \sim N(0, I_d). \label{eq:chi}
	\end{align}
	Combining the definition of $A_t$, we have 
	\begin{align}
		\PP\left(\norm{ \bu_{t,i} }^2 \ge A_t\right) \le \frac{\delta}{b(t+1)^2}. \label{eq:chi_1}
	\end{align}
	Thus, we can obtain that
	\begin{equation*}
		\PP\left(\exists\; t \ge 0, 1 \le i \le b, \norm{\bu_{t,i}}^2 \ge A_t\right) 
		\le \sum_{t=1}^{\infty} \sum_{i=1}^b \PP\left( \norm{ \bu_{t,i} }^2 \ge A_t \right) 
		\le \sum_{t=1}^{\infty} \frac{\delta}{(t+1)^2} = \frac{\pi^2 \delta}{6}.
	\end{equation*}
	That is,
	\begin{align}
		\PP \Big( Y_{t,i} = 0, \forall t \ge 0, 1\le i \le b \Big) \ge 1 -  \frac{\pi^2 \delta}{6}. \label{eq:Y0}
	\end{align}
	
	Next, we will bound the value of $\EE [Y_{t,i}]$ and $\sum_{t=0}^{k} z_{t,k} \EE[Y_{t,i}]$.
	Noting that for $\bu\sim N(0, I_d)$, we have $\bu = \norm{\bu} \cdot \bu'$, where $\bu'\sim \mathrm{Unif}(\mathcal{S}^{d-1})$ and $\mathrm{Unif}(\mathcal{S}^{d-1})$ means that uniform distribution on sphere of dimension $d$. 
	Furthermore, it has the property that $\norm{\bu}$ is independent of $\bu'$.
	
	Then, we can obtain that
	\begin{align*}
		\tr(\Lambda) \cdot I_d + 2 \Lambda =& \EE\left[ \bu^\top \Lambda \bu  \cdot \bu\bu^\top\right] = \EE\left[\norm{\bu}^4\right] \cdot \EE\left[ \bu'^\top\Lambda\bu'\cdot \bu'\bu'^\top\right] = (d^2 + 2d)\cdot \EE\left[ \bu'^\top\Lambda\bu'\cdot \bu'\bu'^\top\right] \\
		\tr(\Lambda) \cdot I_d =& \EE\left[ \bu^\top \Lambda \bu \cdot I_d \right] = \EE\left[ \norm{\bu}^2 \right] \cdot \EE\left[\bu'^\top \Lambda \bu'\cdot I_d\right] = d \cdot  \EE\left[\bu'^\top \Lambda \bu'\cdot I_d\right].
	\end{align*}  
	Thus, we have
	\begin{align*}
		\EE\left[Y_{t,i}\right] 
		=& 
		\EE\left[\bu_{t,i}^\top  \Lambda_t \bu_{t,i} \left(  \bu_{t,i} \bu_{t,i}^\top - I  \right) \mathbbm{1}_{\norm{\bu_{t,i}}^2 > A_t}\right]\\
		=& \EE\left[\left(\norm{\bu_{t,i}}^4 \cdot \bu_{t,i}'^\top  \Lambda_t \bu_{t,i}' \cdot \bu_{t,i}'\bu_{t,i}'^\top  - \norm{\bu_{t,i}}^2 \bu_{t,i}'^\top \Lambda_t \bu_{t,i}'  \right) \mathbbm{1}_{\norm{\bu_{t,i}}^2 > A_t}\right] \\
		=& \EE\left[ \left(\frac{\tr(\Lambda_t)\cdot I_d + 2 \Lambda_t}{d^2 + 2d} \cdot \norm{\bu_{t,i}}^4 - \frac{\tr(\Lambda_t)}{d} \cdot I_d \right)\mathbbm{1}_{\norm{\bu_{t,i}}^2 > A_t}\right],
	\end{align*}
	which leads to 
	\begin{align*}
		\norm{\EE\left[Y_{t,i}\right]}
		\le 
		\max\left\{ 
		\frac{3\tr(\Lambda_t)}{d^2 + 2d} \EE\left[\norm{\bu_{t,i}}^4 \mathbbm{1}_{\norm{\bu_{t,i}}^2 > A_t}\right],
		\frac{\tr(\Lambda_t)}{d} \EE\left[ \norm{\bu_{t,i}}^2 \mathbbm{1}_{\norm{\bu_{t,i}}^2 > A_t} \right]
		\right\}.
	\end{align*}
	Furthermore,
	\begin{align*}
		&\EE\left[\norm{\bu_{t,i}}^4 \mathbbm{1}_{\norm{\bu_{t,i}}^2 > A_t}\right]\\
		=&
		\int_{0}^{\infty} \PP\left(  \norm{\bu_{t,i}}^4 \mathbbm{1}_{\norm{\bu_{t,i}}^2 > A_t} >s \right)\; ds
		=
		\int_{0}^{\infty} \PP\left(  \norm{\bu_{t,i}}^2 > \max\{ \sqrt{s}, A_t \} \right)\; ds\\
		=&\int_{0}^{A_t^2} \PP\left(  \norm{\bu_{t,i}}^2 > A_t  \right)\; ds + \int_{A_t^2}^{\infty} \PP\left( \norm{\bu_{t,i}}^2 > \sqrt{s} \right)\; ds \\
		\stackrel{\eqref{eq:chi}\eqref{eq:chi_1}}{\le}&
		\int_{0}^{A_t^2} \frac{\delta}{(t+1)^2} \; ds + \int_{A_t^2}^{\infty} \exp\left(-\frac{\sqrt{s} - d}{8}\right) \; ds 
		= 
		\frac{\delta A_t^2}{(t+1)^2} + \int_{A_t}^{\infty} 2x \exp\left(-\frac{x - d}{8}\right) \; dx
		\\
		=&
		\frac{\delta A_t^2}{(t+1)^2} + 16(A_t+8)\exp\left( -\frac{A_t - d}{8} \right)
		\le
		\frac{\delta(A_t^2 + 16A_t + 144)}{(t+1)^2}.
	\end{align*}
	Similarly, we have
	\begin{align*}
		\EE\left[\norm{\bu_{t,i}}^2 \mathbbm{1}_{\norm{\bu_{t,i}}^2 > A_t}\right]
		=&
		\int_{0}^{\infty} \PP\left(\norm{\bu_{t,i}}^2 \mathbbm{1}_{\norm{\bu_{t,i}}^2 > A_t} > s\right)\; ds
		= 
		\int_{0}^{\infty} \PP\left(\norm{\bu_{t,i}}^2 >\max\{ s, A_t \}\right)\; ds\\
		\stackrel{\eqref{eq:chi}\eqref{eq:chi_1}}{\le}&
		\int_{0}^{A_t} \frac{\delta}{(t+1)^2}\; ds
		+
		\int_{A_t}^{\infty} \exp\left(-\frac{s - d}{8}\right) \; ds 
		\le
		\frac{\delta(A_t^2 + 8)}{(t+1)^2}.
	\end{align*}
	Thus, we have
	\begin{align*}
		\norm{\EE\left[Y_{t,i}\right]}
		\le \frac{3\tr(\Lambda_t)}{d} \cdot \frac{\delta \left(A_t^2 + 16 A_t + 144\right)}{(t+1)^2}.
	\end{align*}
	
	Thus, with a probability at least $1 - \delta$, it holds that 
	\begin{align*}
		&\left|\sum_{t=0}^kz_{t,k} \cdot  \frac{1}{b}\sum_{i=1}^b\dotprod{Y_{t,i} - \EE\left[Y_{t,i}\right] , \frac{1}{2} V_t^\top \Sigma_t^{-1/2} \left( \Sigma_t^{-1} - H^* \right) \Sigma_t^{-1/2} V_t }\right| \\
		\stackrel{\eqref{eq:Y0}}{=}&
		\left|\sum_{t=0}^kz_{t,k}\dotprod{\EE\left[Y_{t,i}\right] , \frac{1}{2} V_t^\top \Sigma_t^{-1/2} \left( \Sigma_t^{-1} - H^* \right)\Sigma_t^{-1/2} V_t }\right|\\
		\le& \sum_{t=0}^k z_{t,k} \cdot \norm{ \EE[Y_t] } \cdot \sum_{i=1}^d \left|\lambda_i\left(\Sigma_t^{-1/2} \left( \Sigma_t^{-1} - H^* \right) \Sigma_t^{-1/2}\right)\right|\\
		\le&
		\sum_{t=0}^k z_{t,k} \cdot \frac{3\tr(\Lambda_t)}{d} \cdot \frac{\delta \left(A_t^2 + 16 A_t + 144\right)}{(t+1)^2} \cdot \tr(\Sigma_t^{-1}) \cdot \norm{ \Sigma_t^{-1} - H^*},
	\end{align*}
	where the first equality is also because $Y_{t,i}$'s have the independent and identical distribution and   the last two inequalities are because of the von Neumann’s trace theorem.
\end{proof}

\begin{lemma}
	It holds that
	\begin{equation}\label{eq:sto_err}\small
		\begin{aligned}
			& \sum_{t=2}^k \frac{t-1}{k(k-1)} \dotprod{ \frac{1}{2b} \sum_{i=1}^b u_i^\top \Sigma_t^{1/2} H_t \Sigma_t^{1/2} u_i \left(\Sigma_t^{-1/2} u_i u_i^\top \Sigma_t^{-1/2} - \Sigma_t^{-1}\right)- H_t,  \Sigma_t^{-1} - H^*  } \\
			\le &
			\frac{A_k L (\zeta A_k + d \zeta) }{k(k-1)\tau} \cdot \sqrt{\frac{2}{b}\log \frac{k^2}{\delta}} \cdot \sqrt{ \sum_{t=2}^k (t-1)^2 \norm{\Sigma_t - H^*}^2 }.
		\end{aligned}
	\end{equation}
\end{lemma}
\begin{proof}
	\begin{align*}
		&\frac{1}{k(k-1)} \sum_{t=2}^k (t-1) \dotprod{ \frac{1}{2b} \sum_{i=1}^b u_i^\top \Sigma_t^{1/2} H_t \Sigma_t^{1/2} u_i \left(\Sigma_t^{-1/2} u_i u_i^\top \Sigma_t^{-1/2} - \Sigma_t^{-1}\right)- H_t,  \Sigma_t^{-1} - H^*  } \\
		\stackrel{\eqref{eq:E}}{=}& 
		\frac{1}{k(k-1)} \sum_{t=2}^k (t-1) \dotprod{\frac{1}{2} \sum_{i=1}^b E_{t,i},  V_t^\top \Sigma_t^{-1/2} \left(\Sigma_t^{-1} - H^*\right) \Sigma_t^{-1/2} V_t} \\
		\stackrel{\eqref{eq:EXY}}{=}&
		\sum_{t=2}^k \frac{(t-1)}{k(k-1)} \dotprod{\frac{1}{2} \sum_{i=1}^b \left[\Big(X_{t,i} - \EE[X_{t,i}]\Big) + \Big(Y_{t,i} - \EE[Y_{t,i}]\Big)\right],  V_t^\top \Sigma_t^{-1/2} \left(\Sigma_t^{-1} - H^*\right) \Sigma_t^{-1/2} V_t}.
	\end{align*}
	Let $z_{t,k} = \frac{(t-1)}{k(k-1)}$ and $A_t = 8\log \frac{1}{b\delta} + 16 \log( t + 1) + d$ just as set in Lemma~\ref{lem:Y}. 
	Using the facts $\norm{\Lambda_t}_2 \le \norm{\Sigma_t}_2 \cdot \norm{H_t}_2 \le L/\tau$ and $\tr(\Sigma_t^{-1}) \le d \zeta$, resorting to Lemma~\ref{lem:X}, then with a probability at least $1 - \frac{\delta}{k^2}$, it holds that
	\begin{align*}
		&\sum_{t=2}^k \frac{(t-1)}{k(k-1)} \dotprod{\frac{1}{2} \sum_{i=1}^b \left(  X_{t,i} - \EE[X_{t,i}]\right),   V_t^\top \Sigma_t^{-1/2} \left(\Sigma_t^{-1} - H^*\right) \Sigma_t^{-1/2} V_t } \\
		\le&
		\sqrt{ \frac{2}{b} \log \frac{k^2}{\delta} \cdot \sum_{t=2}^k \frac{(t-1)^2}{k^2(k-1)^2} \cdot \left(A_t \norm{\Lambda_t}_2 \left( \zeta A_t +  \tr\left(\Sigma_t^{-1}\right) \right) \cdot \norm{ \Sigma_t^{-1} - H^* }_2\right)^2 }\\
		\le&
		\sqrt{ \frac{2}{b} \log \frac{k^2}{\delta} \cdot \Big(A_k L\tau^{-1} (\zeta A_k + d \zeta) \Big)^2 \cdot  \sum_{t=2}^k \frac{(t-1)^2 \norm{ \Sigma_t ^{-1} - H^* }_2^2}{k^2(k-1)^2}}\\
		=&
		\frac{A_k L (\zeta A_k + d \zeta) }{k(k-1)\tau} \cdot \sqrt{\frac{2}{b}\log \frac{k^2}{\delta}} \cdot \sqrt{ \sum_{t=2}^k (t-1)^2 \norm{\Sigma_t - H^*}^2 }.
	\end{align*}
\end{proof}

\begin{lemma}
	Given $k \ge 2$ and letting $H_t = \Pi_{\cS'}(\nabla^2 f(\mu_t))$ and $H^* = \Pi_{\cS'}(\nabla^2 f(\mu^*))$,  then  it holds that
	\begin{equation} \label{eq:dd}
		\begin{aligned}
			\sum_{t=2}^{k}\norm{H_t - H^*}^2 
			\le 
			\frac{2\gamma^2}{\rho_1\sigma } \cdot \Big( \big(f(\mu_1) - f(\mu^*)\big) +  (k-1) \max\{\Delta_{\alpha,1}, \Delta_{\alpha,2} \}\Big)	
		\end{aligned}
	\end{equation}
	with $\rho_1 =  \frac{c_2\tau\sigma  }{16Lc_1\zeta }$,  $\Delta_{\alpha,1}$ and $\Delta_{\alpha,2}$ being defined in Lemma~\ref{lem:gd_dec_1} and~\ref{lem:dec_local}.
\end{lemma}
\begin{proof}
	Since when $\tau\cdot I \preceq \Sigma^{-1} \preceq \zeta\cdot I $, then, it is easy to check that
	\begin{align}
		\frac{\sigma}{\zeta} \cdot \Sigma_t^{-1} \preceq	\nabla^2 f(\mu_t) \preceq \frac{L}{\tau} \cdot \Sigma_t^{-1}. \label{eq:ddd}
	\end{align}
	Thus, Eqn.~\eqref{eq:dec} shows that
	\begin{align*}
		f(\mu_{k+1}) - f(\mu^*) \le (1 - \rho_1) \Big(f(\mu_k) - f(\mu^*)\Big) + \Delta_{\alpha,2}.
	\end{align*}
	Combining with Lemma~\ref{lem:gd_dec_1}, we can obtain that for $k = 1, 2, \dots$, it holds that 
	\begin{align*}
		f(\mu_{k+1}) - f(\mu^*) \le (1 - \rho_1) \Big(f(\mu_k) - f(\mu^*)\Big) + \max\{ \Delta_{\alpha,1},  \Delta_{\alpha,2}\}. 
	\end{align*}
	By the $\gamma$-Lipschitz Hessian assumption and $\sigma$-strong convexity, we can obtain that
	\begin{align*}
		&\norm{H_t - H^*}^2  \le \norm{\nabla^2 f(\mu_k) - \nabla^2 f(\mu^*)}^2 
		\le \gamma^2 \norm{\mu_k - \mu^*}^2 \le \frac{2\gamma^2}{\sigma} \Big(f(\mu_k) - f(\mu^*)\Big) \\
		\le& \frac{2\gamma^2}{\sigma}\left( (1- \rho_1)^{k-1} \big(f(\mu_1) - f(\mu^*)\big) + \max\{\Delta_{\alpha,1}, \Delta_{\alpha,2} \}\cdot\sum_{i=0}^{k-2} (1 - \rho_1)^i \right)
	\end{align*}
	Thus, 
	\begin{align*}
		&\sum_{t=2}^{k}\norm{\nabla^2 f(\mu_t) - \nabla^2 f(\mu^*)}^2
		\le
		\frac{2\gamma^2}{\sigma}\sum_{t=2}^{k}\left( (1- \rho_1)^{t-1} \big(f(\mu_1) - f(\mu^*)\big) + \max\{\Delta_{\alpha,1}, \Delta_{\alpha,2} \} \sum_{i=0}^{t-2} (1 - \rho_1)^i \right)\\
		=&\frac{2\gamma^2}{\sigma} \cdot \frac{(1-\rho_1)}{\rho_1} \left(\left(1 - (1 - \rho_1)^{k-1}\right)\big(f(\mu_1) - f(\mu^*)\big) + \left((k-1) - \frac{1 - (1 - \rho_1)^{k-1}}{\rho_1}\right) \max\{\Delta_{\alpha,1}, \Delta_{\alpha,2} \}\right) \\
		\le& 
		\frac{2\gamma^2}{\rho_1\sigma } \cdot \Big( \big(f(\mu_1) - f(\mu^*)\big) +  (k-1) \max\{\Delta_{\alpha,1}, \Delta_{\alpha,2} \}\Big).
	\end{align*}
\end{proof}

\begin{proof}[Proof of Lemma~\ref{lem:Sig_dec}]
	Combining  Lemma~\ref{lem:GH} with Eqn.~\eqref{eq:Sig_dec},~\eqref{eq:G_norm_1},~\eqref{eq:alpha_err},~\eqref{eq:sto_err}~\eqref{eq:dd}, we have
	\begin{equation}\label{eq:dec_sig}
		\begin{aligned}
			&\norm{\Sigma_{k+1}^{-1}-H^*}^2  \\
			\le&  
			\frac{1}{k(k-1)} \sum_{t=2}^k (t-1) \dotprod{\BG(\Sigma_t) - H^*, \Sigma_t^{-1} - H^*}
			+
			\frac{1}{k(k-1)} \sum_{t=2}^{k}\frac{t-1}{t}\norm{\TG(\Sigma_t)}^2\\
			\le&  
			\frac{C_1 + C_2 + C_3}{k(k-1)}\cdot \sqrt{ \sum_{t=2}^k (t-1)^2 \norm{\Sigma_t - H^*}^2 }  + \frac{C_4}{k}.  
		\end{aligned}
	\end{equation}
	
	Next, we will prove the result by induction. We assume that $\norm{\Sigma_t - H^*}^2 \le C / t$ holds 
	for all $t = 1,\dots, k$. 
	First, for the case $ t = 1$, we have
	\begin{align*}
		\norm{\Sigma_1^{-1}-H^*}^2 \le \frac{\norm{\Sigma_1^{-1}-H^*}^2}{1} \le \frac{C}{1},
	\end{align*} 
	and 
	\begin{align*}
		\norm{\Sigma_2^{-1}-H^*}^2 
		\stackrel{\eqref{eq:Sig_2_bnd}}{\le}
		\frac{4L^2 (2d+3\log(1/\delta))^2 \left( \zeta (2d + 3\log(1/\delta)) + \sqrt{d} \zeta\right)^2 }{b\tau^2} + \frac{d \zeta^2}{b}
		\le 
		\frac{C}{2}.
	\end{align*}
	Next, we will prove the case $t = k+1$. By Lemma~\ref{lem:sgd_dec}, Eqn.~\eqref{eq:dec_sig} and the definition of $C$, it holds that
	\begin{align*}
		\norm{\Sigma_{k+1}^{-1}-H^*}^2 
		\le
		\frac{C}{k+1}.
	\end{align*}
	Thus, the induction result holds for $t = k+1$ which concludes the proof.
	
\end{proof}

\section{Proof of Theorem~\ref{thm:main_cvg}}

\begin{proof}
	For the case before entering the local region, the result can be obtained by Lemma~\ref{lem:gd_dec_1}.
	For the case after entering the local region, we have
	\begin{align*}
		&\norm{\Sigma_k^{1/2}\Big(\Sigma_k^{-1} - \nabla^2 f(\mu_k)\Big)\Sigma_k^{1/2}}_2 
		\le \norm{\Sigma}_2 \cdot \norm{\Sigma_k^{-1} - \nabla^2 f(\mu_k)}_2
		\le \tau^{-1}\cdot \norm{\Sigma_k^{-1} - \nabla^2 f(\mu_k)}\\
		\le&
		\tau^{-1} \cdot \big(\norm{ \Sigma_k^{-1} - H^*}  + \norm{H^*- \nabla^2 f(\mu_k) }\big)
		\le
		\tau^{-1} \cdot \left(\sqrt{\frac{C}{k}} + \gamma \norm{\mu_k - \mu^*} + \norm{ \nabla^2 f(\mu_k)  - \Pi_{\cS'}\left(\nabla^2 f(\mu_k)\right)}\right)\\
		\le&
		\tau^{-1} \cdot \left(\sqrt{\frac{C}{k}} + \gamma \sqrt{ \frac{2\big(f(\mu_k) - f(\mu^*)\big)}{\sigma}} + \norm{ \nabla^2 f(\mu_k)  - \Pi_{\cS'}\left(\nabla^2 f(\mu_k)\right)}\right),
	\end{align*}
	where the last inequality is because of $f(\cdot)$ is $\sigma$-strongly convex.
	Above equation implies that
	\begin{equation*}
		\small
		\begin{aligned}
			&\left(1 - \tau^{-1} \cdot \left(\sqrt{\frac{C}{k}} + \gamma \sqrt{ \frac{2\big(f(\mu_k) - f(\mu^*)\big)}{\sigma}}+ \norm{ \nabla^2 f(\mu_k)  - \Pi_{\cS'}\left(\nabla^2 f(\mu_k)\right)}\right)\right) \cdot \Sigma_k^{-1} \preceq      H_k \\
			&\preceq \left(1 + \tau^{-1} \cdot \left(\sqrt{\frac{C}{k}} + \gamma \sqrt{ \frac{2\big(f(\mu_k) - f(\mu^*)\big)}{\sigma}} + \norm{ \nabla^2 f(\mu_k)  - \Pi_{\cS'}\left(\nabla^2 f(\mu_k)\right)}\right)\right) \cdot \Sigma_k^{-1}.
		\end{aligned}
	\end{equation*}
	Furthermore, it holds that $ \tau \cdot I \preceq \Sigma_k^{-1} \preceq \zeta \cdot I $ and $\sigma \cdot I \preceq  H \preceq L \cdot I$, which implies Eqn.~\eqref{eq:ddd}.
	Combining above results with Lemma~\ref{lem:dec_local} concludes the proof.	
\end{proof}

\vskip 0.2in
\bibliography{ref.bib}

\end{document}